\pdfoutput=1

\documentclass[10pt,twocolumn,letterpaper]{article}

\usepackage[pagenumbers]{cvpr} %

{}
\usepackage{booktabs}
\usepackage{multirow}
\usepackage{xspace}

\newcommand{\method}{UniHDR\xspace}
\newcommand{\sihdr}{SI-HDR\xspace}

\definecolor{OursRed}{HTML}{D9261C}

\renewcommand{\textfraction}{0.05}

\renewcommand{\dbltopfraction}{0.95}      %
\renewcommand{\dblfloatpagefraction}{0.9}
\usepackage{colortbl}
\definecolor{bestcolor}{RGB}{255,179,179}   %
\definecolor{secondcolor}{RGB}{255,255,179}  %
\newcommand{\legendbox}[2]{{\setlength{\fboxsep}{1pt}\colorbox{#1}{#2}}}  %

\newcommand{\rpar}[1]{\par\vspace{5pt}\noindent\textbf{#1}}

\newif\ifqualfigplaced
\newif\ifpanofigplaced
\newcommand{\qualfigurefloat}{\ifqualfigplaced\else\global\qualfigplacedtrue\begin{figure*}[t]
  \centering
  \includegraphics[width=\textwidth, trim=0 1.50bp 0 0.00bp, clip]{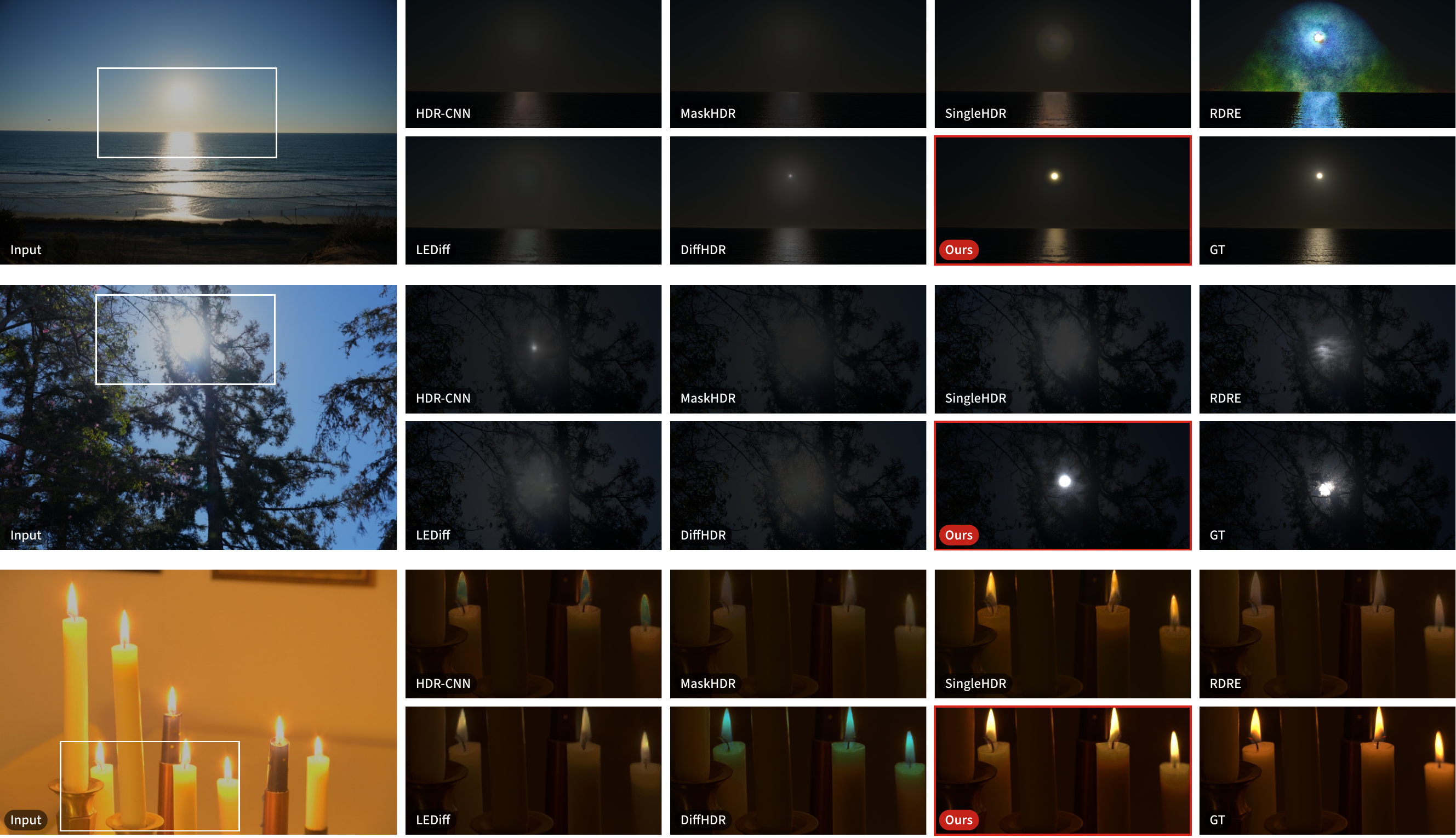}  %
  \caption{\textbf{Qualitative comparison on perspective images.}
  Left: the 8-bit input with the compared window. Right: crops of the compared window. Rows 1 and 2 are camera JPEGs from the 46 scenes we photographed, with a ground truth (GT) merged from exposure brackets. Row 3 is an \sihdr~\cite{hanji2022sihdr} image. All crops are at $-6$\,EV.}
  \label{fig:qual}
\end{figure*}
{}\fi}
\newcommand{\panofigurefloats}{\qualfigurefloat\global\panofigplacedtrue\begin{figure*}[t]
  \centering
  \includegraphics[width=\textwidth, trim=0 11.25bp 0 3.25bp, clip]{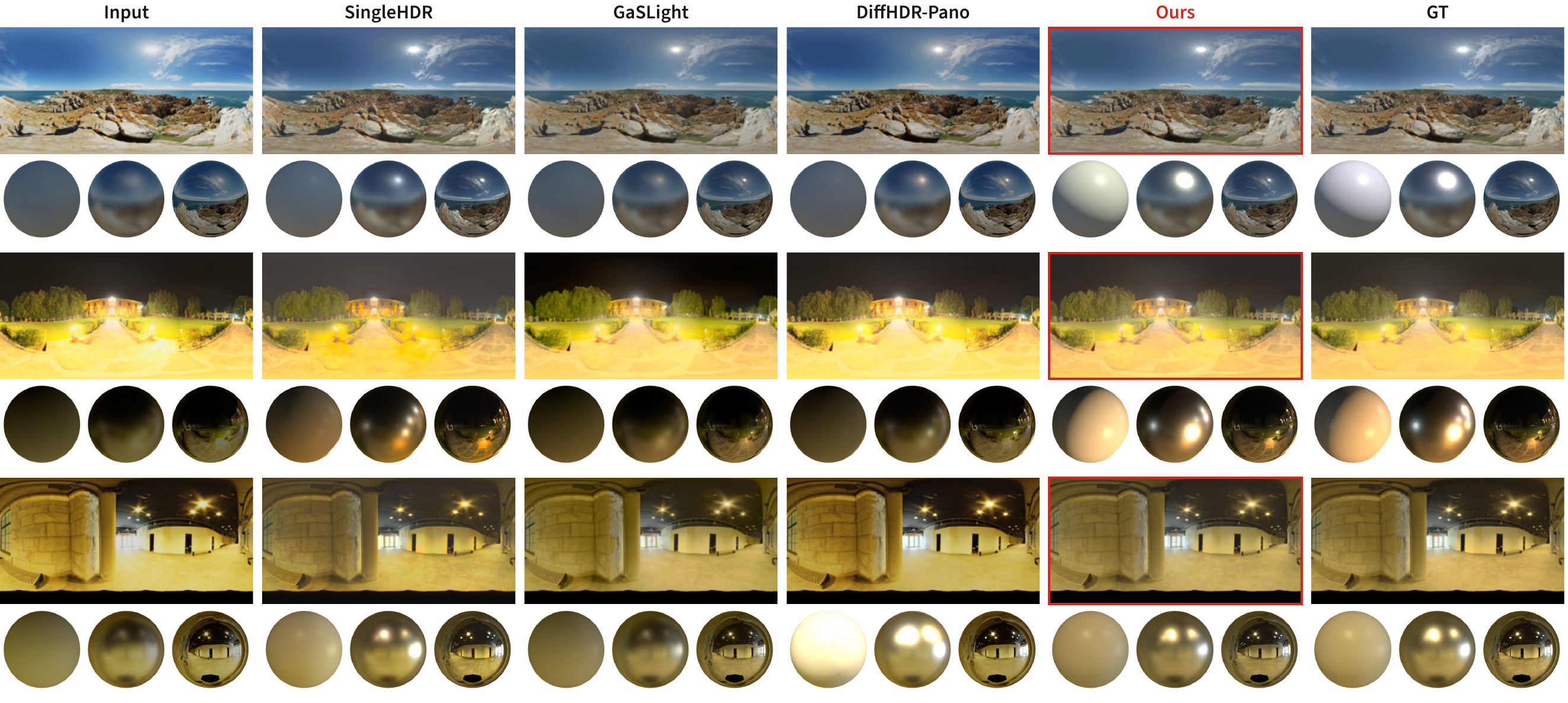}  %
  \caption{\textbf{Panorama lighting comparison.}
  Each scene shows the panorama (top) and three spheres path-traced under it (bottom): diffuse, matte silver and mirror, from left to right. Columns: the 8-bit input used as lighting, three baselines, \method and the ground truth (GT), at one exposure and white point per row. Rows: two outdoor Poly Haven~\cite{polyhaven2026hdris} scenes and a Laval~\cite{gardner2017indoor} interior.}
  \label{fig:pano_spheres}
\end{figure*}
{}\begin{table*}[t]
\centering
\footnotesize
\setlength{\tabcolsep}{2.85pt}
\caption{\textbf{Single-image HDR reconstruction on \sihdr~\cite{hanji2022sihdr}} (181 scenes); metrics as in \cref{sec:protocol}. Mean{\scriptsize$\pm$std} over scenes; FID is one value over $128^2$ crops after Reinhard~\cite{reinhard2002photographic} tone mapping, and PIQE is no-reference. Results are colour-coded by \legendbox{bestcolor}{best} and \legendbox{secondcolor}{second best} per column. The last row scores the ground truth against itself.}
\label{tab:sihdr}
\begin{tabular}{lcccccccccc}
\toprule
\multirow{2}{*}[-2.4pt]{\textbf{Method}} & \multicolumn{6}{c}{\textbf{Full image}} & \multicolumn{4}{c}{\textbf{Clipped region}} \\
\cmidrule(lr){2-7} \cmidrule(lr){8-11}
 & PU21-PSNR $\uparrow$ & PU21-SSIM $\uparrow$ & VSI $\uparrow$ & HDR-VDP-3 $\uparrow$ & FID $\downarrow$ & PIQE $\downarrow$ & PSNR-H $\uparrow$ & HEE $\downarrow$ & HQE $\downarrow$ & HSE $\downarrow$ \\
\midrule
HDR-CNN~\cite{eilertsen2017hdrcnn} & 21.43{\scriptsize$\pm$3.47} & 0.811{\scriptsize$\pm$0.136} & 0.977{\scriptsize$\pm$0.017} & 7.37{\scriptsize$\pm$0.80} & 18.19 & 21.68{\scriptsize$\pm$9.19} & 18.91{\scriptsize$\pm$4.93} & 1.26{\scriptsize$\pm$1.32} & 1.20{\scriptsize$\pm$0.99} & 0.64{\scriptsize$\pm$0.45} \\
ExpandNet~\cite{marnerides2018expandnet} & 21.11{\scriptsize$\pm$4.28} & 0.855{\scriptsize$\pm$0.067} & 0.973{\scriptsize$\pm$0.020} & 7.18{\scriptsize$\pm$0.96} & 17.70 & \cellcolor{secondcolor}15.85{\scriptsize$\pm$5.56} & 17.86{\scriptsize$\pm$5.74} & 1.10{\scriptsize$\pm$1.13} & 1.17{\scriptsize$\pm$0.88} & 0.80{\scriptsize$\pm$0.66} \\
MaskHDR~\cite{santos2020maskhdr} & 21.56{\scriptsize$\pm$3.42} & 0.811{\scriptsize$\pm$0.136} & 0.978{\scriptsize$\pm$0.016} & 7.41{\scriptsize$\pm$0.82} & 18.35 & 22.34{\scriptsize$\pm$9.66} & 19.12{\scriptsize$\pm$4.77} & 1.26{\scriptsize$\pm$1.33} & 1.18{\scriptsize$\pm$0.97} & 0.63{\scriptsize$\pm$0.46} \\
SingleHDR~\cite{liu2020singlehdr} & 21.06{\scriptsize$\pm$3.13} & \cellcolor{secondcolor}0.887{\scriptsize$\pm$0.065} & 0.979{\scriptsize$\pm$0.013} & 7.62{\scriptsize$\pm$0.68} & 17.13 & 21.75{\scriptsize$\pm$9.36} & 16.38{\scriptsize$\pm$4.45} & 0.93{\scriptsize$\pm$0.99} & 0.85{\scriptsize$\pm$0.68} & 0.68{\scriptsize$\pm$0.45} \\
IntrinsicHDR~\cite{dille2024intrinsichdr} & 23.24{\scriptsize$\pm$3.52} & 0.885{\scriptsize$\pm$0.063} & 0.982{\scriptsize$\pm$0.014} & 7.70{\scriptsize$\pm$0.81} & 15.59 & 21.70{\scriptsize$\pm$7.67} & 19.29{\scriptsize$\pm$4.97} & 1.04{\scriptsize$\pm$1.15} & 0.97{\scriptsize$\pm$0.83} & 0.62{\scriptsize$\pm$0.45} \\
RDRE~\cite{dille2026rdre} & \cellcolor{secondcolor}23.43{\scriptsize$\pm$3.88} & 0.841{\scriptsize$\pm$0.102} & \cellcolor{secondcolor}0.985{\scriptsize$\pm$0.013} & \cellcolor{secondcolor}7.74{\scriptsize$\pm$0.86} & 17.19 & 19.76{\scriptsize$\pm$9.02} & \cellcolor{secondcolor}20.89{\scriptsize$\pm$5.61} & 0.97{\scriptsize$\pm$1.12} & 0.85{\scriptsize$\pm$0.79} & \cellcolor{secondcolor}0.55{\scriptsize$\pm$0.43} \\
\midrule
GaSLight~\cite{bolduc2025gaslight} & 22.41{\scriptsize$\pm$2.99} & 0.851{\scriptsize$\pm$0.096} & 0.972{\scriptsize$\pm$0.014} & 7.24{\scriptsize$\pm$0.67} & 22.18 & 26.71{\scriptsize$\pm$11.49} & 17.68{\scriptsize$\pm$4.04} & \cellcolor{secondcolor}0.83{\scriptsize$\pm$0.79} & \cellcolor{secondcolor}0.82{\scriptsize$\pm$0.60} & 0.65{\scriptsize$\pm$0.38} \\
LEDiff~\cite{wang2025lediff} & 18.77{\scriptsize$\pm$3.60} & 0.753{\scriptsize$\pm$0.145} & 0.968{\scriptsize$\pm$0.018} & 6.92{\scriptsize$\pm$0.78} & 18.33 & 23.11{\scriptsize$\pm$8.81} & 16.27{\scriptsize$\pm$5.32} & 1.27{\scriptsize$\pm$1.31} & 1.16{\scriptsize$\pm$0.96} & 0.75{\scriptsize$\pm$0.57} \\
DiffHDR~\cite{yu2026diffhdr} & 22.08{\scriptsize$\pm$3.45} & 0.843{\scriptsize$\pm$0.079} & 0.978{\scriptsize$\pm$0.015} & 7.28{\scriptsize$\pm$0.71} & \cellcolor{secondcolor}14.50 & \cellcolor{bestcolor}15.25{\scriptsize$\pm$6.01} & 18.23{\scriptsize$\pm$4.57} & 1.09{\scriptsize$\pm$1.12} & 0.94{\scriptsize$\pm$0.75} & 0.62{\scriptsize$\pm$0.44} \\
\midrule
 \textbf{\method} (ours) & \cellcolor{bestcolor}27.93{\scriptsize$\pm$3.25} & \cellcolor{bestcolor}0.934{\scriptsize$\pm$0.036} & \cellcolor{bestcolor}0.991{\scriptsize$\pm$0.008} & \cellcolor{bestcolor}8.35{\scriptsize$\pm$0.64} & \cellcolor{bestcolor}8.75 & 37.21{\scriptsize$\pm$15.49} & \cellcolor{bestcolor}23.58{\scriptsize$\pm$4.74} & \cellcolor{bestcolor}0.54{\scriptsize$\pm$0.59} & \cellcolor{bestcolor}0.49{\scriptsize$\pm$0.41} & \cellcolor{bestcolor}0.33{\scriptsize$\pm$0.26} \\
\midrule
GT & $\infty$ & 1.000 & 1.000 & 10.00 & 0.00 & 41.45 & $\infty$ & 0.00 & 0.00 & 0.00 \\
\bottomrule
\end{tabular}
\end{table*}
{}\begin{table*}[t]
\centering
\footnotesize
\setlength{\tabcolsep}{2pt}
\caption{\textbf{Panorama reconstruction scored as lighting} (\cref{sec:protocol}). PU21-PSNR is computed on the panorama, RMSE and angular error on three path-traced spheres, and Energy on the diffuse sphere. Sun is the error of the sun direction on the 58 daytime outdoor scenes. Mean{\scriptsize$\pm$std} over scenes. Results are colour-coded by \legendbox{bestcolor}{best} and \legendbox{secondcolor}{second best} per column.}
\label{tab:pano}
\begin{tabular}{clccccccccc}
\toprule
 & \multirow{2}{*}[-2.4pt]{\textbf{Method}} & \multirow{2}{*}[-2.4pt]{\textbf{PU21-PSNR} $\uparrow$} & \multicolumn{3}{c}{\textbf{Scale-invariant RMSE} $\downarrow$} & \multicolumn{3}{c}{\textbf{Angular error} ($^\circ$) $\downarrow$} & \multirow{2}{*}[-2.4pt]{\textbf{Sun} ($^\circ$) $\downarrow$} & \multirow{2}{*}[-2.4pt]{\textbf{Energy} (stops) $\downarrow$} \\
\cmidrule(lr){4-6} \cmidrule(lr){7-9}
 &  &  & Diffuse & Matte silver & Mirror & Diffuse & Matte silver & Mirror & & \\
\midrule
\multirow{8}{*}[-4.6pt]{\rotatebox[origin=c]{90}{\begin{tabular}{@{}c@{}}Poly Haven~\cite{polyhaven2026hdris}\\ 117 outdoor scenes\end{tabular}}} & HDR-CNN~\cite{eilertsen2017hdrcnn} & 23.73{\scriptsize$\pm$2.51} & 0.405{\scriptsize$\pm$0.670} & 1.605{\scriptsize$\pm$2.480} & 34.712{\scriptsize$\pm$51.598} & 5.24{\scriptsize$\pm$4.49} & 4.25{\scriptsize$\pm$1.73} & 4.34{\scriptsize$\pm$1.07} & 6.54{\scriptsize$\pm$18.38} & 1.04{\scriptsize$\pm$1.05} \\
 & SingleHDR~\cite{liu2020singlehdr} & \cellcolor{secondcolor}30.61{\scriptsize$\pm$3.39} & 0.339{\scriptsize$\pm$0.567} & 1.392{\scriptsize$\pm$2.222} & 34.425{\scriptsize$\pm$51.236} & \cellcolor{secondcolor}4.37{\scriptsize$\pm$3.85} & \cellcolor{secondcolor}3.02{\scriptsize$\pm$1.60} & \cellcolor{secondcolor}2.36{\scriptsize$\pm$0.82} & 9.50{\scriptsize$\pm$18.49} & \cellcolor{secondcolor}0.85{\scriptsize$\pm$0.91} \\
 & IntrinsicHDR~\cite{dille2024intrinsichdr} & 30.23{\scriptsize$\pm$3.76} & 0.374{\scriptsize$\pm$0.643} & 1.505{\scriptsize$\pm$2.419} & 35.162{\scriptsize$\pm$52.750} & 4.92{\scriptsize$\pm$4.86} & 3.32{\scriptsize$\pm$2.21} & 2.40{\scriptsize$\pm$1.38} & 7.39{\scriptsize$\pm$18.09} & 0.93{\scriptsize$\pm$1.01} \\
 & RDRE~\cite{dille2026rdre} & 23.82{\scriptsize$\pm$2.26} & 0.372{\scriptsize$\pm$0.567} & 1.489{\scriptsize$\pm$2.187} & 34.757{\scriptsize$\pm$52.354} & 5.29{\scriptsize$\pm$4.79} & 4.29{\scriptsize$\pm$2.16} & 3.97{\scriptsize$\pm$0.93} & 9.54{\scriptsize$\pm$23.19} & 0.95{\scriptsize$\pm$1.00} \\
\cmidrule(l){2-11}
 & GaSLight~\cite{bolduc2025gaslight} & 25.90{\scriptsize$\pm$3.26} & \cellcolor{secondcolor}0.334{\scriptsize$\pm$0.617} & \cellcolor{secondcolor}1.332{\scriptsize$\pm$2.278} & 34.274{\scriptsize$\pm$51.325} & 4.66{\scriptsize$\pm$3.23} & 3.60{\scriptsize$\pm$1.66} & 3.31{\scriptsize$\pm$1.35} & 6.78{\scriptsize$\pm$18.53} & 0.93{\scriptsize$\pm$0.94} \\
 & LEDiff~\cite{wang2025lediff} & 20.61{\scriptsize$\pm$2.62} & 0.403{\scriptsize$\pm$0.652} & 1.601{\scriptsize$\pm$2.452} & 34.680{\scriptsize$\pm$50.657} & 5.76{\scriptsize$\pm$4.90} & 5.05{\scriptsize$\pm$2.13} & 5.58{\scriptsize$\pm$1.36} & 8.43{\scriptsize$\pm$17.92} & 1.00{\scriptsize$\pm$1.00} \\
 & DiffHDR-Pano~\cite{yu2026promoting} & 23.78{\scriptsize$\pm$2.15} & 0.371{\scriptsize$\pm$0.591} & 1.488{\scriptsize$\pm$2.323} & \cellcolor{secondcolor}33.492{\scriptsize$\pm$51.200} & 5.27{\scriptsize$\pm$4.67} & 4.29{\scriptsize$\pm$1.84} & 4.06{\scriptsize$\pm$0.76} & \cellcolor{secondcolor}5.68{\scriptsize$\pm$19.39} & 0.92{\scriptsize$\pm$1.00} \\
\cmidrule(l){2-11}
 &  \textbf{\method} (ours) & \cellcolor{bestcolor}35.54{\scriptsize$\pm$4.36} & \cellcolor{bestcolor}0.124{\scriptsize$\pm$0.249} & \cellcolor{bestcolor}0.394{\scriptsize$\pm$0.765} & \cellcolor{bestcolor}20.461{\scriptsize$\pm$35.009} & \cellcolor{bestcolor}2.41{\scriptsize$\pm$2.08} & \cellcolor{bestcolor}1.46{\scriptsize$\pm$0.98} & \cellcolor{bestcolor}0.88{\scriptsize$\pm$0.37} & \cellcolor{bestcolor}1.83{\scriptsize$\pm$4.31} & \cellcolor{bestcolor}0.40{\scriptsize$\pm$0.48} \\
\midrule
\multirow{8}{*}[-4.6pt]{\rotatebox[origin=c]{90}{\begin{tabular}{@{}c@{}}Laval~\cite{gardner2017indoor}\\ 150 indoor scenes\end{tabular}}} & HDR-CNN~\cite{eilertsen2017hdrcnn} & 21.52{\scriptsize$\pm$1.88} & 0.133{\scriptsize$\pm$0.208} & 0.520{\scriptsize$\pm$0.857} & 2.767{\scriptsize$\pm$6.255} & \cellcolor{secondcolor}3.09{\scriptsize$\pm$1.34} & 3.84{\scriptsize$\pm$0.93} & 5.14{\scriptsize$\pm$1.27} & -- & 0.56{\scriptsize$\pm$0.39} \\
 & SingleHDR~\cite{liu2020singlehdr} & 22.04{\scriptsize$\pm$2.11} & \cellcolor{secondcolor}0.113{\scriptsize$\pm$0.178} & \cellcolor{secondcolor}0.431{\scriptsize$\pm$0.717} & \cellcolor{secondcolor}2.643{\scriptsize$\pm$5.367} & 3.98{\scriptsize$\pm$2.01} & 4.15{\scriptsize$\pm$1.42} & 5.18{\scriptsize$\pm$1.69} & -- & 0.58{\scriptsize$\pm$0.44} \\
 & IntrinsicHDR~\cite{dille2024intrinsichdr} & 24.75{\scriptsize$\pm$3.19} & 0.138{\scriptsize$\pm$0.224} & 0.555{\scriptsize$\pm$0.940} & 3.069{\scriptsize$\pm$6.587} & 3.17{\scriptsize$\pm$1.86} & 3.09{\scriptsize$\pm$1.44} & 3.40{\scriptsize$\pm$1.48} & -- & 0.47{\scriptsize$\pm$0.45} \\
 & RDRE~\cite{dille2026rdre} & 21.98{\scriptsize$\pm$1.99} & 0.134{\scriptsize$\pm$0.211} & 0.546{\scriptsize$\pm$0.901} & 3.073{\scriptsize$\pm$6.548} & 3.31{\scriptsize$\pm$2.29} & 3.88{\scriptsize$\pm$1.57} & 4.64{\scriptsize$\pm$1.08} & -- & \cellcolor{secondcolor}0.45{\scriptsize$\pm$0.37} \\
\cmidrule(l){2-11}
 & GaSLight~\cite{bolduc2025gaslight} & \cellcolor{secondcolor}25.16{\scriptsize$\pm$2.77} & 0.123{\scriptsize$\pm$0.203} & 0.479{\scriptsize$\pm$0.827} & 2.882{\scriptsize$\pm$6.130} & 3.29{\scriptsize$\pm$2.26} & \cellcolor{secondcolor}2.63{\scriptsize$\pm$1.07} & \cellcolor{secondcolor}2.45{\scriptsize$\pm$1.06} & -- & 0.47{\scriptsize$\pm$0.40} \\
 & LEDiff~\cite{wang2025lediff} & 18.76{\scriptsize$\pm$1.80} & 0.151{\scriptsize$\pm$0.230} & 0.588{\scriptsize$\pm$0.935} & 2.927{\scriptsize$\pm$6.165} & 4.21{\scriptsize$\pm$1.97} & 5.07{\scriptsize$\pm$1.34} & 6.79{\scriptsize$\pm$1.60} & -- & 0.59{\scriptsize$\pm$0.41} \\
 & DiffHDR-Pano~\cite{yu2026promoting} & 22.06{\scriptsize$\pm$1.77} & 0.119{\scriptsize$\pm$0.184} & 0.450{\scriptsize$\pm$0.871} & 2.849{\scriptsize$\pm$6.494} & 4.52{\scriptsize$\pm$2.95} & 4.45{\scriptsize$\pm$1.46} & 5.01{\scriptsize$\pm$0.99} & -- & 0.49{\scriptsize$\pm$0.68} \\
\cmidrule(l){2-11}
 &  \textbf{\method} (ours) & \cellcolor{bestcolor}30.26{\scriptsize$\pm$3.73} & \cellcolor{bestcolor}0.074{\scriptsize$\pm$0.141} & \cellcolor{bestcolor}0.265{\scriptsize$\pm$0.480} & \cellcolor{bestcolor}2.087{\scriptsize$\pm$4.744} & \cellcolor{bestcolor}1.69{\scriptsize$\pm$1.21} & \cellcolor{bestcolor}1.49{\scriptsize$\pm$0.80} & \cellcolor{bestcolor}1.38{\scriptsize$\pm$0.63} & -- & \cellcolor{bestcolor}0.28{\scriptsize$\pm$0.33} \\
\bottomrule
\end{tabular}
\end{table*}
{}}
\newcommand{\panofigureanchor}{%
  \ifnum\value{page}>3 \qualfigurefloat\fi
  \ifpanofigplaced\else\ifnum\value{page}>4 \panofigurefloats\fi\fi}
\newcommand{\panofigurelast}{\ifpanofigplaced\else\panofigurefloats\fi}

{}       %

\definecolor{cvprblue}{rgb}{0.21,0.49,0.74}
\usepackage[pagebackref,breaklinks,colorlinks,allcolors=cvprblue]{hyperref}

\def\confName{CVPR}
\def\confYear{2027}

\title{UniHDR: Universal HDR Reconstruction from a Single Image}
\author{Xiaopan Chu \qquad Guanzhou Ji \qquad Zheng Zheng\\
Northeastern University\\
{\tt\small \{chu.x, g.ji, zh.zheng\}@northeastern.edu}
}

\begin{document}
\twocolumn[{%
\renewcommand\twocolumn[1][]{#1}%
\maketitle
\vspace{-14pt}%
\centering
  \captionsetup{type=figure}
  \includegraphics[width=\textwidth, trim=0 0.50bp 0 0.00bp, clip]{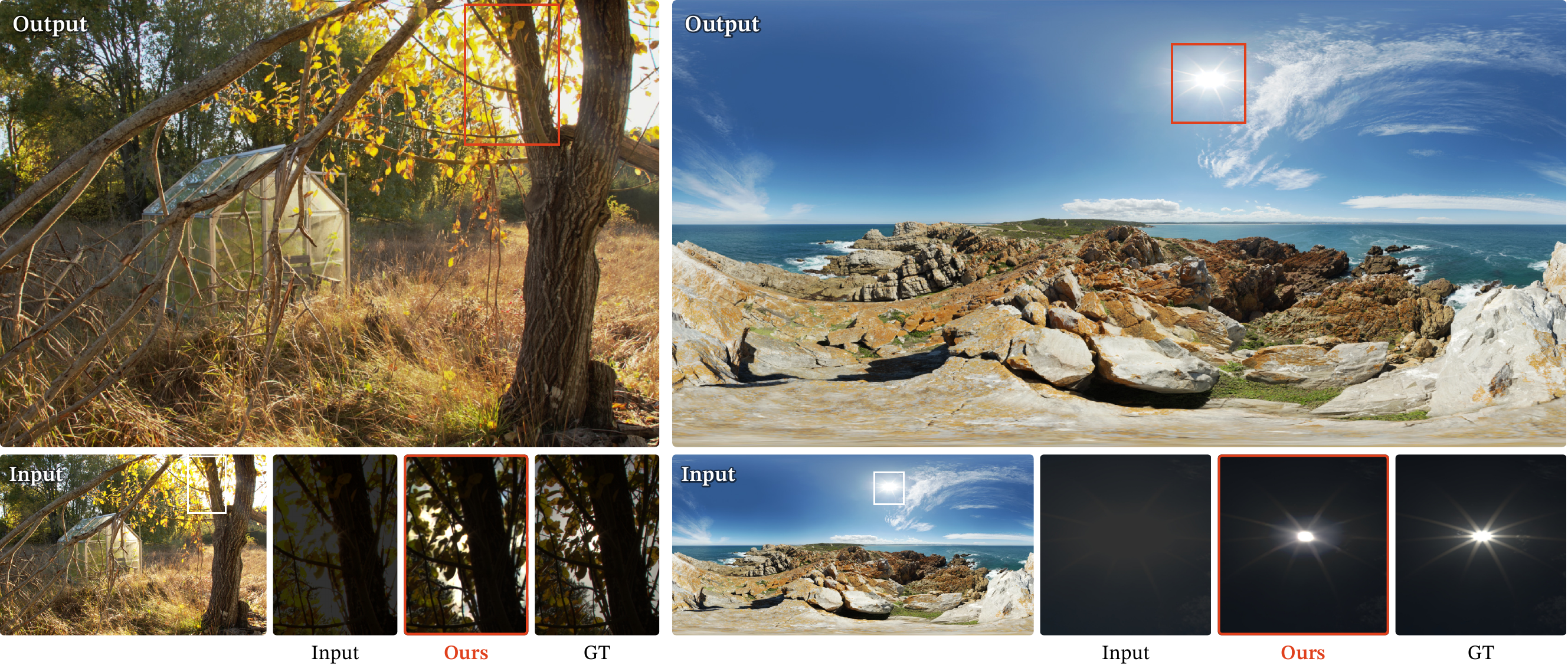}  %
  \captionof{figure}{\textbf{\method} reconstructs HDR radiance from a single 8-bit perspective image (left) or $360^\circ$ panorama (right), keeping the unclipped scene and recovering the clipped light.
  Top: our HDR output tone-mapped at the input's exposure. Bottom: the input and the boxed light source at exposure value (EV) $-5$ in the input, our output and the ground truth (GT).}
  \label{fig:pano}
  \vspace{1em}%
}]

\begin{abstract}
Single-image HDR reconstruction recovers linear radiance from an 8-bit image whose brightest content is clipped.
Diffusion methods sample a new image conditioned on the input, although every unclipped pixel already measures the scene.
We present \method, which anchors the reconstruction to what the camera captured and generates what it clipped.
We fine-tune a pretrained diffusion transformer and sample under \emph{noise-adaptive anchoring}: the unclipped region is reset to the linearised input latent at every step but the last $K$, and $K$ is set by the input's sensor noise.
Without an assumed curve or a separate estimator, we extract the inverse camera response curve from two unanchored steps of the generator to linearise the anchor.
One model serves perspective images and $360^\circ$ panoramas at input resolution.
\method outperforms prior methods in highlight recovery on perspective images, in scene lighting on panoramas scored by path-traced spheres, and in response estimation on camera and film curves. Code and models are publicly available at \url{https://github.com/xpchu/UniHDR}.
\end{abstract}
{}
\section{Introduction}
\label{sec:intro}

Single-image high dynamic range (HDR) reconstruction recovers the linear radiance of a scene from one low dynamic range (LDR) image~\cite{eilertsen2017hdrcnn,liu2020singlehdr,hanji2022sihdr}. An 8-bit image records radiance only up to its clip point, clipping the sun, lamps and bright windows (\cref{fig:pano}). Recovering these highlights is what makes an HDR image useful: it allows the image to be re-exposed, edited or shown on an HDR display, and in a $360^\circ$ panorama those highlights are the light sources of an environment map~\cite{debevec1998rendering}. The task is to restore what the sensor clipped while staying faithful to what it recorded.

Feed-forward networks predict the missing radiance, an exposure stack, or the image one stop (a factor of two) brighter at a time~\cite{eilertsen2017hdrcnn,endo2017drtmo,marnerides2018expandnet,lee2018hdrgan,liu2020singlehdr,santos2020maskhdr,dille2026rdre}, struggling to recover compact light sources many stops above the clip point. Diffusion methods sample a new image conditioned on the input, from a pretrained model adapted to HDR~\cite{wang2025lediff,yu2026diffhdr} or a pixel-space model trained for the task~\cite{andiran2026expanddiff}. Some blend the input back afterwards~\cite{wang2025lediff} or hold it as a constraint at every step~\cite{li2025sagiri,lin2026boosting}. Most single-image methods do not target $360^\circ$ panoramas at all, and the few works that restore panoramas rely on panorama-only networks or separate adapters~\cite{zhang2017outdoorhdr,yu2021lanet,yu2026promoting}.

Our key observation is that an LDR image is a measurement, not only a condition. Where a pixel is not clipped, it gives the radiance up to noise, quantisation and the camera response curve. Radiance is missing only where it is clipped (\cref{sec:measurement}). A free conditional sampler can alter what the camera recorded, whereas holding the constraint to the last step copies those measurement errors into the output. The better choice depends on the input's noise.

\begin{figure*}[t]
  \centering
  \includegraphics[width=\textwidth, trim=0 2.25bp 0 7.75bp, clip]{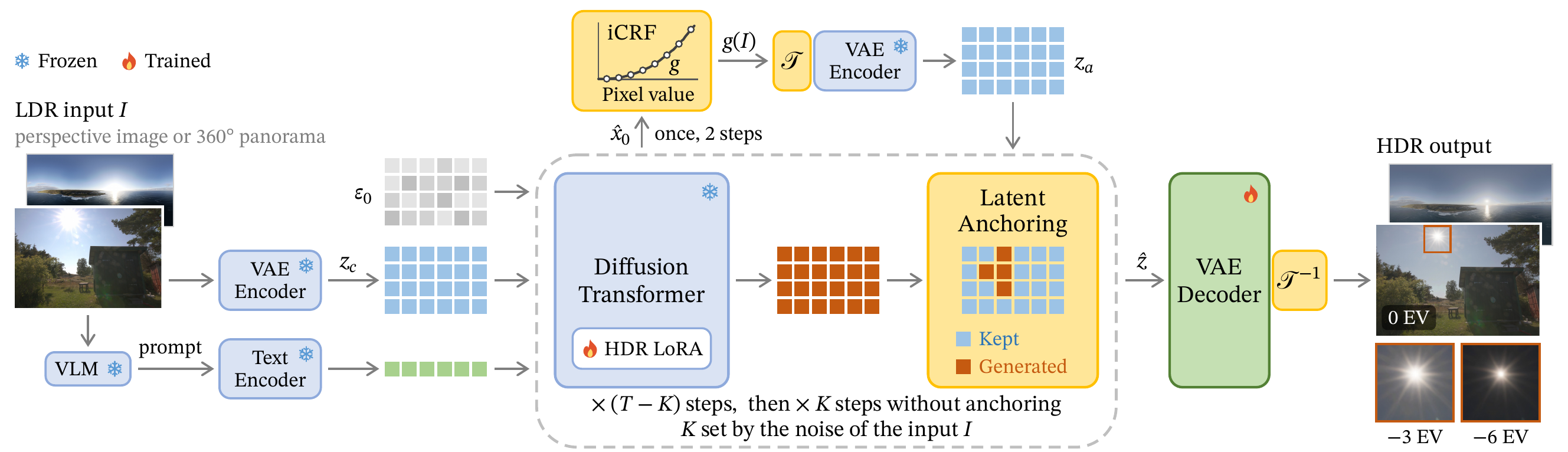}  %
  \caption{\textbf{Overview of \method.}
  The LDR input $I$ conditions the diffusion transformer through $z_c$ and a prompt from a frozen vision-language model (VLM), starting from noise $\varepsilon_0$. A single two-step unanchored pass gives $\hat{x}_0$, from which we extract the inverse camera response function (iCRF) $g$ (\cref{sec:crf}). The curve $g$ linearises the input for the anchor latent $z_a$. During the first $T - K$ steps, unclipped cells are anchored to $z_a$ (\emph{kept}) and clipped cells follow the sample (\emph{generated}). The last $K$ steps run without anchoring. $K$ is set by the sensor noise of $I$, and at $K = 0$ the unclipped pixels are $g(I)$. Right: HDR output at $0$\,EV and the boxed sun at $-3$ and $-6$\,EV.}
  \label{fig:pipeline}
\end{figure*}

We present \method, a single model that reconstructs perspective images and $360^\circ$ panoramas at input resolution in one pass (\cref{fig:pipeline}). It fine-tunes a pretrained image-editing diffusion transformer~\cite{wu2025qwenimage} with low-rank adaptation (LoRA)~\cite{hu2021lora}. A tone-map encoding spanning six decades above the clip point and a fine-tuned decoder let it output HDR radiance (\cref{sec:tonemap}). \method samples under \emph{noise-adaptive anchoring}: after every step but the last $K$, the latent of the unclipped region is reset to the linearised input latent. $K$ is set from a blind estimate of the camera's sensor noise, so a clean input is kept as measured and a noisy one is left to the generator to denoise (\cref{sec:anchor_sample,sec:ablations}).

Linearising the anchor requires the inverse camera response curve, which is unknown for an arbitrary 8-bit image. Without assuming a fixed curve or training a separate estimator, we extract the curve from the generator itself (\cref{sec:crf}). Because training exposes the model to diverse camera response curves, two unanchored steps relate input pixel values to linear radiance up to exposure. The resulting monotone curve linearises the anchor and also serves on its own for single-image response estimation.

On the \sihdr benchmark~\cite{hanji2022sihdr}, \method achieves the best mean of ten methods on every full-reference metric and the lowest Fr\'echet inception distance (FID)~\cite{heusel2017fid}, leading each baseline in paired comparison over scenes (\cref{tab:sihdr}). On held-out panoramas scored by path-traced spheres, it achieves the lowest error on every lighting measure (\cref{tab:pano}). Across measured camera and film response curves, its self-estimated curves outperform dedicated single-image estimators (\cref{tab:crf}). We also captured 46 HDR scenes to demonstrate \method on camera JPEGs (\cref{fig:qual}) and 662 RAW images to calibrate the rule that sets $K$ (\cref{sec:anchor_sample}).

\noindent Our key contributions are as follows:
\begin{itemize}
    \item We propose \method, a single model, built on a pretrained diffusion transformer, that reconstructs perspective images and $360^\circ$ panoramas across camera and film responses at input resolution in one pass (\cref{sec:regimes}).
    \item We introduce noise-adaptive anchoring: the unclipped region is held at the linearised input latent for all but the last $K$ sampling steps, with $K$ set by the sensor noise of the input (\cref{sec:anchor_sample}).
    \item We estimate the inverse camera response curve with the generative model itself: two unanchored steps give the curve of an 8-bit image without a separate estimator or an assumed curve, linearising the anchor (\cref{sec:crf}).
    \item We introduce a highlight radiance error for the clipped region and a protocol that scores panorama reconstructions by path-traced spheres on held-out Laval~\cite{gardner2017indoor} and Poly Haven~\cite{polyhaven2026hdris} scenes, with every baseline re-run (\cref{sec:protocol}).
\end{itemize}

{}
\section{Related Work}
\label{sec:related}

\rpar{Single-image HDR reconstruction} \looseness=-1\relax recovers clipped highlights from one exposure. Feed-forward networks regress radiance or an exposure stack~\cite{eilertsen2017hdrcnn,endo2017drtmo,lee2018hdrgan,marnerides2018expandnet,santos2020maskhdr,chen2021hdrunet}, invert the camera pipeline~\cite{liu2020singlehdr}, extend intrinsic components~\cite{dille2024intrinsichdr} or brighten the image one stop at a time~\cite{dille2026rdre}. Generative methods invert a generative adversarial network~\cite{wang2023glowgan}, map a diffusion latent to an exposure stack without sampling~\cite{fekri2026latenthdr}, or sample conditioned on the input~\cite{wang2025lediff,bolduc2025gaslight,yu2026diffhdr,bemana2025bracket,wu2026x2hdr,talegaonkar2026singleshot,andiran2026expanddiff}, with LEDiff blending the input back afterwards~\cite{wang2025lediff}. For $360^\circ$ scenes, most work estimates lighting from a perspective view~\cite{gardner2017indoor,holdgeoffroy2019outdoor,legendre2019deeplight,zhan2021emlight,wang2022stylelight,dastjerdi2023everlight,phongthawee2024diffusionlight,hilliard2025hdrldm,liang2025luxdit,bolduc2025gaslight}, whereas restoring a complete LDR panorama relies on regression~\cite{zhang2017outdoorhdr,yu2021lanet,wei2021beyond} or a panorama diffusion adapter trained with rendering losses~\cite{yu2026promoting}. \method handles both regimes with one model, tying the unclipped region to the input during sampling instead of using it only as a condition.

\rpar{Sampler-level constraints} enforce consistency with known content in training-free editing and inverse problems~\cite{meng2022sdedit,chung2023dps}. RePaint and Blended (Latent) Diffusion replace the known region after every step with the noised input~\cite{lugmayr2022repaint,avrahami2022blended,avrahami2023bld}, a replacement that Differential Diffusion and Personalize Anything stop early~\cite{levin2025differential,feng2026personalize}. On clipped inputs, Sagiri applies this per-step latent replacement in a fine-tuned model to produce an 8-bit output~\cite{li2025sagiri}, while Lin~\etal\ blend a baseline exposure stack into a frozen inpainting model at every step~\cite{lin2026boosting}. Related HDR methods inpaint the clipped region~\cite{goswami2025ditmo} or blend trajectories under a soft exposure mask~\cite{liu2026expocm}. Our update shares this per-step blend form (\cref{sec:anchor_sample}) but anchors to the linearised input latent. Unlike previous methods for clipped inputs, it holds the constraint for a number of steps set by the sensor noise of the input.

\rpar{Camera response estimation} calibrates curves from exposure brackets~\cite{debevec1997recovering} or models them via empirical curve databases and subspaces~\cite{grossberg2003space,grossberg2004emor,chen2019moderncrf} and in-camera pipelines~\cite{kim2012incamera}. From one image, CRF-net~\cite{li2017crfnet} and SingleHDR~\cite{liu2020singlehdr} predict the curve with a dedicated network, while Sharma~\etal\ weight per-patch estimates by consistency~\cite{sharma2020crf}. Because HDR scores are sensitive to tone and colour shifts from a wrong curve~\cite{hanji2022sihdr}, \method extracts the inverse camera response curve from two unanchored steps of the generator itself to linearise the anchor, with no assumed curve or separate estimator (\cref{sec:crf}).
{}
\section{Method}
\label{sec:method}

\method reconstructs \looseness=-1\relax linear radiance from one 8-bit perspective image or $360^\circ$ panorama with a pretrained diffusion transformer fine-tuned with LoRA, producing a radiance map at input resolution in units of the input's clip point (\cref{fig:pipeline}). We introduce our measurement model and HDR encoding in \cref{sec:measurement}, noise-adaptive anchoring in \cref{sec:anchor_sample}, generator-based response estimation in \cref{sec:crf}, and the implementation in \cref{sec:impl}.

\subsection{Measurement model and HDR encoding}
\label{sec:measurement}
\label{sec:setup}
\label{sec:tonemap}

A camera integrates scene radiance $L$ over an exposure $e$, clips at sensor saturation and maps the result through a camera response curve $f$ to 8-bit pixel values. Expressing radiance in the input's own frame, $x = eL$, places the clip point at $x = 1$ and gives
\begin{equation}
I = f\big(\min(x, 1)\big) + \eta ,
\label{eq:formation}
\end{equation}
where $\eta$ captures sensor noise and quantisation. A pixel is \emph{clipped} when any channel reaches the top of the 8-bit range ($\ge 250$ of 255), partitioning the image into the clipped region $\Omega_c$ and the unclipped region $\Omega_u$. On $\Omega_u$, applying the inverse camera response curve $g = f^{-1}$ recovers $x$ up to $\eta$, so $I$ directly measures the scene. On $\Omega_c$, the observation states only that $x \ge 1$. The task is to reconstruct $x$ across the whole image, keeping what $\Omega_u$ measured and generating what $\Omega_c$ lost.

\begin{figure}[t]
  \centering
  \includegraphics[width=\linewidth, trim=0 3.25bp 0 8.00bp, clip]{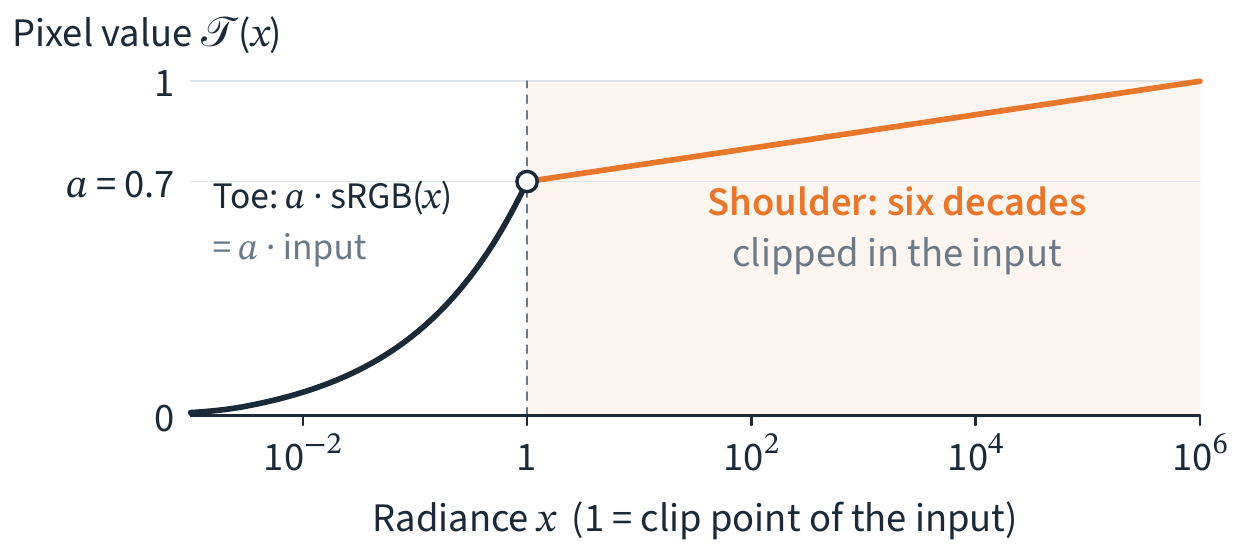}  %
  \caption{\textbf{Tone-map encoding $\mathcal{T}$.} Unclipped radiance ($x \le 1$) maps to $[0, a]$ and six clipped decades share $(a, 1]$ (\cref{eq:tonemap}).}
  \label{fig:encoding}
\end{figure}

Because the variational autoencoder (VAE) $(\mathcal{E}, \mathcal{D})$ expects display-referred images in $[0, 1]$, we propose a piecewise encoding of linear radiance,
\begin{equation}
\mathcal{T}(x) =
\begin{cases}
a\,\mathrm{sRGB}(x), & x \le 1, \\[2pt]
a + (1 - a)\,\dfrac{\log_{10} x}{d}, & 1 < x \le 10^{d},
\end{cases}
\label{eq:tonemap}
\end{equation}
with $a = 0.7$ and $d = 6$ (\cref{fig:encoding}). Below the clip point, $\mathcal{T}(x)$ scales the sRGB rendering~\cite{iec1999srgb} by $a$, so an sRGB input and its encoded radiance agree on $\Omega_u$ up to the factor $a$. Above it, six logarithmic decades share the remaining $1 - a$ of the range, covering the peak of 99.7\,\% of our training pairs, whereas five decades would clip 6.5\,\%. Since one 8-bit step on this shoulder spans 20\,\% of radiance, we store encoded targets at 16 bits.
Because the stock VAE decoder compresses the logarithmic shoulder, we fine-tune only the VAE decoder $\mathcal{D}$ on encoded radiance while keeping the VAE encoder $\mathcal{E}$ frozen to preserve the generator's latent space. Decoding a final sampled latent $z$ gives $\hat{x} = \mathcal{T}^{-1}\big(\mathcal{D}(z)\big)$.

\panofigureanchor   %
\subsection{Noise-adaptive anchoring}
\label{sec:anchor_sample}
\label{sec:prelim}

\panofigureanchor
A diffusion model that takes the input only as a condition repaints the unclipped region. The opposite extreme replaces the latent of $\Omega_u$ with the input's latent after every step until the end of sampling, which carries the input's sensor noise, quantisation and curve error into the output. Which is better depends on the input (\cref{fig:adaptive}): on a clean image, what the camera measured is already the answer, whereas on a noisy one the generator needs steps of its own to remove the noise.

\panofigureanchor
We anchor during all but the last $K$ steps. Let $z_a = \mathcal{E}\big(\mathcal{T}(g(I))\big)$ be the latent of the linearised input, where $g$ is estimated as in \cref{sec:crf}, and let $m \in [0, 1]^{h \times w}$ be the anchor weight on the $h \times w$ latent grid. Conditioned on the raw input latent $z_c = \mathcal{E}(I)$ and a text prompt, the sampler starts from Gaussian noise $z_0 = \varepsilon_0$ and takes $T$ Euler steps $z_{i+1} = z_i + (\sigma_{i+1} - \sigma_i)\, v_\theta$ on the flow-matching~\cite{lipman2023flowmatching} schedule $\sigma_0 = 1 > \dots > \sigma_T = 0$, where $v_\theta$ is the predicted velocity. After each step $i < T - K$, we set
\begin{equation}
z_{i+1} \leftarrow m \odot \big((1 - \sigma_{i+1})\, z_a + \sigma_{i+1}\, \varepsilon_0\big) + (1 - m) \odot z_{i+1} ,
\label{eq:anchor_update}
\end{equation}
and run the last $K$ steps without anchoring. The anchored cells follow the trajectory the input would follow under $\varepsilon_0$, while the clipped cells follow the model. The weight $m$ is 0 on cells containing a clipped pixel and on the ring of cells around them, then rises to 1 over the next three cells via a two-cell dilation and a one-cell Gaussian blur. The generated region thus includes the glow around a clipped source, with a smooth transition across the boundary. For panoramas the dilation and blur wrap across the left and right edges.

\panofigureanchor
$K$ controls how much of the measurement the generator may revise, and we derive it from the input itself. We estimate the sensor noise $\hat{n}$ of the input, in pixel values, with the estimator of Immerk{\ae}r~\cite{immerkaer1996fast} on $8 \times 8$ mid-tone blocks of the green channel, taking the 10th percentile over the blocks. We then set
\begin{equation}
K =
\begin{cases}
0, & \hat{n} < \tau_1, \\[2pt]
\mathrm{round}\Big(1 + (T - 2)\,\dfrac{\log(\hat{n} / \tau_1)}{\log(\tau_2 / \tau_1)}\Big), & \tau_1 \le \hat{n} < \tau_2, \\[6pt]
T, & \hat{n} \ge \tau_2,
\end{cases}
\label{eq:adaptive_k}
\end{equation}
with $\tau_1 = 1.2$ and $\tau_2 = 2.0$ calibrated on 662 RAW images we captured, with sensor noise added. Intuitively, the cleaner the input, the longer the sampler holds it. For $K = 0$ the sampler anchors at every step and the generator supplies only $\Omega_c$. Because even an anchored latent loses detail in its round trip through the VAE encoder and decoder, the output then takes $g(I)$ on $\Omega_u$ directly, a setting we call \emph{paste-back}.

\panofigureanchor
\subsection{A camera response curve from the generator}
\label{sec:crf}

\panofigureanchor
The anchor latent $z_a$ encodes the input as radiance, which requires the inverse camera response curve $g$ of \cref{eq:formation}, yet a camera's curve is unknown from its image. We measured the default curves of 18 recent cameras from their RAW and JPEG pairs~\cite{imagingresource}. They lie 0.4 to 0.5 stops from sRGB. At dark pixel values, their radiance is more than one stop above that of sRGB. An assumed curve matters because \cref{eq:anchor_update} holds every anchored cell to it.

\panofigureanchor
We extract the curve from the generator (\cref{fig:pipeline}). Because training exposes the model to many camera response curves, its scene estimate is a linear radiance image whose relation to the input pixel values approximates $g$. We run the sampler without anchoring for two steps from $\varepsilon_0$, form the one-step clean latent $\hat{z}_0 = z_1 - \sigma_1 v_\theta$ at the second step~\cite{esser2024sd3}, and decode $\hat{x}_0 = \mathcal{T}^{-1}(\mathcal{D}(\hat{z}_0))$. Over the unclipped pixels we take the median of $\hat{x}_0$ in each bin of four pixel values and make the medians monotone. Because the generator lifts the last pixel values before the clip point in some scenes, we extend the curve as a straight line in log radiance above the last bin that it estimates consistently (pixel value 240 to 250).

\panofigureanchor
This gives $g$ up to the exposure of the generator's estimate, which need not place the clip point at 1. We keep that exposure instead of pinning $g(255) = 1$, so that the anchored cells share the exposure frame of the cells the generator fills. The curve $g$ linearises $z_a$ only, whereas the condition latent $z_c$, the clipped-region mask and the VLM prompt keep the raw input $I$. The pre-run costs two transformer evaluations and one decode, and the main run starts from the same $\varepsilon_0$. Because this procedure needs no extra weights and assumes no curve, it also serves on its own as a single-image response estimator for an 8-bit image.

\panofigureanchor
\subsection{Implementation}
\label{sec:regimes}
\label{sec:impl}

\panofigureanchor
The generator \looseness=-1\relax is Qwen-Image-Edit-2511~\cite{wu2025qwenimage,qwen2025imageedit2511} with one rank-64 LoRA module on the transformer~\cite{hu2021lora}, while the text encoder, the VAE encoder and the base transformer weights stay frozen. The prompt concatenates a fixed task sentence, a panorama indicator sentence when applicable, and an image description written by Qwen3.8-27B-FP8~\cite{qwen2026qwen38} from the 8-bit input alone without the words ``HDR'', ``exposure'' or ``dynamic range''. Sampling uses $T = 10$ Euler steps without classifier-free guidance~\cite{ho2021cfg} after the two unanchored steps of \cref{sec:crf}, totalling twelve transformer evaluations per image. Perspective images are generated on a canvas of the input's size and panoramas at $1024 \times 2048$ in one pass. Above 1.5 megapixels we encode $z_c$ on the output canvas instead of the base model's one-megapixel grid. Both regimes share the same LoRA and VAE decoder, differing only in the panorama prompt sentence, the horizontal wrap of $m$ and the canvas. \Cref{sec:protocol} gives the training data and schedule.

\panofigureanchor   %
{}
\panofigureanchor
\section{Experiments}
\label{sec:experiments}

\panofigureanchor   %

\subsection{Setup}
\label{sec:protocol}

\panofigureanchor
\rpar{Training data} are 86,580 LDR--HDR pairs rendered from HDR sources. The sources are 4,170 panoramas (the Laval indoor and outdoor sets~\cite{gardner2017indoor,holdgeoffroy2019outdoor}, Poly Haven~\cite{polyhaven2026hdris} and UrbanSky~\cite{klotz2025urbansky}), perspective views cut out of them, bracketed perspective images~\cite{fairchild2007hdrps,empa_hdr_database} and our own RAW captures. The LDR side is rendered with an automatic exposure, a camera response curve drawn from measured~\cite{grossberg2003space} and parametric families, and measured sensor noise~\cite{darktable_noiseprofiles}. The VAE decoder is fine-tuned separately on the same data (\cref{sec:measurement}). We then train the rank-64 LoRA~\cite{hu2021lora} of \cref{sec:impl} for 10,000 steps on $8\times$ NVIDIA B200 GPUs with a batch of 4 per GPU. It then receives 1,000 updates of DRaFT~\cite{clark2024draft} with a batch of 2 per GPU and a reward on the radiance it generates in the clipped region. Every evaluation scene below is held out from every training stage.

\panofigureanchor
\rpar{Our benchmarks} are 117 outdoor Poly Haven scenes released after our training set was downloaded, 150 indoor panoramas of the Laval set~\cite{gardner2017indoor} held out from training, and the 181 scenes of \sihdr~\cite{hanji2022sihdr} at $1888\times1280$ with the released clip-95 input. Every panorama method receives the same 8-bit input at $1024\times2048$: the ground truth exposed by a camera meter and rendered through the measured response curve of a Canon EOS R5 with its sensor noise at ISO 100. We also demonstrate \method qualitatively on 46 bracketed real-world scenes we captured.

\panofigureanchor
\rpar{Metrics} align each reconstruction to the ground truth by a single scalar, the median luminance ratio over the unclipped input pixels, with the input's white point at 203\,cd/m$^2$~\cite{itu2023bt2408}. To score a panorama by the light it casts, we path-trace diffuse, matte silver and mirror spheres under the reconstruction and the ground truth~\cite{phongthawee2024diffusionlight,liang2025luxdit}. We compare the renders by scale-invariant root-mean-square error (RMSE)~\cite{grosse2009intrinsic} and mean RGB angular error. We also report the sun direction error on daytime outdoor scenes and the diffuse-sphere energy error in stops. For perspective images, we report PU21-PSNR on PU21-encoded~\cite{mantiuk2021pu21} RGB, PU21-SSIM~\cite{wang2004ssim} on PU21-encoded luminance, VSI~\cite{zhang2014vsi}, HDR-VDP-3~\cite{mantiuk2023hdrvdp3} at 60 pixels per degree, PSNR-H~\cite{dille2024intrinsichdr,dille2026rdre} (PU21-PSNR over input pixels above 0.8), patch FID~\cite{heusel2017fid,chai2022patchfid} following LEDiff~\cite{wang2025lediff}, and PIQE~\cite{venkatanath2015piqe} on the PU21-encoded channels.

\panofigureanchor
\rpar{Our highlight radiance error} scores the clipped region alone, since full-image metrics are dominated by the pixels the input already measured and a quality metric restricted to the saturated region separates methods poorly~\cite{hanji2022sihdr,eilertsen2021cheat}. It is computed on the region $\Omega$ of pixels with any channel at or above 0.95 of the input's white, and the prediction is not clipped. Each of its three terms takes a set $\mathcal{S}$ of luminance statistics and averages the absolute log ratio, in stops, between the ground-truth value $s$ and the predicted value $\hat s$:
\begin{equation}
E_{\mathcal{S}} = \frac{1}{|\mathcal{S}|} \sum_{s \in \mathcal{S}} \Bigl|\log_2 \frac{\hat s}{s}\Bigr| .
\label{eq:hre}
\end{equation}
The \emph{highlight energy error} HEE uses the total radiance of the region. The \emph{highlight quantile error} HQE uses its 50th, 90th and 99th percentiles, so it separates a flat plateau from a peaked light source of the same energy. The \emph{highlight spatial error} HSE uses the means over the $8 \times 8$ pixel cells at least half covered by $\Omega$, so it scores where the radiance is placed. For panoramas the energy a diffuse sphere receives plays the same role, since the clipped region of a panorama is its light source.

\panofigureanchor
\rpar{Baselines} comprise nine methods from \cref{sec:related}: the feed-forward methods HDR-CNN~\cite{eilertsen2017hdrcnn}, ExpandNet~\cite{marnerides2018expandnet}, MaskHDR~\cite{santos2020maskhdr}, SingleHDR~\cite{liu2020singlehdr}, IntrinsicHDR~\cite{dille2024intrinsichdr} and RDRE~\cite{dille2026rdre}, and the diffusion methods GaSLight~\cite{bolduc2025gaslight}, LEDiff~\cite{wang2025lediff} and DiffHDR~\cite{yu2026diffhdr} (whose panorama model is DiffHDR-Pano~\cite{yu2026promoting}). On \sihdr we score the benchmark's released reconstructions for the first four and the authors' results for GaSLight. We run the remaining methods and every panorama baseline with their official code and weights (LEDiff via its official $512\times512$ pipeline). \Cref{tab:pano} reports all but ExpandNet and MaskHDR. \method runs with the settings of \cref{sec:impl}, one sample per scene with a fixed seed, using the same weights across every table.

\panofigurelast   %
\subsection{Perspective images}
\label{sec:sihdr}
\Cref{tab:sihdr} compares \method with nine methods on \sihdr. \method achieves the best mean on every full-reference metric and leads each baseline on each of them in paired comparison over the 181 scenes. Against the closest baseline, RDRE, \method gains 4.5\,dB in PU21-PSNR and 2.7\,dB on the highlights scored by PSNR-H. On FID \method scores 8.75 against 14.50 for DiffHDR, the best baseline. The sole exception is the no-reference PIQE, on which \method and the ground truth itself obtain the two worst values in the table.

On the clipped region, \method achieves the lowest energy, quantile and spatial error. GaSLight, which merges generated exposures of the input, ranks second in energy error. \method lowers HQE to 0.49 stops against 0.82 for the best baseline, showing that the generated highlights match the spread of radiance as well as its total. \Cref{fig:qual} compares crops stopped down to reveal the clipped light source. On the two camera JPEGs, \method recovers a compact sun close to that of the ground truth, where the baselines leave a dim glow, a small point or a coloured blob. On the candles of the \sihdr scene, \method restores three bright flames, whereas the baselines leave the same flames dim, grey or tinted.
\subsection{Panoramas}
\label{sec:pano}
\Cref{tab:pano} scores the reconstructed panoramas by the light they cast. On both benchmarks \method achieves the highest panorama PU21-PSNR and the lowest scale-invariant RMSE and angular error on all three spheres. Its sun direction is off by 1.8$^\circ$ on average, against 5.7$^\circ$ for the closest baseline, DiffHDR-Pano. In \cref{fig:pano_spheres} the diffuse sphere under \method is lit as under the ground truth, and the matte silver sphere carries its highlight.

The last column reports the energy a diffuse sphere receives. Outdoors, \method is off by 0.40 stops on average and exceeds one stop in 10 of the 117 scenes. All baselines exceed one stop in at least 40 scenes, because their generated suns are too dim. Indoors, windows and lamps are clipped less severely, and \method is off by 0.28 stops. The closest baselines (RDRE, IntrinsicHDR and GaSLight) are off by 0.45 to 0.47 stops. Noise-adaptive anchoring keeps the unclipped region of the panorama from drifting while the generator supplies light sources with the right energy, as an environment map requires.
\subsection{Camera response}
\label{sec:crf_eval}

\begin{table}[t]
\centering
\footnotesize
\setlength{\tabcolsep}{3pt}
\caption{\textbf{Camera response estimation} on 100 RAW scenes rendered through five camera and five film curves. Abs and shape error in stops (\cref{sec:crf_eval}), median over the scenes of a group. Results are colour-coded by \legendbox{bestcolor}{best} per column.}
\label{tab:crf}
\begin{tabular}{lcccccc}
\toprule
\multirow{2}{*}[-2.4pt]{\textbf{Estimate}} & \multicolumn{2}{c}{\textbf{Cameras}} & \multicolumn{2}{c}{\textbf{Films}} & \multicolumn{2}{c}{\textbf{All}} \\
\cmidrule(lr){2-3} \cmidrule(lr){4-5} \cmidrule(lr){6-7}
 & Abs $\downarrow$ & Shape $\downarrow$ & Abs $\downarrow$ & Shape $\downarrow$ & Abs $\downarrow$ & Shape $\downarrow$ \\
\midrule
sRGB & 0.63 & 0.39 & 0.74 & 0.38 & 0.69 & 0.39 \\
CRF-net~\cite{li2017crfnet} & 0.43 & 0.23 & 0.31 & 0.18 & 0.38 & 0.20 \\
SingleHDR~\cite{liu2020singlehdr} & 0.35 & 0.16 & 0.80 & 0.29 & 0.56 & 0.22 \\
\midrule
\textbf{\method} (ours) & \cellcolor{bestcolor}0.29 & \cellcolor{bestcolor}0.09 & \cellcolor{bestcolor}0.23 & \cellcolor{bestcolor}0.10 & \cellcolor{bestcolor}0.25 & \cellcolor{bestcolor}0.10 \\
\bottomrule
\end{tabular}
\end{table}
{}
\begin{figure}[t]
  \centering
  \includegraphics[width=\linewidth, trim=0 4.00bp 0 2.00bp, clip]{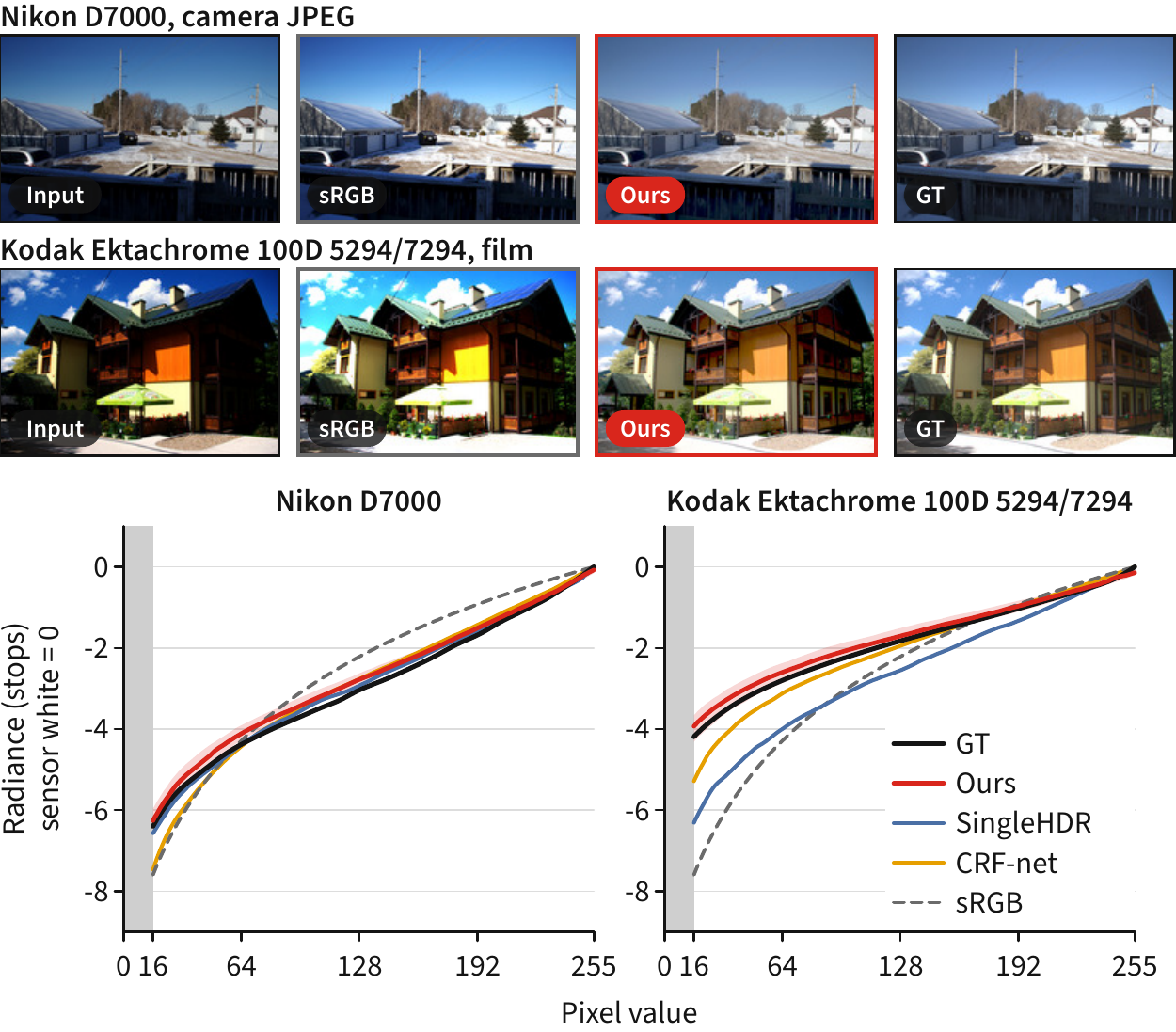}  %
  \caption{\textbf{Camera and film response estimation.} Top: the 8-bit input, the input linearised with the sRGB assumption and with our curve, and the ground-truth (GT) radiance at one exposure. Bottom: median estimated curves over 100 scenes, with the interquartile band for ours and pixel values below 16 masked in grey.}
  \label{fig:crf_curves}
\end{figure}

We evaluate the estimator of \cref{sec:crf} on 100 RAW scenes~\cite{rawpixls} rendered through ten response curves: five camera curves measured from RAW and JPEG pairs~\cite{imagingresource} and five film curves (two colour negatives, two reversal films and one black-and-white film) read from manufacturer data sheets~\cite{kodak2025portra400,kodak2015vision3,fujifilm_velvia50,kodak2018ektachrome100d,harman2018hp5}.
We score each estimate by its mean absolute difference to the ground-truth curve, in stops, over pixel values 16 to 240. We report it directly (\emph{abs}, which includes the estimate's exposure) and after subtracting the median difference (\emph{shape}). \Cref{tab:crf} reports group medians, and \cref{fig:crf_curves} shows the curves and linearised images for one camera and one film.

Across the 1,000 images, our curve achieves a median error of 0.25 stops (0.10 stops in shape), compared with 0.69 (0.39) for the sRGB assumption and 0.38 (0.20) for CRF-net~\cite{li2017crfnet} with the weights of the SingleHDR release. SingleHDR's Linearization-Net~\cite{liu2020singlehdr} lowers the error on cameras to 0.35 stops yet reaches 0.80 stops on films, because its predicted curves span nearly the same range whatever the input. On cameras, our curve sits 0.2 to 0.4 stops too bright in the mid-tones.

\subsection{Ablations}
\label{sec:ablations}

\begin{figure}[t]
  \centering
  \includegraphics[width=\linewidth, trim=0 5.50bp 0 5.50bp, clip]{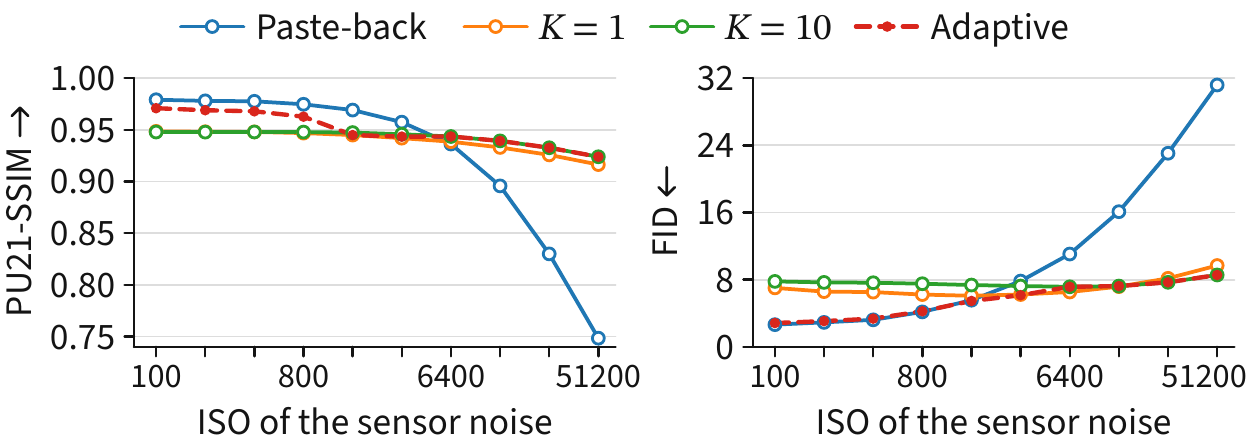}  %
  \caption{\textbf{Noise-adaptive anchoring} on \sihdr re-rendered with Canon EOS R5 noise. Paste-back is $K = 0$, $K = 10$ is no anchoring, and Adaptive sets $K$ from the sensor noise.}
  \label{fig:adaptive}
\end{figure}

\rpar{Noise-adaptive anchoring} is tested on inputs of known noise in \cref{fig:adaptive}: the ground truth of the 181 \sihdr scenes rendered through the response curve of the Canon EOS R5 with its sensor noise at ISO 100 to 51200. Paste-back is the best setting on clean inputs and the worst on noisy ones, because it copies the sensor noise into the output: its FID rises from 2.65 to 31.18. A fixed $K = 1$ or $K = 10$ lets the generator remove the noise and stays nearly level across the range, but it also revises clean inputs. Adaptive reads the noise from each input, so it follows paste-back at low ISO and coincides with $K = 10$ at high ISO. Averaged over the ten levels it reaches a PU21-SSIM of 0.950 and an FID of 5.58, against 0.942 and 7.02 for the best fixed setting on each metric.

\begin{table}[t]
\centering
\footnotesize
\setlength{\tabcolsep}{1pt}
\setlength{\aboverulesep}{1pt}
\setlength{\belowrulesep}{1pt}
\caption{\textbf{Anchor and response curve.} Anchored rows use noise-adaptive anchoring and the first row has none. GT is the true curve. Perspective: \sihdr; panorama: Poly Haven and Laval. Angle: diffuse sphere. Colours mark the \legendbox{bestcolor}{best} and \legendbox{secondcolor}{second best}.}
\label{tab:ablation}
\begin{tabular*}{\linewidth}{@{\extracolsep{\fill}}cccccccc@{}}
\toprule
\multirow{2}{*}[-1.15pt]{\textbf{Anchor}} & \multicolumn{3}{c}{\textbf{Curve}} & \multicolumn{2}{c}{\textbf{Perspective}} & \multicolumn{2}{c}{\textbf{Panorama}} \\
\cmidrule(lr){2-4} \cmidrule(lr){5-6} \cmidrule(lr){7-8}
 & sRGB & Ours & GT & PU21-PSNR $\uparrow$ & FID $\downarrow$ & PU21-PSNR $\uparrow$ & Angle $\downarrow$ \\
\midrule
-- &  &  &  & 27.88 & 9.05 & 29.65 & 2.17 \\
\checkmark & \checkmark &  &  & 27.68 & 8.78 & 23.75 & 2.74 \\
\checkmark &  & \checkmark &  & \cellcolor{secondcolor}27.93 & \cellcolor{secondcolor}8.75 & \cellcolor{secondcolor}32.57 & \cellcolor{secondcolor}2.00 \\
\checkmark &  &  & \checkmark & \cellcolor{bestcolor}28.05 & \cellcolor{bestcolor}8.72 & \cellcolor{bestcolor}35.90 & \cellcolor{bestcolor}1.78 \\
\bottomrule
\end{tabular*}
\end{table}
{}
\rpar{The anchor and its curve} are varied in \cref{tab:ablation} on the 181 \sihdr scenes and the 267 panoramas of Poly Haven and Laval, with noise-adaptive anchoring in the three anchored rows. On panoramas, anchoring with our curve is 2.9\,dB above no anchoring. Anchoring with the sRGB assumption is 5.9\,dB below no anchoring, since it holds the unclipped region to the wrong curve. The ground-truth curve adds a further 3.3\,dB, the room left for the estimate, and the angular error orders the rows the same way. On \sihdr the rows differ little because the benchmark adds strong sensor noise to its simulated inputs~\cite{hanji2022sihdr}: the rule anchors only 29\,\% of the scenes, against 98\,\% of the panoramas.
{}
\section{Conclusion and Limitations}
\label{sec:conclusion}
We introduce \method, a single model, built on a pretrained diffusion transformer, that reconstructs HDR radiance from one LDR perspective image or panorama. Noise-adaptive anchoring keeps the radiance the camera recorded and generates what it clipped. The camera response curve the anchor needs is extracted from the generator itself. With the same weights, \method leads nine previous methods on \sihdr in fidelity and highlight recovery. On panoramas scored by path-traced spheres, its reconstructions light a scene closest to the ground truth. It also estimates camera and film response curves more accurately than dedicated estimators. As with other large generative models, this comes at a cost: the 20-billion-parameter transformer takes about ten seconds per image on one NVIDIA B200 GPU.
{}
\section*{LLM Usage Statement}
We used Claude Code for language polishing and proofreading, to help draft and shorten parts of the text, to assist with implementing the data, training, evaluation and figure scripts and with running the experiments, and to search for related work. All AI-assisted material was reviewed and validated by the authors: the code was tested, each suggested reference was read before it was cited, and all text was checked and revised. The method, the experimental design and the conclusions are the work of the authors, who take full responsibility for this manuscript. Separately, a vision-language model, Qwen3.8-27B-FP8~\cite{qwen2026qwen38}, is a component of the method: it writes the description of the input image that is part of the prompt (\cref{sec:impl}).
{}   %

\twocolumn[{\centering\Large\bfseries Appendix\par\vspace{1.5em}}]   %
\appendix

\noindent This appendix adds to the main paper. \Cref{sec:supp_results} reports further results: more methods, metrics and reconstructions on perspective images and panoramas, the camera response estimate per curve, and ablations. \Cref{sec:supp_method} completes the method: the inverse encoding, the anchor weight, the sensor-noise estimate, paste-back, the extraction of the camera response curve and the sampler. \Cref{sec:supp_train} describes the training data and the three training stages. \Cref{sec:supp_eval} defines the test sets and every metric of the main paper.

\renewcommand{\dbltopfraction}{0.92}\renewcommand{\textfraction}{0.05}\renewcommand{\dblfloatpagefraction}{0.85}\setcounter{dbltopnumber}{2}\setcounter{totalnumber}{4}
\section{Additional results}
\label{sec:supp_results}

\subsection{Perspective images}
\label{sec:supp_more}
\begin{figure*}[p]
\centering
{  \centering
  \includegraphics[width=\linewidth]{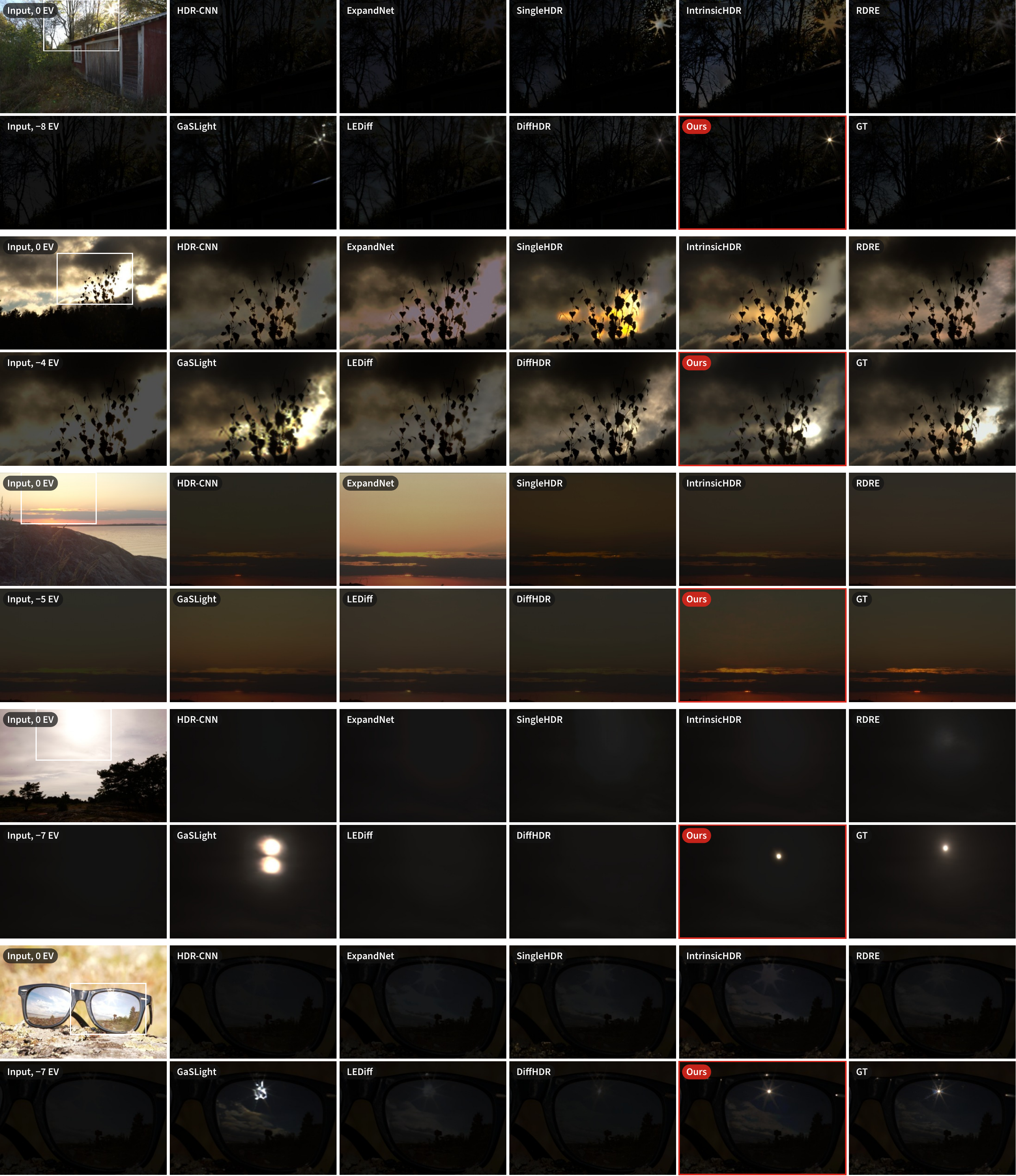}
  \captionof{figure}{\textbf{More results on perspective images.}
  Five \sihdr~\cite{hanji2022sihdr} scenes, two rows each. First tile: the input at $0$\,EV with a window marked. The other tiles show this window stopped down by the amount printed on the second row: the input, eight methods of \cref*{tab:sihdr}, \method and the ground truth (GT).
  Every reconstruction is scaled to the GT by the single scalar of the evaluation and shown as the sRGB encoding of $\mathrm{clip}(x\cdot2^{\mathrm{EV}})$, with $x=1$ at the input's clip point. The stopped-down input is its pixel values raised to the power 2.2.}
  \label{fig:supp_qual_persp}}
\end{figure*}

\begin{figure*}[p]
\centering
{  \centering
  \includegraphics[width=\linewidth]{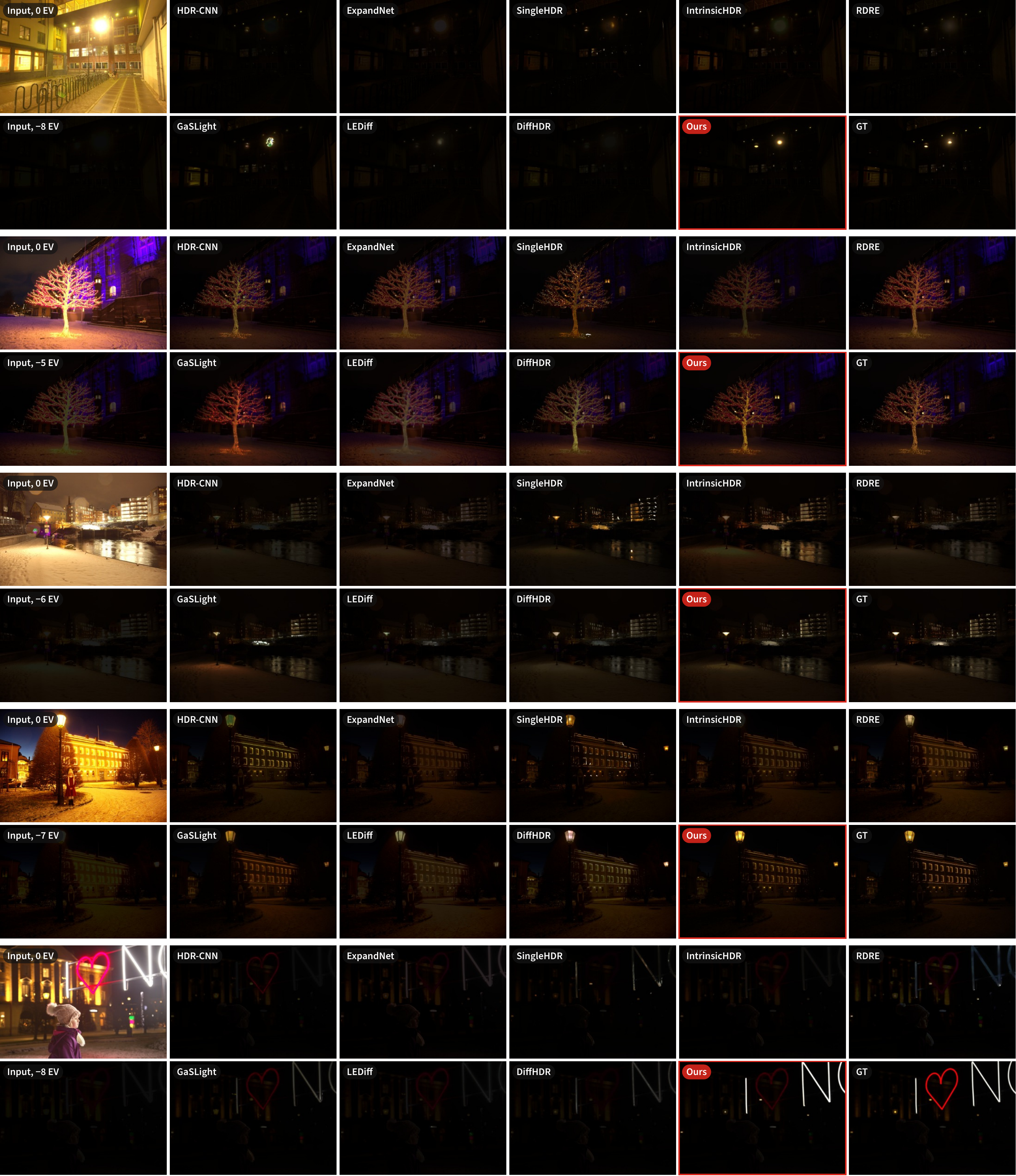}
  \captionof{figure}{\textbf{More results on perspective images (continued).} Five more \sihdr scenes, shown as in \cref{fig:supp_qual_persp}. Every tile shows the whole frame instead of a window.}
  \label{fig:supp_qual_persp_b}}
\end{figure*}

\rpar{More reconstructions.} \Cref{fig:supp_qual_persp,fig:supp_qual_persp_b} show ten further \sihdr scenes. The scenes cover different light sources: the sun, a clouded and a sunset sky, the sun reflected in sunglasses, street lamps and decorative lights at night.

\rpar{Further methods.} \Cref{tab:supp_persp} repeats \cref*{tab:sihdr} with six more methods: HDRUNet~\cite{chen2021hdrunet}, the two exposure-stack methods DrTMO~\cite{endo2017drtmo} and Deep Recursive HDRI~\cite{lee2018hdrgan}, GaSLight with its public weights, the video-diffusion method Vid2HDRImg~\cite{talegaonkar2026singleshot}, and X2HDR~\cite{wu2026x2hdr}.

\rpar{Further metrics.} \Cref{tab:supp_metrics} gives the FID of \cref*{tab:sihdr} and four more measurements for the methods of \cref{tab:supp_persp}. HDR-FLIP~\cite{andersson2021hdrflip} is computed on the aligned images. HDR-VDP-3 is given on its JOD scale, obtained per image from the score $Q$ of \cref*{tab:sihdr}. ColorVideoVDP~\cite{mantiuk2024cvvdp} is computed on the same images in cd/m$^2$, with a linear HDR display of 10{,}000\,cd/m$^2$ peak and 60 pixels per degree. FID and KID~\cite{binkowski2018kid} are computed as in \cref{sec:supp_metrics} with the tone-mapping operators of Reinhard~\cite{reinhard2002photographic}, Durand~\cite{durand2002bilateral} and Liang~\etal~\cite{liang2018hybrid}.

\rpar{Inference time.} The last column of \cref{tab:supp_metrics} is the time per image on one NVIDIA B200, without loading the model. \method takes 8.2 seconds on average for an $1888 \times 1280$ image. The 128 \sihdr images that run without anchoring take 7.4 seconds: they need no camera response curve, so its extraction is skipped. The other images take 10.2 seconds. The time excludes the image description.

\rpar{PIQE and sharpness.} In \cref*{tab:sihdr} the ground truth and \method have the two highest PIQE values. \Cref{tab:supp_piqe} filters the image that PIQE receives. Sharpening the ground truth lowers its PIQE from 41.45 to 31.75, and the score drops on 96\,\% of the scenes. A Gaussian blur of one pixel raises it to 71.35. PIQE therefore follows the local contrast of an image, and a filter that moves the ground truth away from itself improves the score.
\par\noindent\begin{minipage}{\linewidth}\begin{center}
\centering
\footnotesize
\setlength{\tabcolsep}{6pt}
\captionof{table}{\textbf{PIQE after filtering} (181 \sihdr scenes, mean{\scriptsize$\pm$std}, lower is better). The filter is applied to the PU21-encoded image $V \in [0, 1]$ that PIQE receives. Sharpen: $\mathrm{clip}(V + a\,(V - \mathcal{G}_1 * V),\, 0,\, 1)$. Blur: $\mathcal{G}_\sigma * V$, with $\mathcal{G}_\sigma$ a Gaussian of standard deviation $\sigma$.}
\label{tab:supp_piqe}
\begin{tabular}{lcc}
\toprule
\textbf{Filter} & \textbf{GT} & \textbf{\method} \\
\midrule
None & 41.45{\scriptsize$\pm$14.46} & 37.21{\scriptsize$\pm$15.49} \\
\midrule
Sharpen, $a = 1$ & 31.75{\scriptsize$\pm$12.29} & 24.25{\scriptsize$\pm$10.76} \\
\midrule
Blur, $\sigma = 1$\,px & 71.35{\scriptsize$\pm$9.73} & 74.92{\scriptsize$\pm$8.43} \\
Blur, $\sigma = 2$\,px & 88.74{\scriptsize$\pm$7.72} & 88.70{\scriptsize$\pm$8.83} \\
\bottomrule
\end{tabular}
\end{center}
{}\end{minipage}\par

\begin{figure*}[p]
\centering
{  \centering
  \includegraphics[width=\linewidth]{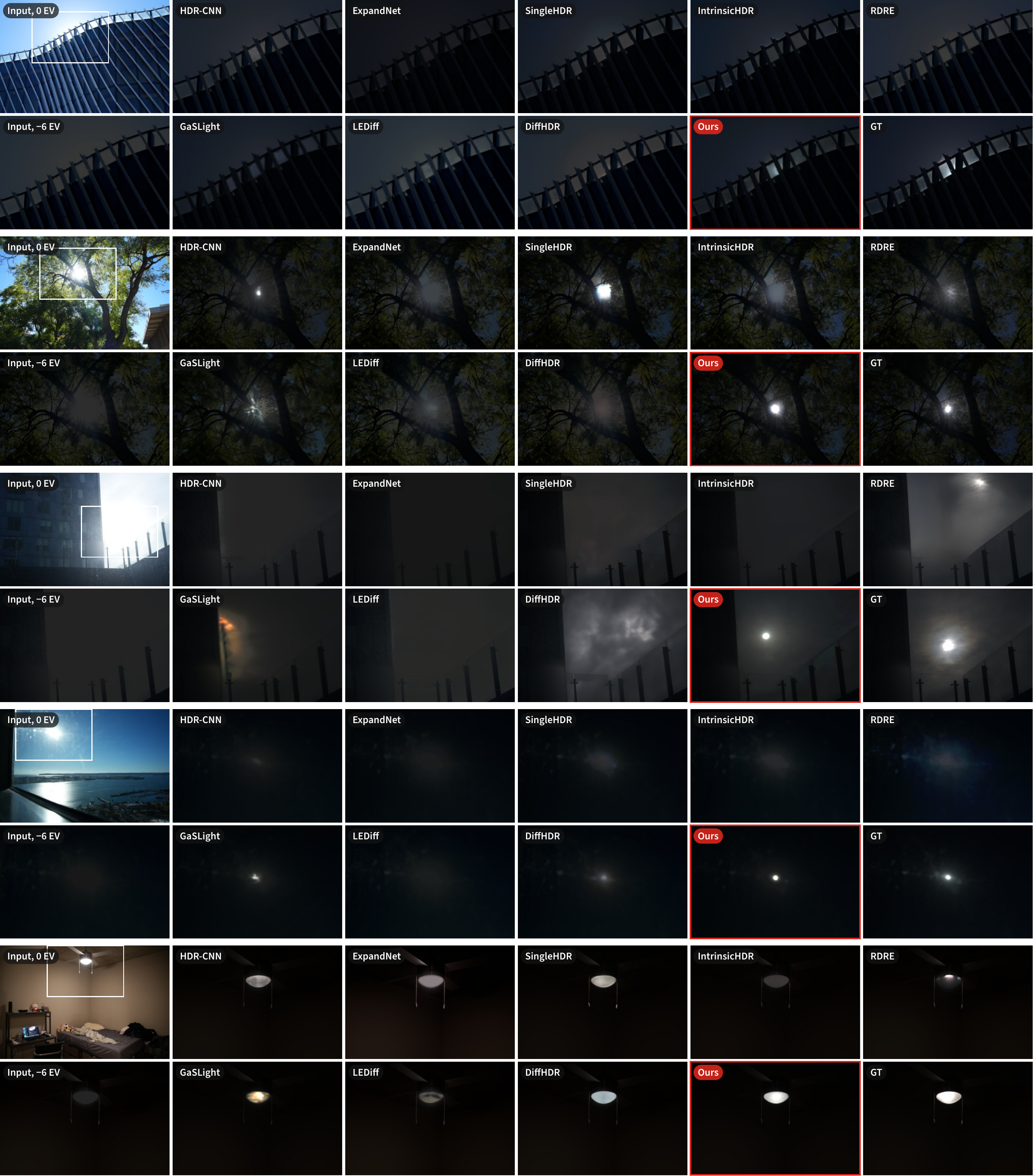}
  \captionof{figure}{\textbf{Results on scenes we photographed.}
  Five of the 46 scenes we photographed, shown as in \cref{fig:supp_qual_persp}. First tile: the camera JPEG at $0$\,EV with a window around the brightest light source. The other tiles show this window stopped down: the input, eight methods of \cref*{tab:sihdr}, \method and the ground truth (GT), which is merged from the exposure bracket. People are blurred in every column.}
  \label{fig:supp_qual_capture}}
\end{figure*}

\rpar{Scenes we photographed.} \Cref{tab:supp_capture} compares the methods of \cref*{tab:sihdr} on the 46 scenes we photographed, with camera JPEGs as input (\cref{sec:supp_sets}). \Cref{fig:supp_qual_capture} shows five of the scenes. \method has the highest PU21-PSNR, HDR-VDP-3 and PSNR-H and the lowest HDR-FLIP and HQE. SingleHDR has the highest PU21-SSIM (0.949 against 0.948) and the lowest HSE, and GaSLight has the lowest HEE (0.65 against 0.69 stops).


\begin{figure*}[p]
\centering
{\centering
\footnotesize
\setlength{\tabcolsep}{2.3pt}
\captionof{table}{\textbf{HDR reconstruction on \sihdr with six more methods} (marked $^\ast$; 181 scenes), with the metrics and conventions of \cref*{tab:sihdr}. Its FID column is in \cref{tab:supp_metrics}, with two more tone mappers. HDRUNet, Vid2HDRImg and X2HDR are run with their official code at the input size. DrTMO is run with its official networks and weights, and its exposure stack is merged with the camera response recovered by the method of Debevec and Malik~\cite{debevec1997recovering}, as its paper describes. Deep Recursive HDRI is run with its official code and weights, which work at $256 \times 256$. Its exposure stack is merged with the HDR Toolbox~\cite{banterle2017advanced} implementation of the method of Debevec and Malik, as its authors did, and the result is upsampled to the input size. GaSLight, public, uses the released weights, whereas the GaSLight row of \cref*{tab:sihdr} scores the reconstructions released by the authors. Results are colour-coded by \legendbox{bestcolor}{best} and \legendbox{secondcolor}{second best} per column over all methods.}
\label{tab:supp_persp}
\begin{tabular}{lccccccccc}
\toprule
\multirow{2}{*}[-2.4pt]{\textbf{Method}} & \multicolumn{5}{c}{\textbf{Full image}} & \multicolumn{4}{c}{\textbf{Clipped region}} \\
\cmidrule(lr){2-6} \cmidrule(lr){7-10}
 & PU21-PSNR $\uparrow$ & PU21-SSIM $\uparrow$ & VSI $\uparrow$ & HDR-VDP-3 $\uparrow$ & PIQE $\downarrow$ & PSNR-H $\uparrow$ & HEE $\downarrow$ & HQE $\downarrow$ & HSE $\downarrow$ \\
\midrule
HDR-CNN~\cite{eilertsen2017hdrcnn} & 21.43{\scriptsize$\pm$3.47} & 0.811{\scriptsize$\pm$0.136} & 0.977{\scriptsize$\pm$0.017} & 7.37{\scriptsize$\pm$0.80} & 21.68{\scriptsize$\pm$9.19} & 18.91{\scriptsize$\pm$4.93} & 1.26{\scriptsize$\pm$1.32} & 1.20{\scriptsize$\pm$0.99} & 0.64{\scriptsize$\pm$0.45} \\
ExpandNet~\cite{marnerides2018expandnet} & 21.11{\scriptsize$\pm$4.28} & 0.855{\scriptsize$\pm$0.067} & 0.973{\scriptsize$\pm$0.020} & 7.18{\scriptsize$\pm$0.96} & \cellcolor{secondcolor}15.85{\scriptsize$\pm$5.56} & 17.86{\scriptsize$\pm$5.74} & 1.10{\scriptsize$\pm$1.13} & 1.17{\scriptsize$\pm$0.88} & 0.80{\scriptsize$\pm$0.66} \\
MaskHDR~\cite{santos2020maskhdr} & 21.56{\scriptsize$\pm$3.42} & 0.811{\scriptsize$\pm$0.136} & 0.978{\scriptsize$\pm$0.016} & 7.41{\scriptsize$\pm$0.82} & 22.34{\scriptsize$\pm$9.66} & 19.12{\scriptsize$\pm$4.77} & 1.26{\scriptsize$\pm$1.33} & 1.18{\scriptsize$\pm$0.97} & 0.63{\scriptsize$\pm$0.46} \\
SingleHDR~\cite{liu2020singlehdr} & 21.06{\scriptsize$\pm$3.13} & \cellcolor{secondcolor}0.887{\scriptsize$\pm$0.065} & 0.979{\scriptsize$\pm$0.013} & 7.62{\scriptsize$\pm$0.68} & 21.75{\scriptsize$\pm$9.36} & 16.38{\scriptsize$\pm$4.45} & 0.93{\scriptsize$\pm$0.99} & 0.85{\scriptsize$\pm$0.68} & 0.68{\scriptsize$\pm$0.45} \\
IntrinsicHDR~\cite{dille2024intrinsichdr} & 23.24{\scriptsize$\pm$3.52} & 0.885{\scriptsize$\pm$0.063} & 0.982{\scriptsize$\pm$0.014} & 7.70{\scriptsize$\pm$0.81} & 21.70{\scriptsize$\pm$7.67} & 19.29{\scriptsize$\pm$4.97} & 1.04{\scriptsize$\pm$1.15} & 0.97{\scriptsize$\pm$0.83} & 0.62{\scriptsize$\pm$0.45} \\
RDRE~\cite{dille2026rdre} & \cellcolor{secondcolor}23.43{\scriptsize$\pm$3.88} & 0.841{\scriptsize$\pm$0.102} & \cellcolor{secondcolor}0.985{\scriptsize$\pm$0.013} & \cellcolor{secondcolor}7.74{\scriptsize$\pm$0.86} & 19.76{\scriptsize$\pm$9.02} & \cellcolor{secondcolor}20.89{\scriptsize$\pm$5.61} & 0.97{\scriptsize$\pm$1.12} & 0.85{\scriptsize$\pm$0.79} & \cellcolor{secondcolor}0.55{\scriptsize$\pm$0.43} \\
\midrule
GaSLight~\cite{bolduc2025gaslight} & 22.41{\scriptsize$\pm$2.99} & 0.851{\scriptsize$\pm$0.096} & 0.972{\scriptsize$\pm$0.014} & 7.24{\scriptsize$\pm$0.67} & 26.71{\scriptsize$\pm$11.49} & 17.68{\scriptsize$\pm$4.04} & \cellcolor{secondcolor}0.83{\scriptsize$\pm$0.79} & \cellcolor{secondcolor}0.82{\scriptsize$\pm$0.60} & 0.65{\scriptsize$\pm$0.38} \\
LEDiff~\cite{wang2025lediff} & 18.77{\scriptsize$\pm$3.60} & 0.753{\scriptsize$\pm$0.145} & 0.968{\scriptsize$\pm$0.018} & 6.92{\scriptsize$\pm$0.78} & 23.11{\scriptsize$\pm$8.81} & 16.27{\scriptsize$\pm$5.32} & 1.27{\scriptsize$\pm$1.31} & 1.16{\scriptsize$\pm$0.96} & 0.75{\scriptsize$\pm$0.57} \\
DiffHDR~\cite{yu2026diffhdr} & 22.08{\scriptsize$\pm$3.45} & 0.843{\scriptsize$\pm$0.079} & 0.978{\scriptsize$\pm$0.015} & 7.28{\scriptsize$\pm$0.71} & \cellcolor{bestcolor}15.25{\scriptsize$\pm$6.01} & 18.23{\scriptsize$\pm$4.57} & 1.09{\scriptsize$\pm$1.12} & 0.94{\scriptsize$\pm$0.75} & 0.62{\scriptsize$\pm$0.44} \\
\midrule
HDRUNet~\cite{chen2021hdrunet}$^\ast$ & 20.75{\scriptsize$\pm$4.20} & 0.878{\scriptsize$\pm$0.143} & 0.978{\scriptsize$\pm$0.016} & 7.47{\scriptsize$\pm$1.04} & 59.17{\scriptsize$\pm$12.86} & 17.81{\scriptsize$\pm$5.52} & 1.17{\scriptsize$\pm$1.22} & 1.05{\scriptsize$\pm$0.86} & 0.75{\scriptsize$\pm$0.60} \\
DrTMO~\cite{endo2017drtmo}$^\ast$ & 20.91{\scriptsize$\pm$4.13} & 0.874{\scriptsize$\pm$0.076} & 0.975{\scriptsize$\pm$0.016} & 7.28{\scriptsize$\pm$0.93} & 19.13{\scriptsize$\pm$7.58} & 15.69{\scriptsize$\pm$5.82} & 1.89{\scriptsize$\pm$1.45} & 1.87{\scriptsize$\pm$1.23} & 1.14{\scriptsize$\pm$0.81} \\
Deep Recursive HDRI~\cite{lee2018hdrgan}$^\ast$ & 19.56{\scriptsize$\pm$3.41} & 0.762{\scriptsize$\pm$0.138} & 0.960{\scriptsize$\pm$0.019} & 6.18{\scriptsize$\pm$0.88} & 91.49{\scriptsize$\pm$4.51} & 15.80{\scriptsize$\pm$4.93} & 1.59{\scriptsize$\pm$1.72} & 1.41{\scriptsize$\pm$1.48} & 0.95{\scriptsize$\pm$1.15} \\
GaSLight~\cite{bolduc2025gaslight}, public$^\ast$ & 21.36{\scriptsize$\pm$3.29} & 0.834{\scriptsize$\pm$0.113} & 0.973{\scriptsize$\pm$0.012} & 7.19{\scriptsize$\pm$0.68} & 24.01{\scriptsize$\pm$11.38} & 17.88{\scriptsize$\pm$4.72} & 0.93{\scriptsize$\pm$0.85} & 0.83{\scriptsize$\pm$0.63} & 0.63{\scriptsize$\pm$0.44} \\
Vid2HDRImg~\cite{talegaonkar2026singleshot}$^\ast$ & 22.39{\scriptsize$\pm$3.17} & 0.845{\scriptsize$\pm$0.103} & 0.973{\scriptsize$\pm$0.016} & 7.17{\scriptsize$\pm$0.84} & 35.04{\scriptsize$\pm$14.06} & 18.25{\scriptsize$\pm$4.36} & 1.34{\scriptsize$\pm$1.27} & 1.32{\scriptsize$\pm$0.90} & 0.64{\scriptsize$\pm$0.40} \\
X2HDR~\cite{wu2026x2hdr}$^\ast$ & 15.42{\scriptsize$\pm$3.61} & 0.745{\scriptsize$\pm$0.108} & 0.942{\scriptsize$\pm$0.021} & 6.16{\scriptsize$\pm$0.67} & 40.54{\scriptsize$\pm$12.88} & 11.61{\scriptsize$\pm$5.17} & 1.53{\scriptsize$\pm$1.15} & 1.77{\scriptsize$\pm$1.08} & 1.53{\scriptsize$\pm$1.03} \\
\midrule
\textbf{\method} (ours) & \cellcolor{bestcolor}27.93{\scriptsize$\pm$3.25} & \cellcolor{bestcolor}0.934{\scriptsize$\pm$0.036} & \cellcolor{bestcolor}0.991{\scriptsize$\pm$0.008} & \cellcolor{bestcolor}8.35{\scriptsize$\pm$0.64} & 37.21{\scriptsize$\pm$15.49} & \cellcolor{bestcolor}23.58{\scriptsize$\pm$4.74} & \cellcolor{bestcolor}0.54{\scriptsize$\pm$0.59} & \cellcolor{bestcolor}0.49{\scriptsize$\pm$0.41} & \cellcolor{bestcolor}0.33{\scriptsize$\pm$0.26} \\
\midrule
GT & $\infty$ & 1.000 & 1.000 & 10.00 & 41.45 & $\infty$ & 0.00 & 0.00 & 0.00 \\
\bottomrule
\end{tabular}
}
{}
\vspace{10pt}
{\centering
\footnotesize
\setlength{\tabcolsep}{3pt}
\captionof{table}{\textbf{Further metrics on \sihdr} (181 scenes) for the methods of \cref{tab:supp_persp} ($^\ast$: not in \cref*{tab:sihdr}). HDR-FLIP~\cite{andersson2021hdrflip}, HDR-VDP-3~\cite{mantiuk2023hdrvdp3} in JOD units and ColorVideoVDP~\cite{mantiuk2024cvvdp} (JOD) are given as mean{\scriptsize$\pm$std} over scenes. FID and KID ($\times 10^3$) are computed on the crops of \cref*{tab:sihdr} after tone mapping with the operators of Reinhard~\etal~\cite{reinhard2002photographic}, Durand and Dorsey~\cite{durand2002bilateral}, and Liang~\etal~\cite{liang2018hybrid}. Results are colour-coded by \legendbox{bestcolor}{best} and \legendbox{secondcolor}{second best} per column. The last column is the time per image in seconds on one NVIDIA B200, without loading the model ($^\dagger$: the network processes a $512 \times 512$ image; --: not measured). GaSLight is timed with its public weights, IntrinsicHDR runs its linearisation on the CPU, and MaskHDR runs at $1920 \times 1280$.}
\label{tab:supp_metrics}
\begin{tabular}{lcccccccccr}
\toprule
\multirow{2}{*}[-2.4pt]{\textbf{Method}} & \multirow{2}{*}[-2.4pt]{HDR-FLIP $\downarrow$} & \multirow{2}{*}[-2.4pt]{HDR-VDP-3 (JOD) $\uparrow$} & \multirow{2}{*}[-2.4pt]{ColorVideoVDP $\uparrow$} & \multicolumn{2}{c}{\textbf{Reinhard}} & \multicolumn{2}{c}{\textbf{Durand}} & \multicolumn{2}{c}{\textbf{Liang}} & \multirow{2}{*}[-2.4pt]{Time (s)} \\
\cmidrule(lr){5-6} \cmidrule(lr){7-8} \cmidrule(lr){9-10}
 & & & & FID $\downarrow$ & KID $\downarrow$ & FID $\downarrow$ & KID $\downarrow$ & FID $\downarrow$ & KID $\downarrow$ & \\
\midrule
HDR-CNN~\cite{eilertsen2017hdrcnn} & 0.394{\scriptsize$\pm$0.176} & 8.18{\scriptsize$\pm$0.71} & 8.04{\scriptsize$\pm$0.87} & 18.19 & 7.71 & 24.33 & 12.63 & 24.87 & 9.91 & $<$\,1 \\
ExpandNet~\cite{marnerides2018expandnet} & 0.408{\scriptsize$\pm$0.146} & 8.00{\scriptsize$\pm$0.88} & 7.87{\scriptsize$\pm$0.88} & 17.70 & 7.22 & 30.30 & 16.53 & 25.05 & 9.94 & $<$\,1 \\
MaskHDR~\cite{santos2020maskhdr} & 0.396{\scriptsize$\pm$0.179} & 8.21{\scriptsize$\pm$0.72} & 7.96{\scriptsize$\pm$0.83} & 18.35 & 7.82 & 28.57 & 15.63 & 25.21 & 9.90 & $<$\,1 \\
SingleHDR~\cite{liu2020singlehdr} & 0.372{\scriptsize$\pm$0.115} & 8.40{\scriptsize$\pm$0.60} & 8.21{\scriptsize$\pm$0.84} & 17.13 & 6.22 & 18.67 & \cellcolor{secondcolor}6.47 & \cellcolor{secondcolor}17.80 & \cellcolor{secondcolor}6.13 & $<$\,1 \\
IntrinsicHDR~\cite{dille2024intrinsichdr} & 0.336{\scriptsize$\pm$0.138} & 8.46{\scriptsize$\pm$0.71} & 8.44{\scriptsize$\pm$0.71} & 15.59 & 5.33 & 26.55 & 12.71 & 21.99 & 7.15 & 1.1 \\
RDRE~\cite{dille2026rdre} & 0.337{\scriptsize$\pm$0.159} & \cellcolor{secondcolor}8.48{\scriptsize$\pm$0.75} & \cellcolor{secondcolor}8.54{\scriptsize$\pm$0.77} & 17.19 & 7.47 & 22.14 & 11.58 & 22.72 & 8.73 & $<$\,1 \\
\midrule
GaSLight~\cite{bolduc2025gaslight} & 0.347{\scriptsize$\pm$0.130} & 8.07{\scriptsize$\pm$0.60} & 7.41{\scriptsize$\pm$1.86} & 22.18 & 8.24 & 23.80 & 7.66 & 26.50 & 9.98 & 5.0$^\dagger$ \\
LEDiff~\cite{wang2025lediff} & 0.507{\scriptsize$\pm$0.194} & 7.78{\scriptsize$\pm$0.73} & 7.43{\scriptsize$\pm$0.85} & 18.33 & 6.36 & 37.18 & 21.04 & 27.86 & 10.60 & 4.3$^\dagger$ \\
DiffHDR~\cite{yu2026diffhdr} & 0.357{\scriptsize$\pm$0.150} & 8.10{\scriptsize$\pm$0.66} & 7.91{\scriptsize$\pm$0.85} & \cellcolor{secondcolor}14.50 & \cellcolor{secondcolor}4.03 & \cellcolor{secondcolor}17.83 & 6.89 & 22.63 & 8.66 & 146 \\
\midrule
HDRUNet~\cite{chen2021hdrunet}$^\ast$ & 0.456{\scriptsize$\pm$0.202} & 8.24{\scriptsize$\pm$0.94} & 7.76{\scriptsize$\pm$0.94} & 23.67 & 8.82 & 26.44 & 12.98 & 26.54 & 10.98 & $<$\,1 \\
DrTMO~\cite{endo2017drtmo}$^\ast$ & \cellcolor{secondcolor}0.322{\scriptsize$\pm$0.147} & 8.08{\scriptsize$\pm$0.84} & 7.53{\scriptsize$\pm$1.02} & 17.40 & 8.24 & 36.41 & 21.55 & 25.34 & 9.81 & -- \\
Deep Recursive HDRI~\cite{lee2018hdrgan}$^\ast$ & 0.468{\scriptsize$\pm$0.140} & 7.08{\scriptsize$\pm$0.89} & 6.68{\scriptsize$\pm$0.92} & 191.56 & 154.28 & 191.11 & 155.22 & 192.47 & 154.24 & -- \\
GaSLight~\cite{bolduc2025gaslight}, public$^\ast$ & 0.428{\scriptsize$\pm$0.159} & 8.03{\scriptsize$\pm$0.61} & 7.54{\scriptsize$\pm$0.86} & 24.30 & 9.17 & 28.76 & 11.55 & 29.49 & 11.32 & 5.0$^\dagger$ \\
Vid2HDRImg~\cite{talegaonkar2026singleshot}$^\ast$ & 0.351{\scriptsize$\pm$0.110} & 8.00{\scriptsize$\pm$0.76} & 7.74{\scriptsize$\pm$0.77} & 18.64 & 5.51 & 35.12 & 19.54 & 19.94 & 6.42 & 28.6 \\
X2HDR~\cite{wu2026x2hdr}$^\ast$ & 0.623{\scriptsize$\pm$0.163} & 7.07{\scriptsize$\pm$0.66} & 6.11{\scriptsize$\pm$1.54} & 32.09 & 13.82 & 47.42 & 29.60 & 36.73 & 15.77 & 18.6 \\
\midrule
\textbf{\method} (ours) & \cellcolor{bestcolor}0.195{\scriptsize$\pm$0.064} & \cellcolor{bestcolor}8.99{\scriptsize$\pm$0.52} & \cellcolor{bestcolor}8.82{\scriptsize$\pm$0.57} & \cellcolor{bestcolor}8.75 & \cellcolor{bestcolor}2.44 & \cellcolor{bestcolor}8.86 & \cellcolor{bestcolor}1.94 & \cellcolor{bestcolor}10.10 & \cellcolor{bestcolor}2.86 & 8.2 \\
\bottomrule
\end{tabular}
}
{}
\end{figure*}

\begin{figure*}[p]
\centering
{  \centering
  \includegraphics[width=\linewidth]{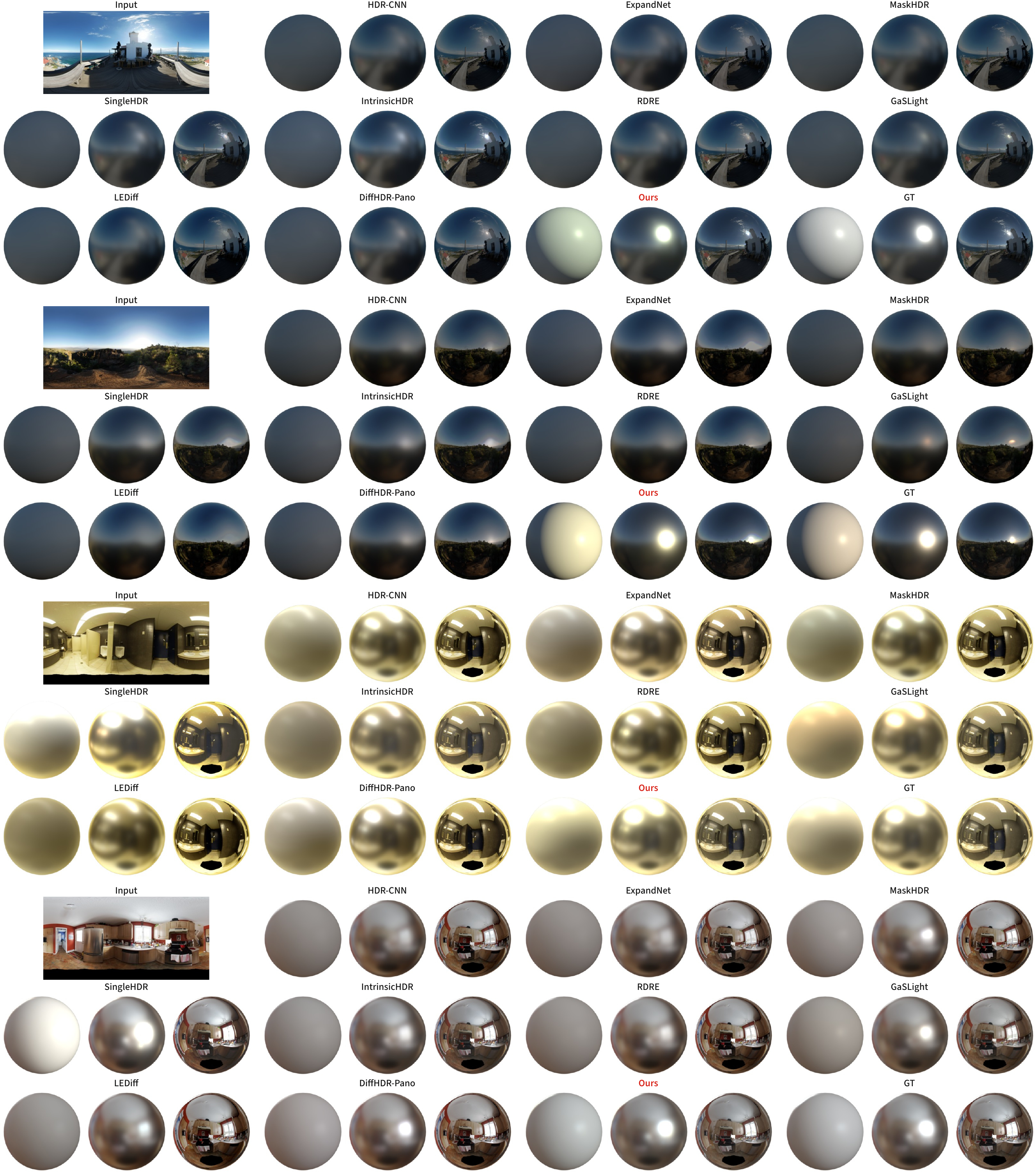}
  \captionof{figure}{\textbf{More panorama results.}
  Two outdoor Poly Haven~\cite{polyhaven2026hdris} scenes and two Laval~\cite{gardner2017indoor} interiors, three rows each. First cell: the 8-bit input panorama. The other cells show a diffuse, a matte silver and a mirror sphere path-traced under the panoramas of the methods of \cref{tab:supp_pano}, \method and the ground truth (GT). The spheres of a scene share one exposure.}
  \label{fig:supp_qual_pano}}
\end{figure*}
\begin{figure*}[p]
\centering
{  \centering
  \includegraphics[width=\linewidth]{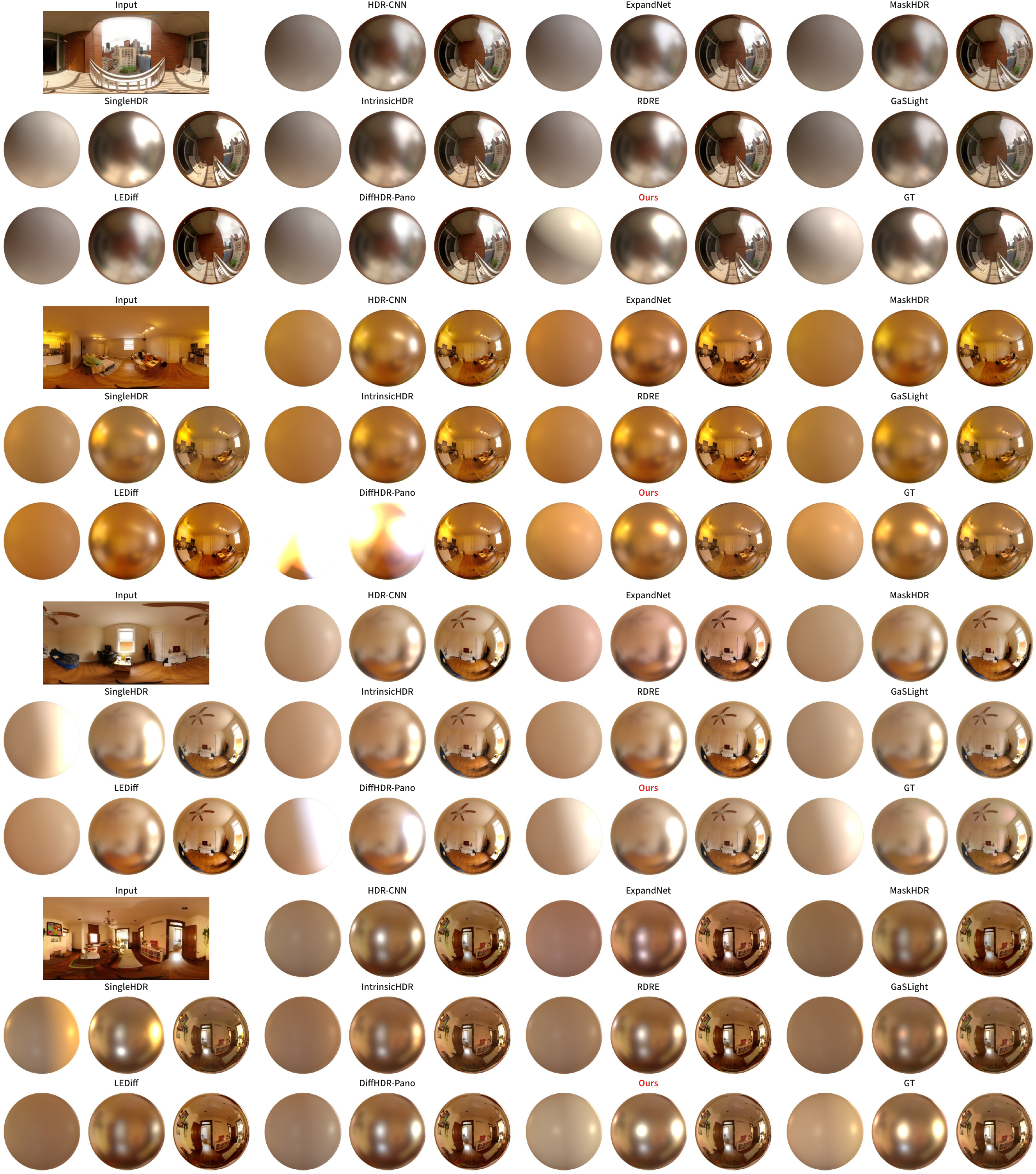}
  \captionof{figure}{\textbf{Panorama results on camera JPEGs.} Four Pano-Pano~\cite{ji2025digital} scenes, whose input is the JPEG written by the camera (\cref{tab:supp_pano}), shown as in \cref{fig:supp_qual_pano}.}
  \label{fig:supp_qual_p2p}}
\end{figure*}
\subsection{Panoramas}
\label{sec:supp_pano_more}
\rpar{More reconstructions.} \Cref{fig:supp_qual_pano} shows four further panoramas, two outdoor scenes lit by the sun and two interiors, as spheres path-traced under each reconstruction.

\Cref{tab:supp_pano} repeats \cref*{tab:pano} with ExpandNet and MaskHDR. Its third block tests real camera input: 177 panoramas of Pano-Pano~\cite{ji2025digital}, with the JPEG that the camera wrote as input. \method is the best on all eight columns. \Cref{fig:supp_qual_p2p} shows four of these scenes.

\begin{figure*}[t]
\centering
{\centering
\footnotesize
\setlength{\tabcolsep}{5pt}
\captionof{table}{\textbf{HDR reconstruction on the 46 scenes we photographed.} Camera JPEGs as input and a ground truth merged from the exposure bracket of each scene (\cref{sec:supp_sets}). Metrics and alignment as in \cref*{tab:sihdr}, mean{\scriptsize$\pm$std} over scenes. Pixels that are still clipped in the shortest exposure of the bracket are excluded. HEE and HQE are defined on 43 scenes and HSE on 41. GaSLight is run with its public weights. Results are colour-coded by \legendbox{bestcolor}{best} and \legendbox{secondcolor}{second best} per column.}
\label{tab:supp_capture}
\begin{tabular}{lcccccccc}
\toprule
\textbf{Method} & PU21-PSNR $\uparrow$ & PU21-SSIM $\uparrow$ & HDR-VDP-3 $\uparrow$ & HDR-FLIP $\downarrow$ & PSNR-H $\uparrow$ & HEE $\downarrow$ & HQE $\downarrow$ & HSE $\downarrow$ \\
\midrule
HDR-CNN~\cite{eilertsen2017hdrcnn} & 20.01{\scriptsize$\pm$1.62} & 0.858{\scriptsize$\pm$0.054} & 7.58{\scriptsize$\pm$0.40} & 0.437{\scriptsize$\pm$0.094} & 19.84{\scriptsize$\pm$5.59} & 0.89{\scriptsize$\pm$0.89} & 1.04{\scriptsize$\pm$0.77} & 0.72{\scriptsize$\pm$0.45} \\
ExpandNet~\cite{marnerides2018expandnet} & 21.91{\scriptsize$\pm$3.21} & 0.908{\scriptsize$\pm$0.053} & 7.61{\scriptsize$\pm$0.79} & 0.361{\scriptsize$\pm$0.089} & 18.97{\scriptsize$\pm$6.06} & 0.91{\scriptsize$\pm$0.91} & 1.28{\scriptsize$\pm$0.92} & 0.82{\scriptsize$\pm$0.59} \\
MaskHDR~\cite{santos2020maskhdr} & 19.82{\scriptsize$\pm$1.62} & 0.857{\scriptsize$\pm$0.054} & 7.54{\scriptsize$\pm$0.40} & 0.441{\scriptsize$\pm$0.097} & 19.26{\scriptsize$\pm$5.93} & 1.06{\scriptsize$\pm$1.12} & 1.35{\scriptsize$\pm$1.05} & 0.84{\scriptsize$\pm$0.58} \\
SingleHDR~\cite{liu2020singlehdr} & \cellcolor{secondcolor}26.06{\scriptsize$\pm$3.55} & \cellcolor{bestcolor}0.949{\scriptsize$\pm$0.029} & \cellcolor{secondcolor}8.53{\scriptsize$\pm$0.56} & \cellcolor{secondcolor}0.246{\scriptsize$\pm$0.085} & \cellcolor{secondcolor}21.66{\scriptsize$\pm$6.64} & 0.71{\scriptsize$\pm$0.72} & 0.82{\scriptsize$\pm$0.63} & \cellcolor{bestcolor}0.44{\scriptsize$\pm$0.34} \\
IntrinsicHDR~\cite{dille2024intrinsichdr} & 24.43{\scriptsize$\pm$4.48} & 0.941{\scriptsize$\pm$0.033} & 8.32{\scriptsize$\pm$0.61} & 0.292{\scriptsize$\pm$0.130} & 19.20{\scriptsize$\pm$5.70} & 0.87{\scriptsize$\pm$1.06} & 1.01{\scriptsize$\pm$1.01} & 0.70{\scriptsize$\pm$0.62} \\
RDRE~\cite{dille2026rdre} & 19.65{\scriptsize$\pm$3.28} & 0.875{\scriptsize$\pm$0.059} & 7.45{\scriptsize$\pm$0.97} & 0.425{\scriptsize$\pm$0.082} & 17.45{\scriptsize$\pm$6.79} & 0.97{\scriptsize$\pm$0.95} & 0.93{\scriptsize$\pm$0.72} & 1.00{\scriptsize$\pm$0.94} \\
\midrule
GaSLight~\cite{bolduc2025gaslight} & 23.36{\scriptsize$\pm$3.60} & 0.891{\scriptsize$\pm$0.050} & 7.78{\scriptsize$\pm$0.56} & 0.332{\scriptsize$\pm$0.142} & 19.38{\scriptsize$\pm$4.50} & \cellcolor{bestcolor}0.65{\scriptsize$\pm$0.62} & \cellcolor{secondcolor}0.78{\scriptsize$\pm$0.56} & 0.63{\scriptsize$\pm$0.36} \\
LEDiff~\cite{wang2025lediff} & 16.93{\scriptsize$\pm$1.94} & 0.797{\scriptsize$\pm$0.065} & 7.00{\scriptsize$\pm$0.39} & 0.556{\scriptsize$\pm$0.105} & 15.81{\scriptsize$\pm$5.31} & 1.08{\scriptsize$\pm$1.02} & 1.34{\scriptsize$\pm$0.91} & 0.98{\scriptsize$\pm$0.63} \\
DiffHDR~\cite{yu2026diffhdr} & 19.64{\scriptsize$\pm$1.98} & 0.833{\scriptsize$\pm$0.060} & 7.21{\scriptsize$\pm$0.34} & 0.440{\scriptsize$\pm$0.073} & 17.89{\scriptsize$\pm$4.79} & 0.87{\scriptsize$\pm$0.90} & 0.98{\scriptsize$\pm$0.75} & 0.80{\scriptsize$\pm$0.41} \\
\midrule
\textbf{\method} (ours) & \cellcolor{bestcolor}28.47{\scriptsize$\pm$2.88} & \cellcolor{secondcolor}0.948{\scriptsize$\pm$0.036} & \cellcolor{bestcolor}8.64{\scriptsize$\pm$0.52} & \cellcolor{bestcolor}0.193{\scriptsize$\pm$0.044} & \cellcolor{bestcolor}23.32{\scriptsize$\pm$5.68} & \cellcolor{secondcolor}0.69{\scriptsize$\pm$0.77} & \cellcolor{bestcolor}0.63{\scriptsize$\pm$0.45} & \cellcolor{secondcolor}0.53{\scriptsize$\pm$0.35} \\
\bottomrule
\end{tabular}
}
{}
\end{figure*}
\begin{figure*}[t]
\centering
{\centering
\footnotesize
\setlength{\tabcolsep}{1.6pt}
\captionof{table}{\textbf{Panorama reconstruction with two more methods}, ExpandNet and MaskHDR (marked $^\ast$), with the columns and conventions of \cref*{tab:pano} (117 Poly Haven and 150 Laval scenes). The third block runs the same methods on 177 panoramas of Pano-Pano, whose input is the JPEG written by the camera (\cref{sec:supp_sets}). Colours rank all rows of a block.}
\label{tab:supp_pano}
\begin{tabular}{clccccccccc}
\toprule
 & \multirow{2}{*}[-2.4pt]{\textbf{Method}} & \multirow{2}{*}[-2.4pt]{\textbf{PU21-PSNR} $\uparrow$} & \multicolumn{3}{c}{\textbf{Scale-invariant RMSE} $\downarrow$} & \multicolumn{3}{c}{\textbf{Angular error} ($^\circ$) $\downarrow$} & \multirow{2}{*}[-2.4pt]{\textbf{Sun} ($^\circ$) $\downarrow$} & \multirow{2}{*}[-2.4pt]{\textbf{Energy} (stops) $\downarrow$} \\
\cmidrule(lr){4-6} \cmidrule(lr){7-9}
 & & & Diffuse & Matte silver & Mirror & Diffuse & Matte silver & Mirror & & \\
\midrule
\multirow{10}{*}[-4.6pt]{\rotatebox[origin=c]{90}{\begin{tabular}{@{}c@{}}Poly Haven~\cite{polyhaven2026hdris}\\ 117 outdoor scenes\end{tabular}}} & HDR-CNN~\cite{eilertsen2017hdrcnn} & 23.73{\scriptsize$\pm$2.51} & 0.405{\scriptsize$\pm$0.670} & 1.605{\scriptsize$\pm$2.480} & 34.712{\scriptsize$\pm$51.598} & 5.24{\scriptsize$\pm$4.49} & 4.25{\scriptsize$\pm$1.73} & 4.34{\scriptsize$\pm$1.07} & 6.54{\scriptsize$\pm$18.38} & 1.04{\scriptsize$\pm$1.05} \\
 & SingleHDR~\cite{liu2020singlehdr} & \cellcolor{secondcolor}30.61{\scriptsize$\pm$3.39} & 0.339{\scriptsize$\pm$0.567} & 1.392{\scriptsize$\pm$2.222} & 34.425{\scriptsize$\pm$51.236} & \cellcolor{secondcolor}4.37{\scriptsize$\pm$3.85} & \cellcolor{secondcolor}3.02{\scriptsize$\pm$1.60} & \cellcolor{secondcolor}2.36{\scriptsize$\pm$0.82} & 9.50{\scriptsize$\pm$18.49} & \cellcolor{secondcolor}0.85{\scriptsize$\pm$0.91} \\
 & IntrinsicHDR~\cite{dille2024intrinsichdr} & 30.23{\scriptsize$\pm$3.76} & 0.374{\scriptsize$\pm$0.643} & 1.505{\scriptsize$\pm$2.419} & 35.162{\scriptsize$\pm$52.750} & 4.92{\scriptsize$\pm$4.86} & 3.32{\scriptsize$\pm$2.21} & 2.40{\scriptsize$\pm$1.38} & 7.39{\scriptsize$\pm$18.09} & 0.93{\scriptsize$\pm$1.01} \\
 & RDRE~\cite{dille2026rdre} & 23.82{\scriptsize$\pm$2.26} & 0.372{\scriptsize$\pm$0.567} & 1.489{\scriptsize$\pm$2.187} & 34.757{\scriptsize$\pm$52.354} & 5.29{\scriptsize$\pm$4.79} & 4.29{\scriptsize$\pm$2.16} & 3.97{\scriptsize$\pm$0.93} & 9.54{\scriptsize$\pm$23.19} & 0.95{\scriptsize$\pm$1.00} \\
\cmidrule(l){2-11}
 & GaSLight~\cite{bolduc2025gaslight} & 25.90{\scriptsize$\pm$3.26} & \cellcolor{secondcolor}0.334{\scriptsize$\pm$0.617} & \cellcolor{secondcolor}1.332{\scriptsize$\pm$2.278} & 34.274{\scriptsize$\pm$51.325} & 4.66{\scriptsize$\pm$3.23} & 3.60{\scriptsize$\pm$1.66} & 3.31{\scriptsize$\pm$1.35} & 6.78{\scriptsize$\pm$18.53} & 0.93{\scriptsize$\pm$0.94} \\
 & LEDiff~\cite{wang2025lediff} & 20.61{\scriptsize$\pm$2.62} & 0.403{\scriptsize$\pm$0.652} & 1.601{\scriptsize$\pm$2.452} & 34.680{\scriptsize$\pm$50.657} & 5.76{\scriptsize$\pm$4.90} & 5.05{\scriptsize$\pm$2.13} & 5.58{\scriptsize$\pm$1.36} & 8.43{\scriptsize$\pm$17.92} & 1.00{\scriptsize$\pm$1.00} \\
 & DiffHDR-Pano~\cite{yu2026promoting} & 23.78{\scriptsize$\pm$2.15} & 0.371{\scriptsize$\pm$0.591} & 1.488{\scriptsize$\pm$2.323} & \cellcolor{secondcolor}33.492{\scriptsize$\pm$51.200} & 5.27{\scriptsize$\pm$4.67} & 4.29{\scriptsize$\pm$1.84} & 4.06{\scriptsize$\pm$0.76} & \cellcolor{secondcolor}5.68{\scriptsize$\pm$19.39} & 0.92{\scriptsize$\pm$1.00} \\
\cmidrule(l){2-11}
 & ExpandNet~\cite{marnerides2018expandnet}$^\ast$ & 22.35{\scriptsize$\pm$3.53} & 0.401{\scriptsize$\pm$0.670} & 1.617{\scriptsize$\pm$2.498} & 35.309{\scriptsize$\pm$52.906} & 5.43{\scriptsize$\pm$4.74} & 4.04{\scriptsize$\pm$1.78} & 3.99{\scriptsize$\pm$0.92} & 18.32{\scriptsize$\pm$28.90} & 1.07{\scriptsize$\pm$1.03} \\
 & MaskHDR~\cite{santos2020maskhdr}$^\ast$ & 23.55{\scriptsize$\pm$2.47} & 0.403{\scriptsize$\pm$0.647} & 1.621{\scriptsize$\pm$2.477} & 35.290{\scriptsize$\pm$52.887} & 5.33{\scriptsize$\pm$4.48} & 4.30{\scriptsize$\pm$1.76} & 4.38{\scriptsize$\pm$1.10} & 9.12{\scriptsize$\pm$24.77} & 1.06{\scriptsize$\pm$1.07} \\
\cmidrule(l){2-11}
 & \textbf{\method} (ours) & \cellcolor{bestcolor}35.54{\scriptsize$\pm$4.36} & \cellcolor{bestcolor}0.124{\scriptsize$\pm$0.249} & \cellcolor{bestcolor}0.394{\scriptsize$\pm$0.765} & \cellcolor{bestcolor}20.461{\scriptsize$\pm$35.009} & \cellcolor{bestcolor}2.41{\scriptsize$\pm$2.08} & \cellcolor{bestcolor}1.46{\scriptsize$\pm$0.98} & \cellcolor{bestcolor}0.88{\scriptsize$\pm$0.37} & \cellcolor{bestcolor}1.83{\scriptsize$\pm$4.31} & \cellcolor{bestcolor}0.40{\scriptsize$\pm$0.48} \\
\midrule
\multirow{10}{*}[-4.6pt]{\rotatebox[origin=c]{90}{\begin{tabular}{@{}c@{}}Laval~\cite{gardner2017indoor}\\ 150 indoor scenes\end{tabular}}} & HDR-CNN~\cite{eilertsen2017hdrcnn} & 21.52{\scriptsize$\pm$1.88} & 0.133{\scriptsize$\pm$0.208} & 0.520{\scriptsize$\pm$0.857} & 2.767{\scriptsize$\pm$6.255} & \cellcolor{secondcolor}3.09{\scriptsize$\pm$1.34} & 3.84{\scriptsize$\pm$0.93} & 5.14{\scriptsize$\pm$1.27} & -- & 0.56{\scriptsize$\pm$0.39} \\
 & SingleHDR~\cite{liu2020singlehdr} & 22.04{\scriptsize$\pm$2.11} & \cellcolor{secondcolor}0.113{\scriptsize$\pm$0.178} & \cellcolor{secondcolor}0.431{\scriptsize$\pm$0.717} & \cellcolor{secondcolor}2.643{\scriptsize$\pm$5.367} & 3.98{\scriptsize$\pm$2.01} & 4.15{\scriptsize$\pm$1.42} & 5.18{\scriptsize$\pm$1.69} & -- & 0.58{\scriptsize$\pm$0.44} \\
 & IntrinsicHDR~\cite{dille2024intrinsichdr} & 24.75{\scriptsize$\pm$3.19} & 0.138{\scriptsize$\pm$0.224} & 0.555{\scriptsize$\pm$0.940} & 3.069{\scriptsize$\pm$6.587} & 3.17{\scriptsize$\pm$1.86} & 3.09{\scriptsize$\pm$1.44} & 3.40{\scriptsize$\pm$1.48} & -- & 0.47{\scriptsize$\pm$0.45} \\
 & RDRE~\cite{dille2026rdre} & 21.98{\scriptsize$\pm$1.99} & 0.134{\scriptsize$\pm$0.211} & 0.546{\scriptsize$\pm$0.901} & 3.073{\scriptsize$\pm$6.548} & 3.31{\scriptsize$\pm$2.29} & 3.88{\scriptsize$\pm$1.57} & 4.64{\scriptsize$\pm$1.08} & -- & \cellcolor{secondcolor}0.45{\scriptsize$\pm$0.37} \\
\cmidrule(l){2-11}
 & GaSLight~\cite{bolduc2025gaslight} & \cellcolor{secondcolor}25.16{\scriptsize$\pm$2.77} & 0.123{\scriptsize$\pm$0.203} & 0.479{\scriptsize$\pm$0.827} & 2.882{\scriptsize$\pm$6.130} & 3.29{\scriptsize$\pm$2.26} & \cellcolor{secondcolor}2.63{\scriptsize$\pm$1.07} & \cellcolor{secondcolor}2.45{\scriptsize$\pm$1.06} & -- & 0.47{\scriptsize$\pm$0.40} \\
 & LEDiff~\cite{wang2025lediff} & 18.76{\scriptsize$\pm$1.80} & 0.151{\scriptsize$\pm$0.230} & 0.588{\scriptsize$\pm$0.935} & 2.927{\scriptsize$\pm$6.165} & 4.21{\scriptsize$\pm$1.97} & 5.07{\scriptsize$\pm$1.34} & 6.79{\scriptsize$\pm$1.60} & -- & 0.59{\scriptsize$\pm$0.41} \\
 & DiffHDR-Pano~\cite{yu2026promoting} & 22.06{\scriptsize$\pm$1.77} & 0.119{\scriptsize$\pm$0.184} & 0.450{\scriptsize$\pm$0.871} & 2.849{\scriptsize$\pm$6.494} & 4.52{\scriptsize$\pm$2.95} & 4.45{\scriptsize$\pm$1.46} & 5.01{\scriptsize$\pm$0.99} & -- & 0.49{\scriptsize$\pm$0.68} \\
\cmidrule(l){2-11}
 & ExpandNet~\cite{marnerides2018expandnet}$^\ast$ & 22.51{\scriptsize$\pm$3.12} & 0.147{\scriptsize$\pm$0.228} & 0.587{\scriptsize$\pm$0.948} & 3.149{\scriptsize$\pm$6.625} & 3.78{\scriptsize$\pm$2.12} & 3.77{\scriptsize$\pm$1.59} & 4.05{\scriptsize$\pm$1.53} & -- & 0.52{\scriptsize$\pm$0.44} \\
 & MaskHDR~\cite{santos2020maskhdr}$^\ast$ & 21.13{\scriptsize$\pm$1.81} & 0.143{\scriptsize$\pm$0.217} & 0.564{\scriptsize$\pm$0.891} & 3.031{\scriptsize$\pm$6.456} & 3.36{\scriptsize$\pm$1.55} & 4.04{\scriptsize$\pm$1.00} & 5.23{\scriptsize$\pm$1.32} & -- & 0.62{\scriptsize$\pm$0.40} \\
\cmidrule(l){2-11}
 & \textbf{\method} (ours) & \cellcolor{bestcolor}30.26{\scriptsize$\pm$3.73} & \cellcolor{bestcolor}0.074{\scriptsize$\pm$0.141} & \cellcolor{bestcolor}0.265{\scriptsize$\pm$0.480} & \cellcolor{bestcolor}2.087{\scriptsize$\pm$4.744} & \cellcolor{bestcolor}1.69{\scriptsize$\pm$1.21} & \cellcolor{bestcolor}1.49{\scriptsize$\pm$0.80} & \cellcolor{bestcolor}1.38{\scriptsize$\pm$0.63} & -- & \cellcolor{bestcolor}0.28{\scriptsize$\pm$0.33} \\
\midrule
\multirow{10}{*}[-4.6pt]{\rotatebox[origin=c]{90}{\begin{tabular}{@{}c@{}}Pano-Pano~\cite{ji2025digital}\\ 177 scenes\end{tabular}}} & HDR-CNN~\cite{eilertsen2017hdrcnn} & 21.15{\scriptsize$\pm$1.90} & 0.091{\scriptsize$\pm$0.092} & 0.328{\scriptsize$\pm$0.327} & \cellcolor{secondcolor}1.077{\scriptsize$\pm$1.033} & 2.80{\scriptsize$\pm$1.40} & 3.80{\scriptsize$\pm$1.14} & 6.10{\scriptsize$\pm$1.68} & -- & 0.37{\scriptsize$\pm$0.30} \\
 & SingleHDR~\cite{liu2020singlehdr} & 26.47{\scriptsize$\pm$2.69} & \cellcolor{secondcolor}0.071{\scriptsize$\pm$0.060} & \cellcolor{secondcolor}0.294{\scriptsize$\pm$0.275} & 1.151{\scriptsize$\pm$1.080} & \cellcolor{secondcolor}2.62{\scriptsize$\pm$1.65} & 2.58{\scriptsize$\pm$0.91} & \cellcolor{secondcolor}3.49{\scriptsize$\pm$0.92} & -- & 0.34{\scriptsize$\pm$0.24} \\
 & IntrinsicHDR~\cite{dille2024intrinsichdr} & \cellcolor{secondcolor}26.56{\scriptsize$\pm$3.01} & 0.112{\scriptsize$\pm$0.121} & 0.448{\scriptsize$\pm$0.487} & 1.362{\scriptsize$\pm$1.410} & 2.82{\scriptsize$\pm$1.66} & 2.90{\scriptsize$\pm$1.01} & 3.64{\scriptsize$\pm$1.00} & -- & 0.42{\scriptsize$\pm$0.35} \\
 & RDRE~\cite{dille2026rdre} & 22.11{\scriptsize$\pm$1.55} & 0.095{\scriptsize$\pm$0.103} & 0.369{\scriptsize$\pm$0.384} & 1.238{\scriptsize$\pm$1.187} & 3.22{\scriptsize$\pm$2.18} & 3.88{\scriptsize$\pm$1.32} & 5.35{\scriptsize$\pm$1.07} & -- & \cellcolor{secondcolor}0.33{\scriptsize$\pm$0.30} \\
\cmidrule(l){2-11}
 & GaSLight~\cite{bolduc2025gaslight} & 25.42{\scriptsize$\pm$2.51} & 0.084{\scriptsize$\pm$0.074} & 0.364{\scriptsize$\pm$0.348} & 1.355{\scriptsize$\pm$1.308} & 2.78{\scriptsize$\pm$2.11} & \cellcolor{secondcolor}2.51{\scriptsize$\pm$1.06} & 3.83{\scriptsize$\pm$2.35} & -- & 0.36{\scriptsize$\pm$0.28} \\
 & LEDiff~\cite{wang2025lediff} & 18.55{\scriptsize$\pm$1.49} & 0.118{\scriptsize$\pm$0.107} & 0.443{\scriptsize$\pm$0.403} & 1.298{\scriptsize$\pm$1.230} & 4.19{\scriptsize$\pm$2.15} & 5.45{\scriptsize$\pm$1.69} & 8.27{\scriptsize$\pm$1.97} & -- & 0.46{\scriptsize$\pm$0.36} \\
 & DiffHDR-Pano~\cite{yu2026promoting} & 21.98{\scriptsize$\pm$1.42} & 0.110{\scriptsize$\pm$0.102} & 0.319{\scriptsize$\pm$0.288} & 1.166{\scriptsize$\pm$1.119} & 5.87{\scriptsize$\pm$4.70} & 4.91{\scriptsize$\pm$2.43} & 5.40{\scriptsize$\pm$0.97} & -- & 0.81{\scriptsize$\pm$0.96} \\
\cmidrule(l){2-11}
 & ExpandNet~\cite{marnerides2018expandnet}$^\ast$ & 21.58{\scriptsize$\pm$2.39} & 0.106{\scriptsize$\pm$0.113} & 0.435{\scriptsize$\pm$0.470} & 1.354{\scriptsize$\pm$1.404} & 3.80{\scriptsize$\pm$1.62} & 4.25{\scriptsize$\pm$1.25} & 5.35{\scriptsize$\pm$1.37} & -- & 0.41{\scriptsize$\pm$0.36} \\
 & MaskHDR~\cite{santos2020maskhdr}$^\ast$ & 20.39{\scriptsize$\pm$1.80} & 0.130{\scriptsize$\pm$0.130} & 0.497{\scriptsize$\pm$0.502} & 1.434{\scriptsize$\pm$1.393} & 3.23{\scriptsize$\pm$1.65} & 4.09{\scriptsize$\pm$1.27} & 6.20{\scriptsize$\pm$1.69} & -- & 0.56{\scriptsize$\pm$0.38} \\
\cmidrule(l){2-11}
 & \textbf{\method} (ours) & \cellcolor{bestcolor}28.66{\scriptsize$\pm$2.50} & \cellcolor{bestcolor}0.051{\scriptsize$\pm$0.045} & \cellcolor{bestcolor}0.197{\scriptsize$\pm$0.165} & \cellcolor{bestcolor}0.892{\scriptsize$\pm$0.864} & \cellcolor{bestcolor}2.05{\scriptsize$\pm$1.44} & \cellcolor{bestcolor}2.00{\scriptsize$\pm$0.62} & \cellcolor{bestcolor}2.73{\scriptsize$\pm$0.74} & -- & \cellcolor{bestcolor}0.20{\scriptsize$\pm$0.21} \\
\bottomrule
\end{tabular}
}
{}
\end{figure*}
\subsection{Camera response}
\label{sec:supp_crf_more}
\rpar{Per curve.} \Cref{tab:supp_crf} breaks \cref*{tab:crf} down by curve, with 100 scenes each.

\rpar{Film input.} Five of the ten curves of \cref*{tab:crf} are film curves. \Cref{tab:supp_film} tests the whole reconstruction on film: the 181 \sihdr scenes are rendered as Kodak Ektachrome 100D reversal film, with one curve per colour layer and with film grain (\cref{sec:supp_sets}). \method reaches a PU21-PSNR of 21.93\,dB and an FID of 13.62. The next method, IntrinsicHDR, reaches 17.18\,dB and 18.39. SingleHDR and IntrinsicHDR have a higher PU21-SSIM than \method (0.892 and 0.857 against 0.847). \Cref{fig:supp_qual_film} shows three of the scenes.

\begin{figure*}[t]
\centering
{\begin{minipage}[t]{0.55\linewidth}
\centering
\footnotesize
\setlength{\tabcolsep}{3.5pt}
\captionof{table}{\textbf{Camera response estimation per curve} (100 scenes each). Abs and shape error in stops, median over the scenes, as in \cref*{tab:crf}. The lowest abs and the lowest shape error of each row are highlighted.}
\label{tab:supp_crf}
\begin{tabular}{lcccccccc}
\toprule
\multirow{2}{*}[-2.4pt]{\textbf{Curve}} & \multicolumn{2}{c}{sRGB} & \multicolumn{2}{c}{CRF-net~\cite{li2017crfnet}} & \multicolumn{2}{c}{SingleHDR~\cite{liu2020singlehdr}} & \multicolumn{2}{c}{\textbf{Ours}} \\
\cmidrule(lr){2-3} \cmidrule(lr){4-5} \cmidrule(lr){6-7} \cmidrule(lr){8-9}
 & Abs & Shape & Abs & Shape & Abs & Shape & Abs & Shape \\
\midrule
\multicolumn{9}{@{}l}{\emph{Cameras}} \\
Canon R5 & 0.59 & 0.41 & 0.41 & 0.21 & 0.30 & 0.17 & \cellcolor{bestcolor}0.27 & \cellcolor{bestcolor}0.09 \\
Fujifilm X-T4 & 0.53 & 0.38 & 0.35 & 0.21 & 0.28 & 0.15 & \cellcolor{bestcolor}0.19 & \cellcolor{bestcolor}0.10 \\
Nikon D850 & 0.71 & 0.45 & 0.50 & 0.24 & 0.39 & 0.21 & \cellcolor{bestcolor}0.38 & \cellcolor{bestcolor}0.07 \\
Nikon D7000 & 0.63 & 0.39 & 0.40 & 0.24 & 0.31 & 0.14 & \cellcolor{bestcolor}0.23 & \cellcolor{bestcolor}0.09 \\
Sony ILCE-1 & 0.80 & 0.39 & 0.51 & 0.23 & 0.46 & 0.17 & \cellcolor{bestcolor}0.38 & \cellcolor{bestcolor}0.10 \\
\midrule
\multicolumn{9}{@{}l}{\emph{Films}} \\
Portra 400 & 0.51 & 0.33 & 0.25 & 0.13 & 0.83 & 0.22 & \cellcolor{bestcolor}0.17 & \cellcolor{bestcolor}0.10 \\
Vision3 500T & 0.45 & \cellcolor{bestcolor}0.10 & 0.32 & 0.13 & 0.33 & 0.25 & \cellcolor{bestcolor}0.21 & 0.13 \\
Velvia 50 & 0.81 & 0.78 & 0.31 & 0.21 & 0.79 & 0.46 & \cellcolor{bestcolor}0.30 & \cellcolor{bestcolor}0.08 \\
Ektachrome 100D & 0.82 & 0.75 & 0.31 & 0.20 & 0.81 & 0.43 & \cellcolor{bestcolor}0.21 & \cellcolor{bestcolor}0.07 \\
HP5 Plus & 0.71 & 0.38 & \cellcolor{bestcolor}0.40 & 0.20 & 1.13 & 0.20 & 0.41 & \cellcolor{bestcolor}0.13 \\
\bottomrule
\end{tabular}
\end{minipage}\hfill
\begin{minipage}[t]{0.41\linewidth}
\centering
\footnotesize
\setlength{\tabcolsep}{4pt}
\captionof{table}{\textbf{HDR reconstruction from film input.} The 181 \sihdr scenes rendered as reversal film (\cref{sec:supp_sets}). Metrics as in \cref*{tab:sihdr}. GaSLight is run with its public weights.}
\label{tab:supp_film}
\begin{tabular}{lccc}
\toprule
\textbf{Method} & PU21-PSNR $\uparrow$ & PU21-SSIM $\uparrow$ & FID $\downarrow$ \\
\midrule
HDR-CNN~\cite{eilertsen2017hdrcnn} & 12.31{\scriptsize$\pm$2.67} & 0.583{\scriptsize$\pm$0.197} & 26.57 \\
ExpandNet~\cite{marnerides2018expandnet} & 16.73{\scriptsize$\pm$3.54} & 0.837{\scriptsize$\pm$0.070} & 25.34 \\
MaskHDR~\cite{santos2020maskhdr} & 12.29{\scriptsize$\pm$2.69} & 0.599{\scriptsize$\pm$0.203} & 27.72 \\
SingleHDR~\cite{liu2020singlehdr} & 15.74{\scriptsize$\pm$2.97} & \cellcolor{bestcolor}0.892{\scriptsize$\pm$0.044} & 24.43 \\
IntrinsicHDR~\cite{dille2024intrinsichdr} & \cellcolor{secondcolor}17.18{\scriptsize$\pm$2.64} & \cellcolor{secondcolor}0.857{\scriptsize$\pm$0.070} & \cellcolor{secondcolor}18.39 \\
RDRE~\cite{dille2026rdre} & 13.01{\scriptsize$\pm$2.17} & 0.645{\scriptsize$\pm$0.178} & 22.04 \\
\midrule
GaSLight~\cite{bolduc2025gaslight} & 14.55{\scriptsize$\pm$2.58} & 0.641{\scriptsize$\pm$0.215} & 47.51 \\
LEDiff~\cite{wang2025lediff} & 10.78{\scriptsize$\pm$2.76} & 0.509{\scriptsize$\pm$0.186} & 35.93 \\
DiffHDR~\cite{yu2026diffhdr} & 12.66{\scriptsize$\pm$2.85} & 0.594{\scriptsize$\pm$0.157} & 27.31 \\
\midrule
\textbf{\method} (ours) & \cellcolor{bestcolor}21.93{\scriptsize$\pm$2.84} & 0.847{\scriptsize$\pm$0.097} & \cellcolor{bestcolor}13.62 \\
\bottomrule
\end{tabular}
\end{minipage}
}
{}
\vspace{10pt}
{\centering
\footnotesize
\setlength{\tabcolsep}{5.5pt}
\captionof{table}{\textbf{Noise-adaptive anchoring per ISO setting} (181 scenes per setting): paste-back, every fixed number of free steps $K$ and the adaptive rule. \Cref*{fig:adaptive} plots these values. Results are colour-coded by \legendbox{bestcolor}{best} and \legendbox{secondcolor}{second best} per column.}
\label{tab:supp_ladder}
\begin{tabular}{lccccccccccc}
\toprule
\textbf{ISO} & 100 & 200 & 400 & 800 & 1600 & 3200 & 6400 & 12800 & 25600 & 51200 & Mean \\
\midrule
\multicolumn{12}{@{}l}{\textbf{PU21-SSIM} $\uparrow$} \\
Paste-back & \cellcolor{bestcolor}0.979 & \cellcolor{bestcolor}0.978 & \cellcolor{bestcolor}0.978 & \cellcolor{bestcolor}0.975 & \cellcolor{bestcolor}0.969 & \cellcolor{bestcolor}0.957 & 0.936 & 0.896 & 0.830 & 0.748 & 0.925 \\
$K = 1$ & 0.949 & 0.948 & 0.948 & 0.947 & 0.945 & 0.942 & 0.939 & 0.933 & 0.926 & 0.916 & 0.939 \\
$K = 2$ & 0.947 & 0.947 & 0.947 & 0.946 & 0.944 & 0.942 & 0.940 & 0.935 & 0.928 & \cellcolor{secondcolor}0.919 & 0.939 \\
$K = 3$ & 0.947 & 0.947 & 0.947 & 0.946 & 0.945 & 0.942 & 0.939 & 0.932 & 0.920 & 0.905 & 0.937 \\
$K = 4$ & 0.947 & 0.947 & 0.947 & 0.946 & 0.945 & 0.942 & 0.939 & 0.932 & 0.921 & 0.904 & 0.937 \\
$K = 5$ & 0.947 & 0.947 & 0.947 & 0.946 & 0.945 & 0.942 & 0.939 & 0.933 & 0.921 & 0.905 & 0.937 \\
$K = 6$ & 0.947 & 0.947 & 0.947 & 0.946 & 0.945 & 0.943 & 0.940 & 0.933 & 0.922 & 0.907 & 0.938 \\
$K = 7$ & 0.947 & 0.947 & 0.947 & 0.946 & 0.945 & 0.943 & 0.940 & 0.934 & 0.924 & 0.908 & 0.938 \\
$K = 8$ & 0.947 & 0.947 & 0.947 & 0.946 & 0.945 & 0.943 & 0.941 & 0.935 & 0.926 & 0.911 & 0.939 \\
$K = 9$ & 0.947 & 0.947 & 0.947 & 0.946 & 0.946 & 0.944 & \cellcolor{secondcolor}0.942 & \cellcolor{secondcolor}0.937 & \cellcolor{secondcolor}0.929 & 0.917 & 0.940 \\
$K = 10$ & 0.948 & 0.948 & 0.948 & 0.948 & \cellcolor{secondcolor}0.947 & \cellcolor{secondcolor}0.946 & \cellcolor{bestcolor}0.944 & \cellcolor{bestcolor}0.939 & \cellcolor{bestcolor}0.933 & \cellcolor{bestcolor}0.924 & \cellcolor{secondcolor}0.942 \\
Adaptive & \cellcolor{secondcolor}0.971 & \cellcolor{secondcolor}0.969 & \cellcolor{secondcolor}0.968 & \cellcolor{secondcolor}0.963 & 0.945 & 0.943 & \cellcolor{bestcolor}0.944 & \cellcolor{bestcolor}0.939 & \cellcolor{bestcolor}0.933 & \cellcolor{bestcolor}0.924 & \cellcolor{bestcolor}0.950 \\
\midrule
\multicolumn{12}{@{}l}{\textbf{FID} $\downarrow$} \\
Paste-back & \cellcolor{bestcolor}2.65 & \cellcolor{bestcolor}2.92 & \cellcolor{bestcolor}3.22 & \cellcolor{bestcolor}4.16 & \cellcolor{secondcolor}5.56 & 7.83 & 11.06 & 16.10 & 23.05 & 31.18 & 10.77 \\
$K = 1$ & 7.02 & 6.58 & 6.52 & 6.23 & 6.09 & 6.23 & 6.53 & 7.17 & 8.16 & 9.69 & 7.02 \\
$K = 2$ & 6.89 & 6.47 & 6.42 & 6.13 & 5.92 & \cellcolor{bestcolor}5.93 & \cellcolor{bestcolor}6.06 & \cellcolor{bestcolor}6.59 & \cellcolor{secondcolor}7.71 & \cellcolor{secondcolor}9.55 & \cellcolor{secondcolor}6.77 \\
$K = 3$ & 6.83 & 6.47 & 6.39 & 6.13 & 6.03 & 6.19 & 6.73 & 7.98 & 10.07 & 12.84 & 7.57 \\
$K = 4$ & 6.83 & 6.49 & 6.40 & 6.14 & 6.00 & 6.17 & 6.64 & 7.83 & 9.90 & 12.65 & 7.51 \\
$K = 5$ & 6.87 & 6.52 & 6.44 & 6.17 & 6.02 & \cellcolor{secondcolor}6.13 & 6.57 & 7.67 & 9.70 & 12.47 & 7.46 \\
$K = 6$ & 6.92 & 6.57 & 6.50 & 6.23 & 6.05 & 6.14 & 6.49 & 7.52 & 9.46 & 12.21 & 7.41 \\
$K = 7$ & 6.99 & 6.69 & 6.59 & 6.31 & 6.09 & 6.15 & 6.43 & 7.33 & 9.12 & 11.82 & 7.35 \\
$K = 8$ & 7.12 & 6.83 & 6.72 & 6.44 & 6.18 & 6.17 & 6.33 & 7.04 & 8.62 & 11.07 & 7.25 \\
$K = 9$ & 7.37 & 7.15 & 7.02 & 6.74 & 6.44 & 6.32 & \cellcolor{secondcolor}6.25 & \cellcolor{secondcolor}6.65 & 7.73 & 9.57 & 7.12 \\
$K = 10$ & 7.79 & 7.68 & 7.64 & 7.52 & 7.36 & 7.24 & 7.15 & 7.24 & \cellcolor{bestcolor}7.69 & \cellcolor{bestcolor}8.56 & 7.59 \\
Adaptive & \cellcolor{secondcolor}2.84 & \cellcolor{secondcolor}3.07 & \cellcolor{secondcolor}3.36 & \cellcolor{secondcolor}4.24 & \cellcolor{bestcolor}5.47 & 6.15 & 7.15 & 7.24 & \cellcolor{bestcolor}7.69 & \cellcolor{bestcolor}8.56 & \cellcolor{bestcolor}5.58 \\
\bottomrule
\end{tabular}
}
{}
\end{figure*}
\begin{figure*}[t]
\centering
{  \centering
  \includegraphics[width=\linewidth]{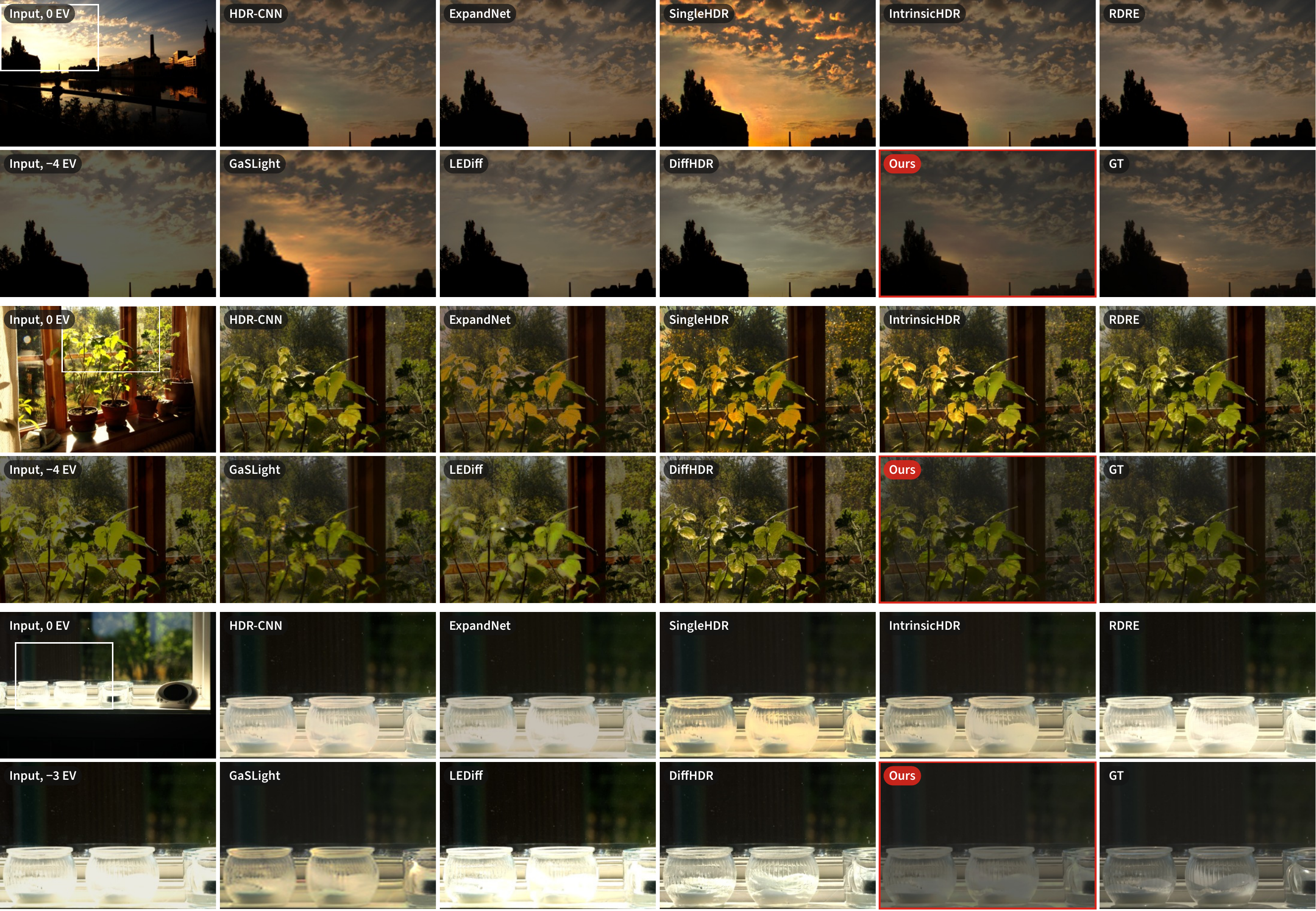}
  \captionof{figure}{\textbf{Results on film input.} Three of the \sihdr scenes rendered as reversal film (\cref{tab:supp_film}), shown as in \cref{fig:supp_qual_persp}.}
  \label{fig:supp_qual_film}}
\end{figure*}
\subsection{Ablations}
\label{sec:supp_adaptive}
\rpar{Per ISO setting.} \Cref{tab:supp_ladder} lists the values plotted in \cref*{fig:adaptive}. Up to ISO 800 the adaptive setting stays close to paste-back, and from ISO 6400 on it equals $K = 10$. At ISO 1600 and 3200 most scenes receive an intermediate $K$. Adaptive has the lowest FID at ISO 1600, and at ISO 3200 its FID of 6.15 lies 0.22 above the lowest. Its PU21-SSIM is below that of paste-back at both ISO settings: 0.945 against 0.969 at ISO 1600 and 0.943 against 0.957 at ISO 3200.

\rpar{Chosen settings.} \Cref{tab:supp_picks} counts the settings that \cref*{eq:adaptive_k} chooses on the test sets. The panorama inputs carry the sensor noise of ISO 100, and 252 of the 267 panoramas are reconstructed with paste-back. The \sihdr inputs carry strong simulated noise, and 128 of the 181 scenes run without anchoring.
\par\noindent\begin{minipage}{\linewidth}\begin{center}
\centering
\footnotesize
\setlength{\tabcolsep}{6pt}
\captionof{table}{\textbf{Settings chosen by noise-adaptive anchoring} on the test sets, in scenes. The rule is \cref*{eq:adaptive_k}.}
\label{tab:supp_picks}
\begin{tabular}{lrrrr}
\toprule
\textbf{Test set} & \textbf{Scenes} & $K = 0$ & $1 \le K \le 9$ & $K = 10$ \\
\midrule
\sihdr & 181 & 2 & 51 & 128 \\
Poly Haven & 117 & 113 & 3 & 1 \\
Laval & 150 & 139 & 7 & 4 \\
\bottomrule
\end{tabular}
\end{center}
{}\end{minipage}\par

\begin{figure*}[t]
\centering
{\centering
\footnotesize
\setlength{\tabcolsep}{5pt}
\captionof{table}{\textbf{Training ablation} on \sihdr (181 scenes). CRF: the training inputs pass through varied camera response curves (\cref{sec:supp_data}). Without it they pass through the sRGB curve only. DRaFT: reward fine-tuning (\cref{sec:supp_reward}). Every row is sampled with noise-adaptive anchoring, and the last row is \method. Metrics as in \cref*{tab:sihdr}. Results are colour-coded by \legendbox{bestcolor}{best} and \legendbox{secondcolor}{second best} per column.}
\label{tab:supp_stage}
\begin{tabular}{cccccccccc}
\toprule
CRF & DRaFT & PU21-PSNR $\uparrow$ & PU21-SSIM $\uparrow$ & HDR-VDP-3 $\uparrow$ & FID $\downarrow$ & PSNR-H $\uparrow$ & HEE $\downarrow$ & HQE $\downarrow$ & HSE $\downarrow$ \\
\midrule
-- & -- & 23.64{\scriptsize$\pm$4.00} & 0.902{\scriptsize$\pm$0.080} & 7.89{\scriptsize$\pm$0.70} & 11.08 & 20.43{\scriptsize$\pm$5.05} & 0.84{\scriptsize$\pm$0.83} & 0.76{\scriptsize$\pm$0.60} & 0.54{\scriptsize$\pm$0.40} \\
\checkmark & -- & \cellcolor{secondcolor}24.52{\scriptsize$\pm$3.57} & \cellcolor{secondcolor}0.923{\scriptsize$\pm$0.043} & \cellcolor{secondcolor}8.00{\scriptsize$\pm$0.69} & \cellcolor{secondcolor}10.27 & 20.05{\scriptsize$\pm$4.79} & 0.79{\scriptsize$\pm$0.77} & 0.81{\scriptsize$\pm$0.56} & 0.59{\scriptsize$\pm$0.43} \\
-- & \checkmark & 23.93{\scriptsize$\pm$3.99} & 0.905{\scriptsize$\pm$0.077} & 7.99{\scriptsize$\pm$0.63} & 13.17 & \cellcolor{secondcolor}20.51{\scriptsize$\pm$4.70} & \cellcolor{secondcolor}0.68{\scriptsize$\pm$0.65} & \cellcolor{secondcolor}0.62{\scriptsize$\pm$0.50} & \cellcolor{secondcolor}0.52{\scriptsize$\pm$0.39} \\
\midrule
\checkmark & \checkmark & \cellcolor{bestcolor}27.93{\scriptsize$\pm$3.25} & \cellcolor{bestcolor}0.934{\scriptsize$\pm$0.036} & \cellcolor{bestcolor}8.35{\scriptsize$\pm$0.64} & \cellcolor{bestcolor}8.75 & \cellcolor{bestcolor}23.58{\scriptsize$\pm$4.74} & \cellcolor{bestcolor}0.54{\scriptsize$\pm$0.59} & \cellcolor{bestcolor}0.49{\scriptsize$\pm$0.41} & \cellcolor{bestcolor}0.33{\scriptsize$\pm$0.26} \\
\bottomrule
\end{tabular}
}
{}
\end{figure*}
\rpar{Training.} \Cref{tab:supp_stage} removes two parts of the training. Without CRF, the training inputs pass through the sRGB curve only, and without DRaFT the model is not fine-tuned on the reward. The two rows without CRF are one model before and after reward fine-tuning, and the row with CRF alone is the model after supervised fine-tuning. Reward fine-tuning lowers the three highlight radiance errors in both cases: HEE falls from 0.84 to 0.68 stops without CRF and from 0.79 to 0.54 stops with it. \method, trained with both, is the best in every column.

\section{Method details}
\label{sec:supp_method}

\subsection{Encoding and latent grid}
\label{sec:supp_encoding}
The inverse of the encoding $\mathcal{T}$ of \cref*{eq:tonemap} is
\begin{equation}
\mathcal{T}^{-1}(y) =
\begin{cases}
\mathrm{sRGB}^{-1}(y / a), & y \le a, \\[2pt]
10^{\,d\,(y - a)/(1 - a)}, & y > a,
\end{cases}
\label{eq:supp_inverse}
\end{equation}
with $a = 0.7$ and $d = 6$. Radiance above $10^{d}$ is clamped to $y = 1$. Encoded targets are stored at 16 bits.

The VAE maps an $H \times W$ image to a latent of 16 channels on an $H/8 \times W/8$ grid. We call one position of this grid a \emph{cell}, so $h = H/8$ and $w = W/8$ in the main paper. The transformer reads the latent in patches of $2 \times 2$ cells. Image latents are the mode of the VAE posterior.

\subsection{Anchor weight}
\label{sec:supp_weight}
Let $c \in \{0, 1\}^{h \times w}$ mark the cells that contain at least one clipped pixel of the input at full resolution. The anchor weight of \cref*{eq:anchor_update} is
\begin{equation}
m = 1 - \max\big(\mathcal{G}_{1} * D_{2}(c),\; D_{1}(c)\big),
\label{eq:supp_weight}
\end{equation}
where $D_{r}$ dilates a mask by $r$ cells with a square window and $\mathcal{G}_{1}$ is a Gaussian blur with a standard deviation of one cell. The term $D_{1}(c)$ keeps $m$ at 0 on the marked cells and on the ring around them. The dilation and the blur reflect at the top and bottom border. For panoramas they wrap across the left and right border.

\subsection{Sensor-noise estimate}
\label{sec:supp_noise}
We compute the estimate $\hat{n}$ of \cref*{eq:adaptive_k} on the green channel of the 8-bit input, in pixel values from 0 to 255. We filter it with the Laplacian difference mask of Immerk{\ae}r~\cite{immerkaer1996fast} and take the absolute response $R$. A pixel is \emph{mid-tone} when its value lies in $[30, 230]$ and the mean of its $5 \times 5$ neighbourhood lies in $[40, 170]$. We keep the $8 \times 8$ blocks in which at least 90\,\% of the pixels are mid-tone. Each kept block gives the estimate $\sqrt{\pi/2}\,/\,6$ times the mean of $R$ over its mid-tone pixels, and $\hat{n}$ is the 10th percentile of these estimates. The low percentile selects the flattest blocks, where $R$ comes from the sensor noise and not from texture. If fewer than 50 blocks are kept, the image has no estimate and we set $K = T$.

\Cref{tab:supp_calib} shows the calibration of the two thresholds. We took 662 RAW images that we shot at ISO 100 and developed them to linear radiance. We added the measured sensor noise of the Canon EOS R5 at each ISO setting~\cite{darktable_noiseprofiles}, rendered the result through its camera response curve~\cite{imagingresource} to 8 bits, and computed $\hat{n}$. The threshold $\tau_1 = 1.2$ is the median estimate near ISO 800, and $\tau_2 = 2.0$ lies between the medians at ISO 3200 and ISO 6400.
\begin{table}[t]
\centering
\footnotesize
\setlength{\tabcolsep}{4pt}
\caption{\textbf{Calibration of the thresholds.} Median sensor-noise estimate $\hat{n}$, in pixel values, over our RAW images rendered with the sensor noise of each ISO setting (660 of the 662 images give an estimate). The thresholds are $\tau_1 = 1.2$ and $\tau_2 = 2.0$.}
\label{tab:supp_calib}
\begin{tabular}{lcccccccc}
\toprule
ISO & 100 & 200 & 400 & 800 & 1600 & 3200 & 6400 & 12800 \\
\midrule
$\hat{n}$ & 0.91 & 0.99 & 1.04 & 1.21 & 1.47 & 1.88 & 2.50 & 3.51 \\
\bottomrule
\end{tabular}
\end{table}
{}

\subsection{Paste-back}
\label{sec:supp_pasteback}
For $K = 0$ the output is composed in pixel space from the linearised input $g(I)$ and the generated radiance $\hat{x}^{\mathrm{gen}}$. Here $g$ is normalised so that $g(255) = 1$, which puts the output in units of the input's clip point:
\begin{equation}
\hat{x}_{p} =
\begin{cases}
s_{\mathrm{pb}}\,\hat{x}^{\mathrm{gen}}_{p}, & p \in \Omega_c, \\[2pt]
g(I)_{p}, & p \in \Omega_u,
\end{cases}
\label{eq:supp_pasteback}
\end{equation}
where the scalar $s_{\mathrm{pb}}$ brings the generated radiance to this unit,
\begin{equation}
s_{\mathrm{pb}} = \operatorname*{median}_{p \in \mathcal{M}} \frac{Y\big(g(I)\big)_{p}}{Y\big(\hat{x}^{\mathrm{gen}}\big)_{p}} .
\label{eq:supp_pastescale}
\end{equation}
In \cref{eq:supp_pastescale}, $Y$ is the Rec.\,709 luminance and $\mathcal{M}$ is the set of pixels whose three input values all lie between 20 and 230. The boundary between the two regions follows the pixels of $\Omega_c$ without blending.

\subsection{Extraction of the camera response curve}
\label{sec:supp_crf}
We extract the curve of \cref*{sec:crf} from the decoded one-step estimate $\hat{x}_{0}$. We use the pixels whose largest channel lies between 6 and 253 and pool the three channels, so one curve serves all of them. We group the pixel values into 64 bins of four values. A bin enters the fit when it holds at least 200 samples, and its value is the median of $\hat{x}_{0}$ over these samples. We make the medians monotone in $\log_2$ with the pool-adjacent-violators algorithm and interpolate linearly in $\log_2$ between the bin centres. Outside the covered bins the curve continues with the shape of the sRGB curve.

Near the clip point the medians can be unreliable, so we decide per image where the measured curve ends. Let $r$ be the median interquartile range of $\log_2 \hat{x}_{0}$ over the bins of pixel values 200 to 239. We accept the bins of pixel values 244 to 247 and 248 to 251 in this order while their interquartile range is at most $\max(0.3,\, 3 r)$, which places the last trusted value $c^{\ast}$ at 240, 245 or 250. Above $c^{\ast}$ the curve continues as a straight line in $\log_2$ radiance with the slope of the bins in $[c^{\ast} - 30,\, c^{\ast}]$.

The resulting curve keeps the exposure of the generator's estimate (\cref*{sec:crf}). The anchor latent is $z_a = \mathcal{E}\big(\mathcal{T}(g(I))\big)$ with this curve applied per channel. If fewer than four bins enter the fit, we fall back to the sRGB curve.

\subsection{Sampler and prompt}
\label{sec:supp_sampler}
\rpar{Schedule.} The $T = 10$ steps use the flow-matching schedule of the base model with its resolution-dependent shift. Starting from $u_i = 1 - i\,(1 - 1/T)/(T - 1)$ for $i = 0, \dots, T - 1$, we set
\begin{equation}
\sigma'_i = \frac{e^{\mu}}{e^{\mu} + 1/u_i - 1}, \qquad \mu = 0.5 + 0.4\,\frac{n - 256}{7936},
\label{eq:supp_shift}
\end{equation}
where $n = h w / 4$ is the number of output patches. The levels are then stretched so that the last one is 0.02,
\begin{equation}
\sigma_i = 1 - \big(1 - \sigma'_i\big)\,\frac{1 - 0.02}{1 - \sigma'_{T-1}},
\label{eq:supp_stretch}
\end{equation}
and $\sigma_T = 0$. For a $1280 \times 1888$ image this gives $\sigma = (1, 0.95, 0.89, 0.82, 0.74, 0.65, 0.54, 0.40, 0.24, 0.02, 0)$. With $K = 1$ the last anchored level is therefore $\sigma_9 = 0.02$, and with $K = 9$ the anchor is applied once, at $\sigma_1 = 0.95$.

\rpar{Sampling.} Each image is sampled once. The initial noise $\varepsilon_0$ is drawn from a seed fixed per image and is shared by the two unanchored steps of \cref*{sec:crf} and by the main run. Without classifier-free guidance, each step evaluates the transformer once.

\rpar{Prompt.} The task sentence is: ``Re-render this photo as a tone-mapped HDR image: keep the scene exactly as it is --- the same objects, layout and photographic style, inventing nothing --- while fully restoring the detail lost in over-exposed or crushed areas.'' For panoramas we add: ``This photo is a 360-degree equirectangular panorama: the left and right edges wrap around seamlessly, and the top and bottom edges are the zenith and nadir.'' The image description follows. It is one paragraph of 60 to 100 words, written by the vision-language model from the 8-bit input, about the setting, the main objects and every light source. For panoramas it avoids positions relative to the image frame.

\section{Training details}
\label{sec:supp_train}

\subsection{Training data}
\label{sec:supp_data}
The training pairs are built from HDR panoramas of Laval indoor, Laval outdoor~\cite{holdgeoffroy2019outdoor}, Poly Haven and UrbanSky~\cite{klotz2025urbansky}, and from perspective views cut out of them. The HDR side of a pair is scaled so that $x = 1$ is the clip point of its LDR side and is stored encoded with $\mathcal{T}$. The LDR side is rendered from the same radiance. A metering rule sets the exposure, and signal-dependent sensor noise is added. The result passes through a camera response curve drawn from a wide range of measured and synthetic curves and is quantised to 8 bits.

\subsection{VAE decoder}
\label{sec:supp_decoder}
We fine-tune the VAE decoder on the encoded HDR targets and keep the VAE encoder frozen. The losses cover the image below the clip point and the radiance above it. Part of each batch uses the composite latents that anchored sampling produces, so the decoder also learns to decode them.

\subsection{Supervised fine-tuning}
\label{sec:supp_sft}
We fine-tune the transformer with a LoRA. Training exposes the model to the anchored latents it meets during sampling. The noisy latent is formed from the anchor latent on the unclipped cells and from the ground-truth latent on the clipped cells. The target velocity points from this state to the ground truth. On the clipped cells this is the usual flow-matching target. On the anchored cells the model also learns to remove the noise, the quantisation and the curve error of the input. Part of the samples are trained without anchoring, so the model also learns the unanchored steps that \cref*{sec:crf} and the last $K$ steps rely on.

\subsection{Reward fine-tuning}
\label{sec:supp_reward}
The supervised objective measures errors in the latent space, where a small error on the shoulder of $\mathcal{T}$ is a large error in radiance. We therefore continue with reward fine-tuning: DRaFT~\cite{clark2024draft} back-propagates a differentiable reward through the last steps of an anchored sampling run. After the exposure of the estimate is removed, the reward compares the energy and the distribution of the reconstructed radiance in the clipped region with those of the ground truth. A regulariser keeps the model close to its state before reward fine-tuning.

\section{Evaluation details}
\label{sec:supp_eval}

\subsection{Test sets}
\label{sec:supp_sets}
\rpar{\sihdr.} We use all 181 scenes with the released clip-95 inputs at $1888 \times 1280$. The benchmark simulates each input from the ground truth: sensor noise is added in linear radiance, the result passes through a measured camera response curve drawn at random, and the exposure is set so that 5\,\% of the pixels saturate~\cite{hanji2022sihdr}.

\rpar{Panoramas.} The Poly Haven set~\cite{polyhaven2026hdris} holds the 117 outdoor environment maps that were published after we downloaded our training data. We use the official files at $1024 \times 2048$. The Laval set~\cite{gardner2017indoor} starts from 250 indoor panoramas held out from training. We removed, by feature matching, every panorama of a place that also appears in training and kept the 150 with the weakest matches. We downsample them to $1024 \times 2048$.

\rpar{Panorama inputs.} Every method receives the same 8-bit input, rendered from the ground truth $G$ in four steps. (i)~A metering model picks the exposure $e$ such that the mean luminance of $\min(eG, 2)$, weighted by solid angle, equals 0.18. (ii)~The radiance is clipped, $x = \min(eG, 1)$, and the ground truth is expressed in the same unit, so the clip point of the input is 1. (iii)~We add sensor noise of variance $\alpha x + \beta$ per channel with the measured parameters of the Canon EOS R5 at ISO 100~\cite{darktable_noiseprofiles}. (iv)~The result passes through the camera response curve of the Canon EOS R5, measured from RAW and JPEG pairs~\cite{imagingresource}, and is rounded to 8 bits.

\rpar{Pano-Pano.} We use 177 panoramas of Pano-Pano~\cite{ji2025digital}, captured with a Ricoh Theta Z1 as brackets of nine JPEG frames from 4 to 1/8000 seconds at ISO 100~\cite{ji2023virtual}. We merge each bracket to linear radiance with one response curve of the camera, recovered by the method of Debevec and Malik~\cite{debevec1997recovering}, and downsample the result to $1024 \times 2048$ as the ground truth. The input is the JPEG frame of the bracket nearest to the exposure that the metering of step~(i) picks with the radiance clipped at 1, downsampled to the same size.

\rpar{Noise ladder.} The inputs of \cref*{fig:adaptive} are rendered from the 181 \sihdr ground-truth images at ten ISO settings from 100 to 51200. They follow steps~(ii) to (iv) of the panorama inputs, with the noise parameters of step~(iii) taken at each ISO setting. The exposure is set so that the 95th percentile of the ground-truth channel values reaches the clip point. The ten inputs of a scene share one noise realisation, scaled by the parameters of each ISO setting.

\rpar{Response curves.} The benchmark of \cref*{tab:crf} starts from 100 RAW images~\cite{rawpixls}. We develop them to linear radiance at $1920 \times 1280$ with the camera white balance and clip them at the sensor white. Each image is then rendered through ten curves without noise, which gives 1,000 inputs. The camera curves are those of the Canon EOS R5, Nikon D850, Fujifilm X-T4, Sony ILCE-1 and Nikon D7000. The film curves are those of Kodak Ektachrome 100D~\cite{kodak2018ektachrome100d}, Fujifilm Velvia 50~\cite{fujifilm_velvia50}, Kodak Portra 400~\cite{kodak2025portra400}, Kodak Vision3 500T~\cite{kodak2015vision3} and Ilford HP5 Plus~\cite{harman2018hp5}. None of the ten curves is used in training.

\rpar{Scenes we photographed.} We photographed 46 scenes with a Sony ILCE-7RM5 as exposure brackets. The input is the JPEG that the camera rendered for one frame of the bracket, downsampled to $1920 \times 1280$. The ground truth is merged from the RAW frames of the bracket in linear radiance and downsampled to the same size. A light source that is still clipped in the shortest exposure has no valid ground truth. We therefore exclude from the metrics the pixels whose largest channel reaches 0.95 in that frame. People in these photographs are blurred in every image we show. All scores are computed on the unblurred images.

\rpar{Film input.} The film inputs are rendered from the 181 \sihdr ground-truth images. The exposure is set so that the 95th percentile of the ground-truth channel values reaches the clip point, as for the noise ladder. Each channel then passes through the characteristic curve of its colour layer of Kodak Ektachrome 100D, taken from the data sheet of the film~\cite{kodak2018ektachrome100d}. Film grain is added per channel with the granularity of the data sheet for a 35\,mm frame, and with the spatial and cross-channel correlation that we measured on scans of this film. The result is stored as an 8-bit image without lossy compression.

\subsection{Alignment and display frame}
\label{sec:supp_align}
Let $I \in [0, 1]$ be the 8-bit input divided by 255, $G$ the ground truth and $P$ a prediction, both in linear RGB with negative values set to 0, and let $Y(\cdot)$ be the Rec.\,709 luminance. We align the prediction with the scalar
\begin{equation}
k = \operatorname*{median}_{p \in \mathcal{A}} \frac{Y(G_p)}{Y(P_p)},
\label{eq:supp_align}
\end{equation}
where $\mathcal{A}$ holds the pixels with every channel of $I$ below 0.95 and $Y(I_p) > 0.05$. This set is smaller than $\Omega_u$. The scalar removes the absolute exposure, which a single LDR image does not determine. The camera response curve, the colour and the highlights are left to the method, so we apply no other correction. One scalar per image serves the three channels, and the prediction is not clipped.

The image metrics are computed in absolute luminance. We scale both images by $\rho = 203 / \omega$, where $\omega$ is the radiance of the input's white point in the unit of $G$, so that the white of the input is displayed at 203\,cd/m$^2$~\cite{itu2023bt2408}. For panoramas $\omega = 1$ by construction. For \sihdr the exposure of the input is not given, so we estimate $\omega$ from the ground truth. Let $f$ be the fraction of input pixels with a channel at or above 0.99. Then $\omega$ is the $(1 - f)$ quantile of the largest channel of $G$. The metrics compare $\tilde{P} = k\,\rho\,P$ with $\tilde{G} = \rho\,G$.

The lower rows of a Laval panorama are not observed and are zero in the ground truth. We exclude them from the panorama metrics and set them to zero in every environment map before rendering.

\begin{table*}[t]
\centering
\footnotesize
\setlength{\tabcolsep}{7pt}
\caption{\textbf{Baseline outputs that are scored.} Released reconstructions are stored at $1888 \times 1280$.}
\label{tab:supp_baselines}
\begin{tabular}{lll}
\toprule
\textbf{Method} & \textbf{\sihdr} & \textbf{Panoramas} \\
\midrule
HDR-CNN & released by the benchmark & official code at $1024 \times 2048$ \\
ExpandNet & released by the benchmark & official code at $1024 \times 2048$ \\
MaskHDR & released by the benchmark & official code at $1024 \times 2048$ \\
SingleHDR & released by the benchmark & official code with the refinement network \\
IntrinsicHDR & official code at input size & official code at input size \\
RDRE & official code at input size, 24 steps & official code at input size, 24 steps \\
GaSLight & released by the authors & public weights, $512 \times 512$ exposures merged at input size \\
LEDiff & official pipeline at $512 \times 512$, blended with the input & as on \sihdr \\
DiffHDR & official code at input size, 50 steps & DiffHDR-Pano at $1024 \times 2048$, 50 steps \\
\bottomrule
\end{tabular}
\end{table*}
{}

\subsection{Image metrics}
\label{sec:supp_metrics}
PU21-PSNR, PSNR-H, PU21-SSIM, VSI and PIQE operate on the PU21 encoding~\cite{mantiuk2021pu21} of $\tilde{P}$ and $\tilde{G}$. HDR-VDP-3 takes their absolute luminance, and FID uses its own tone mapping. We use the PU21 variant for banding and glare, which maps luminance between 0.005 and $10^4$\,cd/m$^2$ to approximately perceptually uniform values. Our implementations of PU21-PSNR, PU21-SSIM and VSI reproduce the authors' MATLAB code to machine precision.

\rpar{PU21-PSNR} encodes each RGB channel and uses a peak of 256,
\begin{equation}
\text{PU21-PSNR} = 10 \log_{10} \frac{256^2}{\frac{1}{3N} \sum_{p, c} \big(\mathrm{PU}(\tilde{P}_{p,c}) - \mathrm{PU}(\tilde{G}_{p,c})\big)^2},
\label{eq:supp_psnr}
\end{equation}
with $N$ the number of pixels. \textbf{PSNR-H} restricts the sum to the pixels with any channel of $I$ above 0.8. \textbf{PU21-SSIM} is the SSIM~\cite{wang2004ssim} of the PU21-encoded luminance with a dynamic range of 256. \textbf{VSI}~\cite{zhang2014vsi} is computed on the PU21-encoded RGB image.

\rpar{HDR-VDP-3.} We report the quality score $Q$ of HDR-VDP-3~\cite{mantiuk2023hdrvdp3}, version 3.0.7. We run the metric in its \emph{quality} task on the absolute luminance of the two images at 60 pixels per degree, with all other options at their defaults.

\rpar{FID.} We first scale the ground truth and the aligned prediction $kP$ of a scene by one factor, which sets the median luminance of the ground truth to 0.18. Each image is then tone-mapped with the global operator of Reinhard~\cite{reinhard2002photographic}, $x' = x / (1 + Y(x))$, clipped to $[0, 1]$, encoded with the sRGB curve and rounded to 8 bits. From every scene we draw 60 random $128 \times 128$ crops at positions shared by all methods and the ground truth, which gives 10,860 patches per method. FID~\cite{heusel2017fid} is computed between the patches of a method and those of the ground truth with the standard Inception features.

\rpar{PIQE}~\cite{venkatanath2015piqe} is applied to the PU21 encoding of $\tilde{P}$ and needs no ground truth.

\subsection{Highlight radiance error}
\label{sec:supp_hre}
The three terms of \cref*{eq:hre} are computed on the aligned luminances $\hat{L} = Y(\tilde{P})$ and $L = Y(\tilde{G})$ over the region $\Omega$ of pixels with any channel of $I$ at or above 0.95. This region is larger than the clipped region $\Omega_c$ of the method, which starts at pixel value 250, so it also covers the roll-off of the camera response curve next to the clip point. The statistics are
\begin{align}
\mathcal{S}_{\mathrm{HEE}} &= \Big\{ \textstyle\sum_{p \in \Omega} L_p \Big\}, \label{eq:supp_hee} \\
\mathcal{S}_{\mathrm{HQE}} &= \big\{ Q_{50}(L_\Omega),\; Q_{90}(L_\Omega),\; Q_{99}(L_\Omega) \big\}, \label{eq:supp_hqe} \\
\mathcal{S}_{\mathrm{HSE}} &= \big\{ \bar{L}_{c} : c \in \mathcal{C} \big\}, \label{eq:supp_hse}
\end{align}
where $Q_q$ is the $q$-th percentile over the pixels of $\Omega$, $\bar{L}_c$ is the mean over the $8 \times 8$ pixel cell $c$ and $\mathcal{C}$ holds the cells with at least half of their pixels in $\Omega$. For HQE and HSE we add $10^{-4}$\,cd/m$^2$ to both values before the logarithm. Because the ratios are taken after the alignment of \cref{eq:supp_align}, a global exposure offset of the prediction does not enter them. They measure the clipped region relative to the unclipped region.

\subsection{Path-traced spheres}
\label{sec:supp_spheres}
\rpar{Rendering.} We light a unit sphere with the environment map and render it with the Cycles path tracer of Blender, following the scenes of the DiffusionLight evaluation~\cite{phongthawee2024diffusionlight}. The map is the aligned prediction $kP$ or the ground truth $G$ at $1024 \times 2048$, used as it is without tone mapping. The three materials share a base colour of 0.8: a diffuse material with a roughness of 0.36, a metal with a roughness of 0.33 (matte silver) and a metal with a roughness of 0 (mirror). Each render has $960 \times 540$ pixels and 4,096 samples per pixel, without denoising or clamping. The metrics use the pixels that show the sphere.

\rpar{Metrics.} Let $\hat{R}$ and $R$ be the renders under the prediction and the ground truth, and $N$ the number of sphere pixels. The scale-invariant RMSE is
\begin{equation}
\text{si-RMSE} = \sqrt{\frac{1}{N} \min_{\gamma} \sum_{p} \big\| \gamma \hat{R}_p - R_p \big\|_2^2} ,
\label{eq:supp_sirmse}
\end{equation}
in the unit of the ground truth, where the clip point of the input is 1. The angular error is the mean angle between the RGB vectors $\hat{R}_p$ and $R_p$, in degrees. The energy error of the diffuse sphere is
\begin{equation}
\text{Energy} = \left| \log_2 \frac{\sum_{p, c} \hat{R}_{p,c}}{\sum_{p, c} R_{p,c}} \right| ,
\label{eq:supp_energy_sphere}
\end{equation}
in stops. Unlike \cref{eq:supp_sirmse} it keeps the scale of the prediction, so it measures whether the light sources carry the right energy relative to the unclipped region.

\rpar{Sun direction.} We locate the sun in the prediction and in the ground truth with the same two-stage search: the brightest $20^\circ$ block of a low-resolution luminance map, then the luminance-weighted centroid within $1.5^\circ$ of the brightest point of the full-resolution map. The error is the angle between the two directions. We report it on the 58 Poly Haven scenes that have a sun above 50 times the clip point and are not night scenes.

\subsection{Response-curve error}
\label{sec:supp_crferr}
For each input we measure the ground-truth curve $t$ from the input and its linear image, as the median radiance of each bin of four pixel values, in $\log_2$. With $\hat{g}$ an estimated curve and $d_c = \log_2 \hat{g}(c) - t(c)$ at pixel value $c$,
\begin{equation}
\text{Abs} = \frac{1}{|\mathcal{K}|} \sum_{c \in \mathcal{K}} |d_c|, \qquad
\text{Shape} = \frac{1}{|\mathcal{K}|} \sum_{c \in \mathcal{K}} \big| d_c - \operatorname*{median}_{\mathcal{K}} d \big| ,
\label{eq:supp_crferr}
\end{equation}
over the pixel values $\mathcal{K} = \{16, \dots, 240\}$. The sRGB curve, CRF-net~\cite{li2017crfnet} and the Linearization-Net of SingleHDR~\cite{liu2020singlehdr} give curves with $\hat{g}(255) = 1$. Our curve keeps the exposure of the generator (\cref*{sec:crf}). \Cref*{tab:crf} reports the median over the 500 inputs of a group and over all 1,000.

\subsection{Baselines}
\label{sec:supp_baselines}
\Cref{tab:supp_baselines} lists what is scored for every baseline. We read every output as linear radiance, set negative values to 0 and apply the alignment of \cref{eq:supp_align} to every method alike. Methods that work at a fixed resolution are run as their official code prescribes and their output is resized to the input size.

\subsection{Statistics}
\label{sec:supp_stats}
\Cref*{tab:sihdr,tab:pano} and \cref{tab:supp_persp,tab:supp_pano} report the mean and the sample standard deviation over scenes.
{}


{
    \small
     \begin{thebibliography}{92}
\providecommand{\natexlab}[1]{#1}
\providecommand{\url}[1]{\texttt{#1}}
\expandafter\ifx\csname urlstyle\endcsname\relax
  \providecommand{\doi}[1]{doi: #1}\else
  \providecommand{\doi}{doi: \begingroup \urlstyle{rm}\Url}\fi

\bibitem[Andersson et~al.(2021)Andersson, Nilsson, Shirley, and
  Akenine-M{\"o}ller]{andersson2021hdrflip}
Pontus Andersson, Jim Nilsson, Peter Shirley, and Tomas Akenine-M{\"o}ller.
\newblock {Visualizing Errors in Rendered High Dynamic Range Images}.
\newblock In \emph{Eurographics 2021 - Short Papers}, pages 25--28. The
  Eurographics Association, 2021.

\bibitem[And{\i}ran et~al.(2026)And{\i}ran, Yang, Lu, Salzmann, and
  S{\"{u}}sstrunk]{andiran2026expanddiff}
Mehmet~Emre And{\i}ran, Zhuoqian Yang, Liying Lu, Mathieu Salzmann, and Sabine
  S{\"{u}}sstrunk.
\newblock {ExpandDiff: Dynamic Range Expanding Diffusion for Single-Image {HDR}
  Reconstruction}.
\newblock \emph{CoRR}, abs/2609.39624, 2026.

\bibitem[Avrahami et~al.(2022)Avrahami, Lischinski, and
  Fried]{avrahami2022blended}
Omri Avrahami, Dani Lischinski, and Ohad Fried.
\newblock {Blended Diffusion for Text-driven Editing of Natural Images}.
\newblock In \emph{{IEEE/CVF} Conference on Computer Vision and Pattern
  Recognition, {CVPR} 2022, New Orleans, LA, USA, June 18-24, 2022}, pages
  18187--18197. {IEEE}, 2022.

\bibitem[Avrahami et~al.(2023)Avrahami, Fried, and Lischinski]{avrahami2023bld}
Omri Avrahami, Ohad Fried, and Dani Lischinski.
\newblock {Blended Latent Diffusion}.
\newblock \emph{{ACM} Trans. Graph.}, 42\penalty0 (4):\penalty0 149:1--149:11,
  2023.

\bibitem[Banterle et~al.(2017)Banterle, Artusi, Debattista, and
  Chalmers]{banterle2017advanced}
Francesco Banterle, Alessandro Artusi, Kurt Debattista, and Alan Chalmers.
\newblock \emph{{Advanced High Dynamic Range Imaging}}.
\newblock A K Peters/CRC Press, 2 edition, 2017.

\bibitem[Bemana et~al.(2025)Bemana, Leimk{\"{u}}hler, Myszkowski, Seidel, and
  Ritschel]{bemana2025bracket}
Mojtaba Bemana, Thomas Leimk{\"{u}}hler, Karol Myszkowski, Hans{-}Peter Seidel,
  and Tobias Ritschel.
\newblock {Bracket Diffusion: {HDR} Image Generation by Consistent {LDR}
  Denoising}.
\newblock \emph{Comput. Graph. Forum}, 44\penalty0 (2), 2025.

\bibitem[Bińkowski et~al.(2018)Bińkowski, Sutherland, Arbel, and
  Gretton]{binkowski2018kid}
Mikołaj Bińkowski, Danica~J. Sutherland, Michael Arbel, and Arthur Gretton.
\newblock {Demystifying {MMD} {GANs}}.
\newblock In \emph{International Conference on Learning Representations
  (ICLR)}, 2018.

\bibitem[Bolduc et~al.(2025)Bolduc, Hold{-}Geoffroy, and
  Lalonde]{bolduc2025gaslight}
Christophe Bolduc, Yannick Hold{-}Geoffroy, and Jean{-}Fran{\c{c}}ois Lalonde.
\newblock {GaSLight: Gaussian Splats for Spatially-Varying Lighting in {HDR}}.
\newblock In \emph{{IEEE/CVF} International Conference on Computer Vision,
  {ICCV} 2025, Honolulu, HI, USA, October 19-25, 2025}, pages 29120--29130.
  {IEEE}, 2025.

\bibitem[Chai et~al.(2022)Chai, Gharbi, Shechtman, Isola, and
  Zhang]{chai2022patchfid}
Lucy Chai, Micha{\"{e}}l Gharbi, Eli Shechtman, Phillip Isola, and Richard
  Zhang.
\newblock {Any-Resolution Training for High-Resolution Image Synthesis}.
\newblock In \emph{Computer Vision - {ECCV} 2022 - 17th European Conference,
  Tel Aviv, Israel, October 23-27, 2022, Proceedings, Part {XVI}}, pages
  170--188. Springer, 2022.

\bibitem[Chen et~al.(2019)Chen, McCloskey, and Yu]{chen2019moderncrf}
Can Chen, Scott McCloskey, and Jingyi Yu.
\newblock {Analyzing Modern Camera Response Functions}.
\newblock In \emph{{IEEE} Winter Conference on Applications of Computer Vision,
  {WACV} 2019, Waikoloa Village, HI, USA, January 7-11, 2019}, pages
  1961--1969. {IEEE}, 2019.

\bibitem[Chen et~al.(2021)Chen, Liu, Zhang, Qiao, and Dong]{chen2021hdrunet}
Xiangyu Chen, Yihao Liu, Zhengwen Zhang, Yu Qiao, and Chao Dong.
\newblock {HDRUNet: Single Image {HDR} Reconstruction With Denoising and
  Dequantization}.
\newblock In \emph{{IEEE} Conference on Computer Vision and Pattern Recognition
  Workshops, {CVPR} Workshops 2021, virtual, June 19-25, 2021}, pages 354--363.
  Computer Vision Foundation / {IEEE}, 2021.

\bibitem[Chung et~al.(2023)Chung, Kim, McCann, Klasky, and Ye]{chung2023dps}
Hyungjin Chung, Jeongsol Kim, Michael~Thompson McCann, Marc~Louis Klasky, and
  Jong~Chul Ye.
\newblock {Diffusion Posterior Sampling for General Noisy Inverse Problems}.
\newblock In \emph{The Eleventh International Conference on Learning
  Representations, {ICLR} 2023, Kigali, Rwanda, May 1-5, 2023}. OpenReview.net,
  2023.

\bibitem[Clark et~al.(2024)Clark, Vicol, Swersky, and Fleet]{clark2024draft}
Kevin Clark, Paul Vicol, Kevin Swersky, and David~J. Fleet.
\newblock {Directly Fine-Tuning Diffusion Models on Differentiable Rewards}.
\newblock In \emph{The Twelfth International Conference on Learning
  Representations, {ICLR} 2024, Vienna, Austria, May 7-11, 2024}.
  OpenReview.net, 2024.

\bibitem[{darktable developers}(2026)]{darktable_noiseprofiles}
{darktable developers}.
\newblock {darktable: camera noise profiles ({data/noiseprofiles.json})}.
\newblock \url{https://github.com/darktable-org/darktable}, 2026.

\bibitem[Dastjerdi et~al.(2023)Dastjerdi, Eisenmann, Hold{-}Geoffroy, and
  Lalonde]{dastjerdi2023everlight}
Mohammad Reza~Karimi Dastjerdi, Jonathan Eisenmann, Yannick Hold{-}Geoffroy,
  and Jean{-}Fran{\c{c}}ois Lalonde.
\newblock {EverLight: Indoor-Outdoor Editable {HDR} Lighting Estimation}.
\newblock In \emph{{IEEE/CVF} International Conference on Computer Vision,
  {ICCV} 2023, Paris, France, October 1-6, 2023}, pages 7386--7395. {IEEE},
  2023.

\bibitem[Debevec(1998)]{debevec1998rendering}
Paul~E. Debevec.
\newblock {Rendering Synthetic Objects into Real Scenes: Bridging Traditional
  and Image-based Graphics with Global Illumination and High Dynamic Range
  Photography}.
\newblock In \emph{Proceedings of the 25th Annual Conference on Computer
  Graphics and Interactive Techniques, {SIGGRAPH} 1998, Orlando, FL, USA, July
  19-24, 1998}, pages 189--198. {ACM}, 1998.

\bibitem[Debevec and Malik(1997)]{debevec1997recovering}
Paul~E. Debevec and Jitendra Malik.
\newblock {Recovering high dynamic range radiance maps from photographs}.
\newblock In \emph{Proceedings of the 24th Annual Conference on Computer
  Graphics and Interactive Techniques, {SIGGRAPH} 1997, Los Angeles, CA, USA,
  August 3-8, 1997}, pages 369--378. {ACM}, 1997.

\bibitem[Dille et~al.(2024)Dille, Careaga, and Aksoy]{dille2024intrinsichdr}
Sebastian Dille, Chris Careaga, and Ya\u{g}{\i}z Aksoy.
\newblock {Intrinsic Single-Image {HDR} Reconstruction}.
\newblock In \emph{Computer Vision - {ECCV} 2024 - 18th European Conference,
  Milan, Italy, September 29-October 4, 2024, Proceedings, Part {XI}}, pages
  161--177. Springer, 2024.

\bibitem[Dille et~al.(2026)Dille, Fu, Miangoleh, and Aksoy]{dille2026rdre}
Sebastian Dille, Keru Fu, S.~Mahdi~H. Miangoleh, and Ya\u{g}{\i}z Aksoy.
\newblock {Recurrent Dynamic Range Extension}.
\newblock In \emph{Proc. SIGGRAPH Asia}, 2026.

\bibitem[Durand and Dorsey(2002)]{durand2002bilateral}
Fr{\'e}do Durand and Julie Dorsey.
\newblock {Fast Bilateral Filtering for the Display of High-Dynamic-Range
  Images}.
\newblock \emph{ACM Transactions on Graphics}, 21\penalty0 (3):\penalty0
  257--266, 2002.

\bibitem[{Eastman Kodak Company}(2015)]{kodak2015vision3}
{Eastman Kodak Company}.
\newblock {{KODAK VISION3 500T} Color Negative Film 5219 / 7219 / SO-219,
  Technical Data}.
\newblock KODAK Publication No. H-1-5219t,
  \url{https://www.kodak.com/content/products-brochures/Film/VISION3-500T-Color-Negative-Film-7219-TECHNICAL-DATA.pdf},
  2015.

\bibitem[{Eastman Kodak Company}(2018)]{kodak2018ektachrome100d}
{Eastman Kodak Company}.
\newblock {{KODAK EKTACHROME 100D} Color Reversal Film / 7294, Technical Data}.
\newblock Publication H1-7294t, 2018.

\bibitem[Eilertsen et~al.(2017)Eilertsen, Kronander, Denes, Mantiuk, and
  Unger]{eilertsen2017hdrcnn}
Gabriel Eilertsen, Joel Kronander, Gyorgy Denes, Rafa{\l}~K. Mantiuk, and Jonas
  Unger.
\newblock {{HDR} image reconstruction from a single exposure using deep CNNs}.
\newblock \emph{{ACM} Trans. Graph.}, 36\penalty0 (6):\penalty0 178:1--178:15,
  2017.

\bibitem[Eilertsen et~al.(2021)Eilertsen, Hajisharif, Hanji, Tsirikoglou,
  Mantiuk, and Unger]{eilertsen2021cheat}
Gabriel Eilertsen, Saghi Hajisharif, Param Hanji, Apostolia Tsirikoglou,
  Rafa{\l}~K. Mantiuk, and Jonas Unger.
\newblock {How to cheat with metrics in single-image {HDR} reconstruction}.
\newblock In \emph{{IEEE/CVF} International Conference on Computer Vision
  Workshops, {ICCVW} 2021, Montreal, QC, Canada, October 11-17, 2021}, pages
  3981--3990. {IEEE}, 2021.

\bibitem[{EMPA Media Technology}(2013)]{empa_hdr_database}
{EMPA Media Technology}.
\newblock {Empa HDR Image Database}.
\newblock \url{http://www.empamedia.ethz.ch/hdrdatabase/index.php}, 2013.
\newblock Original site offline; archived copy of 2017-09-29 at
  web.archive.org.

\bibitem[Endo et~al.(2017)Endo, Kanamori, and Mitani]{endo2017drtmo}
Yuki Endo, Yoshihiro Kanamori, and Jun Mitani.
\newblock {Deep reverse tone mapping}.
\newblock \emph{{ACM} Trans. Graph.}, 36\penalty0 (6):\penalty0 177:1--177:10,
  2017.

\bibitem[Esser et~al.(2024)Esser, Kulal, Blattmann, Entezari, M{\"{u}}ller,
  Saini, Levi, Lorenz, Sauer, Boesel, Podell, Dockhorn, English, and
  Rombach]{esser2024sd3}
Patrick Esser, Sumith Kulal, Andreas Blattmann, Rahim Entezari, Jonas
  M{\"{u}}ller, Harry Saini, Yam Levi, Dominik Lorenz, Axel Sauer, Frederic
  Boesel, Dustin Podell, Tim Dockhorn, Zion English, and Robin Rombach.
\newblock {Scaling Rectified Flow Transformers for High-Resolution Image
  Synthesis}.
\newblock In \emph{Forty-first International Conference on Machine Learning,
  {ICML} 2024, Vienna, Austria, July 21-27, 2024}, pages 12606--12633. {PMLR} /
  OpenReview.net, 2024.

\bibitem[Fairchild(2007)]{fairchild2007hdrps}
Mark~D. Fairchild.
\newblock {The {HDR} Photographic Survey}.
\newblock In \emph{15th Color and Imaging Conference, {CIC} 2007, Albuquerque,
  New Mexico, USA, November 5-9, 2007}, pages 233--238. Society for Imaging
  Science and Technology, 2007.

\bibitem[Fekri et~al.(2026)Fekri, Li, Chen, and Altamirano]{fekri2026latenthdr}
Pedram Fekri, WenChen Li, William Chen, and Peter Altamirano.
\newblock {LatentHDR: Decoupling Exposure from Diffusion via Conditional
  Latent-to-Latent Mapping for Text/Image-to-Panoramic {HDR}}.
\newblock \emph{CoRR}, abs/2605.11115, 2026.

\bibitem[Feng et~al.(2026)Feng, Huang, Li, and Sheng]{feng2026personalize}
Haoran Feng, Zehuan Huang, Lin Li, and Lu Sheng.
\newblock {Personalize Anything for Free with Diffusion Transformer}.
\newblock In \emph{Fortieth {AAAI} Conference on Artificial Intelligence,
  Thirty-Eighth Conference on Innovative Applications of Artificial
  Intelligence, Sixteenth Symposium on Educational Advances in Artificial
  Intelligence, {AAAI} 2026, Singapore, January 20-27, 2026}, pages 3921--3929.
  {AAAI} Press, 2026.

\bibitem[{Fujifilm}(2006)]{fujifilm_velvia50}
{Fujifilm}.
\newblock {{FUJICHROME Velvia 50 Professional [RVP 50]} Data Sheet}.
\newblock Ref. No. AF3-0221E2, 2006.
\newblock Undated; PDF created 2006.

\bibitem[Gardner et~al.(2017)Gardner, Sunkavalli, Yumer, Shen, Gambaretto,
  Gagn{\'{e}}, and Lalonde]{gardner2017indoor}
Marc{-}Andr{\'{e}} Gardner, Kalyan Sunkavalli, Ersin Yumer, Xiaohui Shen,
  Emiliano Gambaretto, Christian Gagn{\'{e}}, and Jean{-}Fran{\c{c}}ois
  Lalonde.
\newblock {Learning to predict indoor illumination from a single image}.
\newblock \emph{{ACM} Trans. Graph.}, 36\penalty0 (6):\penalty0 176:1--176:14,
  2017.

\bibitem[Goswami et~al.(2025)Goswami, Singh, Banterle, Debattista, and
  Bashford-Rogers]{goswami2025ditmo}
Abhishek Goswami, Aru~Ranjan Singh, Francesco Banterle, Kurt Debattista, and
  Thomas Bashford-Rogers.
\newblock {Semantic Aware Diffusion Inverse Tone Mapping}.
\newblock \emph{Journal of Physics: Conference Series}, 3128\penalty0
  (1):\penalty0 012009, 2025.

\bibitem[Grossberg and Nayar(2003)]{grossberg2003space}
Michael~D. Grossberg and Shree~K. Nayar.
\newblock {What is the Space of Camera Response Functions?}
\newblock In \emph{2003 {IEEE} Computer Society Conference on Computer Vision
  and Pattern Recognition {(CVPR} 2003), 16-22 June 2003, Madison, WI, {USA}},
  pages 602--609. {IEEE} Computer Society, 2003.

\bibitem[Grossberg and Nayar(2004)]{grossberg2004emor}
Michael~D. Grossberg and Shree~K. Nayar.
\newblock {Modeling the Space of Camera Response Functions}.
\newblock \emph{{IEEE} Trans. Pattern Anal. Mach. Intell.}, 26\penalty0
  (10):\penalty0 1272--1282, 2004.

\bibitem[Grosse et~al.(2009)Grosse, Johnson, Adelson, and
  Freeman]{grosse2009intrinsic}
Roger~B. Grosse, Micah~K. Johnson, Edward~H. Adelson, and William~T. Freeman.
\newblock {Ground truth dataset and baseline evaluations for intrinsic image
  algorithms}.
\newblock In \emph{{IEEE} 12th International Conference on Computer Vision,
  {ICCV} 2009, Kyoto, Japan, September 27 - October 4, 2009}, pages 2335--2342.
  {IEEE} Computer Society, 2009.

\bibitem[Hanji et~al.(2022)Hanji, Mantiuk, Eilertsen, Hajisharif, and
  Unger]{hanji2022sihdr}
Param Hanji, Rafa{\l} Mantiuk, Gabriel Eilertsen, Saghi Hajisharif, and Jonas
  Unger.
\newblock {Comparison of single image {HDR} reconstruction methods - the
  caveats of quality assessment}.
\newblock In \emph{{SIGGRAPH} '22: Special Interest Group on Computer Graphics
  and Interactive Techniques Conference, Vancouver, BC, Canada, August 7 - 11,
  2022}, pages 1:1--1:8. {ACM}, 2022.

\bibitem[{HARMAN technology}(2018)]{harman2018hp5}
{HARMAN technology}.
\newblock {{ILFORD HP5 PLUS} Technical Information}.
\newblock
  \url{https://www.ilfordphoto.com/amfile/file/download/file/1903/product/691/},
  2018.

\bibitem[Heusel et~al.(2017)Heusel, Ramsauer, Unterthiner, Nessler, and
  Hochreiter]{heusel2017fid}
Martin Heusel, Hubert Ramsauer, Thomas Unterthiner, Bernhard Nessler, and Sepp
  Hochreiter.
\newblock {GANs Trained by a Two Time-Scale Update Rule Converge to a Local
  Nash Equilibrium}.
\newblock In \emph{Advances in Neural Information Processing Systems 30: Annual
  Conference on Neural Information Processing Systems 2017, December 4-9, 2017,
  Long Beach, CA, {USA}}, pages 6626--6637, 2017.

\bibitem[Hilliard et~al.(2025)Hilliard, Hilton, and
  Guillemaut]{hilliard2025hdrldm}
Jack~Oliver Hilliard, Adrian Hilton, and Jean{-}Yves Guillemaut.
\newblock {{HDR} Environment Map Estimation with Latent Diffusion Models}.
\newblock \emph{CoRR}, abs/2507.21261, 2025.

\bibitem[Ho and Salimans(2021)]{ho2021cfg}
Jonathan Ho and Tim Salimans.
\newblock {Classifier-Free Diffusion Guidance}.
\newblock In \emph{NeurIPS 2021 Workshop on Deep Generative Models and
  Downstream Applications}, 2021.

\bibitem[Hold{-}Geoffroy et~al.(2019)Hold{-}Geoffroy, Athawale, and
  Lalonde]{holdgeoffroy2019outdoor}
Yannick Hold{-}Geoffroy, Akshaya Athawale, and Jean{-}Fran{\c{c}}ois Lalonde.
\newblock {Deep Sky Modeling for Single Image Outdoor Lighting Estimation}.
\newblock In \emph{{IEEE} Conference on Computer Vision and Pattern
  Recognition, {CVPR} 2019, Long Beach, CA, USA, June 16-20, 2019}, pages
  6927--6935. Computer Vision Foundation / {IEEE}, 2019.

\bibitem[Hu et~al.(2022)Hu, Shen, Wallis, Allen{-}Zhu, Li, Wang, Wang, and
  Chen]{hu2021lora}
Edward~J. Hu, Yelong Shen, Phillip Wallis, Zeyuan Allen{-}Zhu, Yuanzhi Li,
  Shean Wang, Lu Wang, and Weizhu Chen.
\newblock {LoRA: Low-Rank Adaptation of Large Language Models}.
\newblock In \emph{The Tenth International Conference on Learning
  Representations, {ICLR} 2022, Virtual Event, April 25-29, 2022}.
  OpenReview.net, 2022.

\bibitem[{Imaging Resource}(2026)]{imagingresource}
{Imaging Resource}.
\newblock {Imaging Resource camera reviews: sample image galleries}.
\newblock \url{https://www.imaging-resource.com/}, 2026.
\newblock RAW and JPEG test shots; accessed 2026-10.

\bibitem[Immerk{\ae}r(1996)]{immerkaer1996fast}
John Immerk{\ae}r.
\newblock {Fast Noise Variance Estimation}.
\newblock \emph{Comput. Vis. Image Underst.}, 64\penalty0 (2):\penalty0
  300--302, 1996.

\bibitem[{International Electrotechnical Commission}(1999)]{iec1999srgb}
{International Electrotechnical Commission}.
\newblock {{IEC 61966-2-1:1999} Multimedia systems and equipment -- Colour
  measurement and management -- Part 2-1: Colour management -- Default {RGB}
  colour space -- {sRGB}}.
\newblock \url{https://webstore.iec.ch/en/publication/6169}, 1999.
\newblock Amendment 1, 2003.

\bibitem[{ITU-R}(2023)]{itu2023bt2408}
{ITU-R}.
\newblock {Guidance for Operational Practices in {HDR} Television Production}.
\newblock Report ITU-R BT.2408-7, International Telecommunication Union, 2023.

\bibitem[Ji et~al.(2023)Ji, Sawyer, and Narasimhan]{ji2023virtual}
Guanzhou Ji, Azadeh~O. Sawyer, and Srinivasa~G. Narasimhan.
\newblock {Virtual Home Staging: Inverse Rendering and Editing an Indoor
  Panorama under Natural Illumination}.
\newblock In \emph{Advances in Visual Computing ({ISVC} 2023)}, pages 329--342.
  Springer Nature Switzerland, 2023.

\bibitem[Ji et~al.(2025)Ji, Sawyer, and Narasimhan]{ji2025digital}
Guanzhou Ji, Azadeh~O. Sawyer, and Srinivasa~G. Narasimhan.
\newblock {Digital Kitchen Remodeling: Editing and Relighting Intricate Indoor
  Scenes from a Single Panorama}.
\newblock \emph{CoRR}, abs/2504.16086, 2025.

\bibitem[Kim et~al.(2012)Kim, Lin, Lu, S{\"{u}}sstrunk, Lin, and
  Brown]{kim2012incamera}
Seon~Joo Kim, Hai~Ting Lin, Zheng Lu, Sabine S{\"{u}}sstrunk, Stephen Lin, and
  Michael~S. Brown.
\newblock {A New In-Camera Imaging Model for Color Computer Vision and Its
  Application}.
\newblock \emph{{IEEE} Trans. Pattern Anal. Mach. Intell.}, 34\penalty0
  (12):\penalty0 2289--2302, 2012.

\bibitem[Klotz and Nayar(2025)]{klotz2025urbansky}
Jeremy Klotz and Shree~K. Nayar.
\newblock {Minimal sensing for orienting a solar panel}.
\newblock \emph{Solar Energy}, 300:\penalty0 113833, 2025.

\bibitem[{Kodak Alaris}(2025)]{kodak2025portra400}
{Kodak Alaris}.
\newblock {{KODAK PROFESSIONAL PORTRA 400} Film, Technical Data}.
\newblock KODAK Publication No. E-4050, 2025.

\bibitem[Lee et~al.(2018)Lee, An, and Kang]{lee2018hdrgan}
Siyeong Lee, Gwon~Hwan An, and Suk{-}Ju Kang.
\newblock {Deep Recursive {HDRI:} Inverse Tone Mapping Using Generative
  Adversarial Networks}.
\newblock In \emph{Computer Vision - {ECCV} 2018 - 15th European Conference,
  Munich, Germany, September 8-14, 2018, Proceedings, Part {II}}, pages
  613--628. Springer, 2018.

\bibitem[LeGendre et~al.(2019)LeGendre, Ma, Fyffe, Flynn, Charbonnel, Busch,
  and Debevec]{legendre2019deeplight}
Chloe LeGendre, Wan{-}Chun Ma, Graham Fyffe, John Flynn, Laurent Charbonnel,
  Jay Busch, and Paul~E. Debevec.
\newblock {DeepLight: Learning Illumination for Unconstrained Mobile Mixed
  Reality}.
\newblock In \emph{{IEEE} Conference on Computer Vision and Pattern
  Recognition, {CVPR} 2019, Long Beach, CA, USA, June 16-20, 2019}, pages
  5918--5928. Computer Vision Foundation / {IEEE}, 2019.

\bibitem[Levin and Fried(2025)]{levin2025differential}
Eran Levin and Ohad Fried.
\newblock {Differential Diffusion: Giving Each Pixel Its Strength}.
\newblock \emph{Comput. Graph. Forum}, 44\penalty0 (2), 2025.

\bibitem[Li et~al.(2025)Li, Ma, Zeng, Xu, Fang, Zhang, Wang, and
  Chen]{li2025sagiri}
Baiang Li, Sizhuo Ma, Yanhong Zeng, Xiaogang Xu, Youqing Fang, Zhao Zhang, Jian
  Wang, and Kai Chen.
\newblock {Exposure-Limited Image Enhancement with Generative Diffusion Prior}.
\newblock In \emph{{IEEE} International Conference on Computational
  Photography, {ICCP} 2025, Toronto, ON, Canada, July 21-23, 2025}, pages
  1--10. {IEEE}, 2025.

\bibitem[Li and Peers(2017)]{li2017crfnet}
Han Li and Pieter Peers.
\newblock {CRF-net: Single Image Radiometric Calibration using CNNs}.
\newblock In \emph{Proceedings of the 14th European Conference on Visual Media
  Production {(CVMP} 2017), London, United Kingdom, December 11-13, 2017},
  pages 5:1--5:9. {ACM}, 2017.

\bibitem[Liang et~al.(2025)Liang, He, Gojcic, Gilitschenski, Fidler,
  Vijaykumar, and Wang]{liang2025luxdit}
Ruofan Liang, Kai He, Zan Gojcic, Igor Gilitschenski, Sanja Fidler, Nandita
  Vijaykumar, and Zian Wang.
\newblock {LuxDiT: Lighting Estimation with Video Diffusion Transformer}.
\newblock In \emph{Advances in Neural Information Processing Systems 38: Annual
  Conference on Neural Information Processing Systems 2025, NeurIPS 2025, San
  Diego, CA, USA, December 2-7, 2025 / Mexico City, Mexico, November 30 -
  December 5, 2025}, 2025.

\bibitem[Liang et~al.(2018)Liang, Xu, Zhang, Cao, and Zhang]{liang2018hybrid}
Zhetong Liang, Jun Xu, David Zhang, Zisheng Cao, and Lei Zhang.
\newblock {A Hybrid $\ell_1$-$\ell_0$ Layer Decomposition Model for Tone
  Mapping}.
\newblock In \emph{IEEE/CVF Conference on Computer Vision and Pattern
  Recognition (CVPR)}, pages 4758--4766, 2018.

\bibitem[Lin et~al.(2026)Lin, Chen, Hu, Lin, and Liu]{lin2026boosting}
Yo{-}Tin Lin, Su{-}Kai Chen, Hou{-}Ning Hu, Yen{-}Yu Lin, and Yu{-}Lun Liu.
\newblock {{HDR} Reconstruction Boosting with Training-Free and
  Exposure-Consistent Diffusion}.
\newblock In \emph{{IEEE/CVF} Winter Conference on Applications of Computer
  Vision, {WACV} 2026, Tucson, AZ, USA, March 6-10, 2026}, pages 7513--7523.
  {IEEE}, 2026.

\bibitem[Lipman et~al.(2023)Lipman, Chen, Ben{-}Hamu, Nickel, and
  Le]{lipman2023flowmatching}
Yaron Lipman, Ricky T.~Q. Chen, Heli Ben{-}Hamu, Maximilian Nickel, and Matthew
  Le.
\newblock {Flow Matching for Generative Modeling}.
\newblock In \emph{The Eleventh International Conference on Learning
  Representations, {ICLR} 2023, Kigali, Rwanda, May 1-5, 2023}. OpenReview.net,
  2023.

\bibitem[Liu et~al.(2026)Liu, Liu, Wang, Chen, Zeng, and Liu]{liu2026expocm}
Aoyu Liu, Zhen Liu, Ziyi Wang, Dian Chen, Bing Zeng, and Shuaicheng Liu.
\newblock {ExpoCM: Exposure-Aware One-Step Generative Single-Image {HDR}
  Reconstruction}.
\newblock In \emph{{IEEE/CVF} Conference on Computer Vision and Pattern
  Recognition, {CVPR} 2026, Denver, CO, USA, June 3-7, 2026}, pages
  29909--29918. Computer Vision Foundation, 2026.

\bibitem[Liu et~al.(2020)Liu, Lai, Chen, Kao, Yang, Chuang, and
  Huang]{liu2020singlehdr}
Yu{-}Lun Liu, Wei{-}Sheng Lai, Yu{-}Sheng Chen, Yi{-}Lung Kao, Ming{-}Hsuan
  Yang, Yung{-}Yu Chuang, and Jia{-}Bin Huang.
\newblock {Single-Image {HDR} Reconstruction by Learning to Reverse the Camera
  Pipeline}.
\newblock In \emph{2020 {IEEE/CVF} Conference on Computer Vision and Pattern
  Recognition, {CVPR} 2020, Seattle, WA, USA, June 13-19, 2020}, pages
  1648--1657. Computer Vision Foundation / {IEEE}, 2020.

\bibitem[Lugmayr et~al.(2022)Lugmayr, Danelljan, Romero, Yu, Timofte, and
  Gool]{lugmayr2022repaint}
Andreas Lugmayr, Martin Danelljan, Andr{\'{e}}s Romero, Fisher Yu, Radu
  Timofte, and Luc~Van Gool.
\newblock {RePaint: Inpainting using Denoising Diffusion Probabilistic Models}.
\newblock In \emph{{IEEE/CVF} Conference on Computer Vision and Pattern
  Recognition, {CVPR} 2022, New Orleans, LA, USA, June 18-24, 2022}, pages
  11451--11461. {IEEE}, 2022.

\bibitem[Mantiuk and Azimi(2021)]{mantiuk2021pu21}
Rafa{\l}~K. Mantiuk and Maryam Azimi.
\newblock {{PU21:} {A} novel perceptually uniform encoding for adapting
  existing quality metrics for {HDR}}.
\newblock In \emph{Picture Coding Symposium, {PCS} 2021, Bristol, United
  Kingdom, June 29 - July 2, 2021}, pages 1--5. {IEEE}, 2021.

\bibitem[Mantiuk et~al.(2023)Mantiuk, Hammou, and Hanji]{mantiuk2023hdrvdp3}
Rafa{\l}~K. Mantiuk, Dounia Hammou, and Param Hanji.
\newblock {{HDR-VDP-3:} {A} multi-metric for predicting image differences,
  quality and contrast distortions in high dynamic range and regular content}.
\newblock \emph{CoRR}, abs/2304.13625, 2023.

\bibitem[Mantiuk et~al.(2024)Mantiuk, Hanji, Ashraf, Asano, and
  Chapiro]{mantiuk2024cvvdp}
Rafa{\l}~K. Mantiuk, Param Hanji, Maliha Ashraf, Yuta Asano, and Alexandre
  Chapiro.
\newblock {ColorVideoVDP}: A visual difference predictor for image, video and
  display distortions.
\newblock \emph{ACM Transactions on Graphics}, 43\penalty0 (4):\penalty0 1--20,
  2024.

\bibitem[Marnerides et~al.(2018)Marnerides, Bashford{-}Rogers, Hatchett, and
  Debattista]{marnerides2018expandnet}
Demetris Marnerides, Thomas Bashford{-}Rogers, Jonathan Hatchett, and Kurt
  Debattista.
\newblock {ExpandNet: {A} Deep Convolutional Neural Network for High Dynamic
  Range Expansion from Low Dynamic Range Content}.
\newblock \emph{Comput. Graph. Forum}, 37\penalty0 (2):\penalty0 37--49, 2018.

\bibitem[Meng et~al.(2022)Meng, He, Song, Song, Wu, Zhu, and
  Ermon]{meng2022sdedit}
Chenlin Meng, Yutong He, Yang Song, Jiaming Song, Jiajun Wu, Jun{-}Yan Zhu, and
  Stefano Ermon.
\newblock {SDEdit: Guided Image Synthesis and Editing with Stochastic
  Differential Equations}.
\newblock In \emph{The Tenth International Conference on Learning
  Representations, {ICLR} 2022, Virtual Event, April 25-29, 2022}.
  OpenReview.net, 2022.

\bibitem[Phongthawee et~al.(2024)Phongthawee, Chinchuthakun, Sinsunthithet,
  Jampani, Raj, Khungurn, and Suwajanakorn]{phongthawee2024diffusionlight}
Pakkapon Phongthawee, Worameth Chinchuthakun, Nontaphat Sinsunthithet, Varun
  Jampani, Amit Raj, Pramook Khungurn, and Supasorn Suwajanakorn.
\newblock {DiffusionLight: Light Probes for Free by Painting a Chrome Ball}.
\newblock In \emph{{IEEE/CVF} Conference on Computer Vision and Pattern
  Recognition, {CVPR} 2024, Seattle, WA, USA, June 16-22, 2024}, pages 98--108.
  {IEEE}, 2024.

\bibitem[{PIXLS.US community}(2026)]{rawpixls}
{PIXLS.US community}.
\newblock {raw.pixls.us: a library of camera raw samples}.
\newblock \url{https://raw.pixls.us/}, 2026.
\newblock Accessed 2026-11.

\bibitem[{Poly Haven}(2026)]{polyhaven2026hdris}
{Poly Haven}.
\newblock {HDRIs}.
\newblock \url{https://polyhaven.com/hdris}, 2026.
\newblock All assets licensed CC0 (\url{https://polyhaven.com/license}).
  Accessed 2026-11-03.

\bibitem[{Qwen Team}(2025)]{qwen2025imageedit2511}
{Qwen Team}.
\newblock {Qwen-Image-Edit-2511}.
\newblock \url{https://huggingface.co/Qwen/Qwen-Image-Edit-2511}, 2025.
\newblock Model card; checkpoint released December 2025. Accessed 2026-10-04.

\bibitem[{Qwen Team}(2026)]{qwen2026qwen38}
{Qwen Team}.
\newblock {Qwen3.8-27B-FP8}.
\newblock \url{https://huggingface.co/Qwen/Qwen3.8-27B-FP8}, 2026.
\newblock Model card; released August 2026. Accessed 2026-10-04.

\bibitem[Reinhard et~al.(2002)Reinhard, Stark, Shirley, and
  Ferwerda]{reinhard2002photographic}
Erik Reinhard, Michael~M. Stark, Peter Shirley, and James~A. Ferwerda.
\newblock {Photographic tone reproduction for digital images}.
\newblock \emph{{ACM} Trans. Graph.}, 21\penalty0 (3):\penalty0 267--276, 2002.

\bibitem[Santos et~al.(2020)Santos, Ren, and Kalantari]{santos2020maskhdr}
Marcel~Santana Santos, Tsang~Ing Ren, and Nima~Khademi Kalantari.
\newblock {Single image {HDR} reconstruction using a {CNN} with masked features
  and perceptual loss}.
\newblock \emph{{ACM} Trans. Graph.}, 39\penalty0 (4):\penalty0 80, 2020.

\bibitem[Sharma et~al.(2020)Sharma, Tan, and Cheong]{sharma2020crf}
Aashish Sharma, Robby~T. Tan, and Loong{-}Fah Cheong.
\newblock {Single-Image Camera Response Function Using Prediction Consistency
  and Gradual Refinement}.
\newblock In \emph{Computer Vision - {ACCV} 2020 - 15th Asian Conference on
  Computer Vision, Kyoto, Japan, November 30 - December 4, 2020, Revised
  Selected Papers, Part {VI}}, pages 19--35. Springer, 2020.

\bibitem[Talegaonkar et~al.(2026)Talegaonkar, He, McKenna, and
  Antipa]{talegaonkar2026singleshot}
Chinmay Talegaonkar, Jinshi He, Christopher McKenna, and Nicholas Antipa.
\newblock {Single-Shot {HDR} Recovery via a Video Diffusion Prior}.
\newblock \emph{CoRR}, abs/2605.11628, 2026.

\bibitem[{Venkatanath N.} et~al.(2015){Venkatanath N.}, {Praneeth D.}, {Maruthi
  Chandrasekhar Bh.}, Channappayya, and Medasani]{venkatanath2015piqe}
{Venkatanath N.}, {Praneeth D.}, {Maruthi Chandrasekhar Bh.}, Sumohana~S.
  Channappayya, and Swarup~S. Medasani.
\newblock {Blind image quality evaluation using perception based features}.
\newblock In \emph{Twenty First National Conference on Communications, {NCC}
  2015, Mumbai, India, February 27 - March 1, 2015}, pages 1--6. {IEEE}, 2015.

\bibitem[Wang et~al.(2023)Wang, Serrano, Pan, Chen, Myszkowski, Seidel,
  Theobalt, and Leimk{\"{u}}hler]{wang2023glowgan}
Chao Wang, Ana Serrano, Xingang Pan, Bin Chen, Karol Myszkowski, Hans{-}Peter
  Seidel, Christian Theobalt, and Thomas Leimk{\"{u}}hler.
\newblock {GlowGAN: Unsupervised Learning of {HDR} Images from {LDR} Images in
  the Wild}.
\newblock In \emph{{IEEE/CVF} International Conference on Computer Vision,
  {ICCV} 2023, Paris, France, October 1-6, 2023}, pages 10475--10485. {IEEE},
  2023.

\bibitem[Wang et~al.(2025)Wang, Xia, Leimk{\"{u}}hler, Myszkowski, and
  Zhang]{wang2025lediff}
Chao Wang, Zhihao Xia, Thomas Leimk{\"{u}}hler, Karol Myszkowski, and Xuaner
  Zhang.
\newblock {LEDiff: Latent Exposure Diffusion for {HDR} Generation}.
\newblock In \emph{{IEEE/CVF} Conference on Computer Vision and Pattern
  Recognition, {CVPR} 2025, Nashville, TN, USA, June 11-15, 2025}, pages
  453--464. Computer Vision Foundation / {IEEE}, 2025.

\bibitem[Wang et~al.(2022)Wang, Yang, Loy, and Liu]{wang2022stylelight}
Guangcong Wang, Yinuo Yang, Chen~Change Loy, and Ziwei Liu.
\newblock {StyleLight: {HDR} Panorama Generation for Lighting Estimation and
  Editing}.
\newblock In \emph{Computer Vision - {ECCV} 2022 - 17th European Conference,
  Tel Aviv, Israel, October 23-27, 2022, Proceedings, Part {XV}}, pages
  477--492. Springer, 2022.

\bibitem[Wang et~al.(2004)Wang, Bovik, Sheikh, and Simoncelli]{wang2004ssim}
Zhou Wang, Alan~C. Bovik, Hamid~R. Sheikh, and Eero~P. Simoncelli.
\newblock {Image quality assessment: from error visibility to structural
  similarity}.
\newblock \emph{{IEEE} Trans. Image Process.}, 13\penalty0 (4):\penalty0
  600--612, 2004.

\bibitem[Wei et~al.(2021)Wei, Guan, Liu, Kang, Li, Wu, and Hua]{wei2021beyond}
Wei Wei, Li Guan, Yue Liu, Hao Kang, Haoxiang Li, Ying Wu, and Gang Hua.
\newblock {Beyond Visual Attractiveness: Physically Plausible Single Image
  {HDR} Reconstruction for Spherical Panoramas}.
\newblock \emph{CoRR}, abs/2103.12926, 2021.

\bibitem[Wu et~al.(2025)Wu, Li, Zhou, Lin, Gao, Yan, Yin, Bai, Xu, Chen, Chen,
  Tang, Zhang, Wang, Yang, Yu, Cheng, Liu, Li, Zhang, Meng, Wei, Ni, Chen, Cao,
  Peng, Qu, Wu, Wang, Yu, Wen, Feng, Xu, Wang, Zhang, Zhu, Wu, Cai, and
  Liu]{wu2025qwenimage}
Chenfei Wu, Jiahao Li, Jingren Zhou, Junyang Lin, Kaiyuan Gao, Kun Yan,
  Sheng-ming Yin, Shuai Bai, Xiao Xu, Yilei Chen, Yuxiang Chen, Zecheng Tang,
  Zekai Zhang, Zhengyi Wang, An Yang, Bowen Yu, Chen Cheng, Dayiheng Liu,
  Deqing Li, Hang Zhang, Hao Meng, Hu Wei, Jingyuan Ni, Kai Chen, Kuan Cao,
  Liang Peng, Lin Qu, Minggang Wu, Peng Wang, Shuting Yu, Tingkun Wen, Wensen
  Feng, Xiaoxiao Xu, Yi Wang, Yichang Zhang, Yongqiang Zhu, Yujia Wu, Yuxuan
  Cai, and Zenan Liu.
\newblock {Qwen-Image Technical Report}.
\newblock \emph{CoRR}, abs/2508.02324, 2025.

\bibitem[Wu et~al.(2026)Wu, Su, Ma, Liao, and Mantiuk]{wu2026x2hdr}
Ronghuan Wu, Wanchao Su, Kede Ma, Jing Liao, and Rafał~K. Mantiuk.
\newblock {X2HDR: HDR Image Generation in a Perceptually Uniform Space}.
\newblock In \emph{Advances in Neural Information Processing Systems
  (NeurIPS)}, 2026.

\bibitem[Yu et~al.(2021)Yu, Liu, Long, Dong, Zou, and Xiao]{yu2021lanet}
Hanning Yu, Wentao Liu, Chengjiang Long, Bo Dong, Qin Zou, and Chunxia Xiao.
\newblock {Luminance Attentive Networks for {HDR} Image and Panorama
  Reconstruction}.
\newblock \emph{Comput. Graph. Forum}, 40\penalty0 (7):\penalty0 181--192,
  2021.

\bibitem[Yu et~al.(2026{\natexlab{a}})Yu, Ma, He, Isikdogan, Xu, Smirnov,
  Salamanca, Mi, Delgado, Yu, Philip, Li, Wang, and Debevec]{yu2026diffhdr}
Zhengming Yu, Li Ma, Mingming He, Leo Isikdogan, Yuancheng Xu, Dmitriy Smirnov,
  Pablo Salamanca, Dao Mi, Pablo Delgado, Ning Yu, Julien Philip, Xin Li,
  Wenping Wang, and Paul Debevec.
\newblock {DiffHDR: Re-exposing LDR Videos with Video Diffusion Models}.
\newblock In \emph{Computer Vision -- ECCV 2026}, pages 336--357. Springer
  Nature Switzerland, 2026{\natexlab{a}}.

\bibitem[Yu et~al.(2026{\natexlab{b}})Yu, Ma, He, Zhang, Wang, Yu, and
  Debevec]{yu2026promoting}
Zhengming Yu, Li Ma, Mingming He, Jingdong Zhang, Jionghao Wang, Ning Yu, and
  Paul Debevec.
\newblock {Promoting {LDR} Panoramas to {HDRI} with Rendering-Space Losses}.
\newblock In \emph{Proceedings of the Special Interest Group on Computer
  Graphics and Interactive Techniques Conference Posters, {SIGGRAPH} Posters
  2026, Los Angeles, CA, USA, July 19-23, 2026}, pages 72:1--72:3. {ACM},
  2026{\natexlab{b}}.

\bibitem[Zhan et~al.(2021)Zhan, Zhang, Yu, Chang, Lu, Ma, and
  Xie]{zhan2021emlight}
Fangneng Zhan, Changgong Zhang, Yingchen Yu, Yuan Chang, Shijian Lu, Feiying
  Ma, and Xuansong Xie.
\newblock {EMLight: Lighting Estimation via Spherical Distribution
  Approximation}.
\newblock In \emph{Thirty-Fifth {AAAI} Conference on Artificial Intelligence,
  {AAAI} 2021, Thirty-Third Conference on Innovative Applications of Artificial
  Intelligence, {IAAI} 2021, The Eleventh Symposium on Educational Advances in
  Artificial Intelligence, {EAAI} 2021, Virtual Event, February 2-9, 2021},
  pages 3287--3295. {AAAI} Press, 2021.

\bibitem[Zhang and Lalonde(2017)]{zhang2017outdoorhdr}
Jinsong Zhang and Jean{-}Fran{\c{c}}ois Lalonde.
\newblock {Learning High Dynamic Range from Outdoor Panoramas}.
\newblock In \emph{{IEEE} International Conference on Computer Vision, {ICCV}
  2017, Venice, Italy, October 22-29, 2017}, pages 4529--4538. {IEEE} Computer
  Society, 2017.

\bibitem[Zhang et~al.(2014)Zhang, Shen, and Li]{zhang2014vsi}
Lin Zhang, Ying Shen, and Hongyu Li.
\newblock {{VSI:} {A} Visual Saliency-Induced Index for Perceptual Image
  Quality Assessment}.
\newblock \emph{{IEEE} Trans. Image Process.}, 23\penalty0 (10):\penalty0
  4270--4281, 2014.

\end{thebibliography}
}
\end{document}